\documentclass[%
    reprint,
    superscriptaddress,
    amsmath,
    amssymb,
    aps,
    pra,
]{revtex4-2}

\usepackage{xcolor}
\usepackage{booktabs}
\usepackage{multirow}

\usepackage{algorithm}
\usepackage{algorithmic}

\usepackage{graphicx}
\usepackage{dcolumn}
\usepackage{bm}
\usepackage{hyperref}
\usepackage[caption=false]{subfig}

\begin{document}

\title{Nonlocal Monte Carlo via Reinforcement Learning}

\author{Dmitrii Dobrynin}
\email{d.dobrynin@fz-juelich.de}
\affiliation{
    Peter Grünberg Institut (PGI-14), Forschungszentrum Jülich GmbH, Jülich, Germany
}
\affiliation{
    RWTH Aachen University, Aachen, Germany
}

\author{Masoud Mohseni}
\email{masoud.mohseni@hpe.com}
\affiliation{
    Emergent Machine Intelligence, Hewlett Packard Labs, CA, USA
}

\author{John Paul Strachan}
\email{j.strachan@fz-juelich.de}
\affiliation{
    Peter Grünberg Institut (PGI-14), Forschungszentrum Jülich GmbH, Jülich, Germany
}
\affiliation{
    RWTH Aachen University, Aachen, Germany
}

\date{\today}

\begin{abstract}

Optimizing or sampling complex cost functions of combinatorial optimization problems is a longstanding challenge across disciplines and applications. When employing family of conventional algorithms based on Markov Chain Monte Carlo (MCMC) such as  simulated annealing or parallel tempering, one assumes homogeneous (equilibrium) temperature profiles across input. This instance independent approach was shown to be ineffective for the hardest benchmarks near a computational phase transition when the so-called overlap-gap-property holds. In these regimes conventional MCMC struggles to unfreeze rigid variables, escape suboptimal basins of attraction, and sample high-quality and diverse solutions. In order to mitigate these challenges, Nonequilibrium Nonlocal Monte Carlo (NMC) algorithms were proposed that leverage inhomogeneous temperature profiles thereby accelerating exploration of the configuration space without compromising its exploitation. Here, we employ deep reinforcement learning (RL) to train the nonlocal transition policies of NMC which were previously designed phenomenologically. We demonstrate that the resulting solver can be trained solely by observing energy changes of the configuration space exploration as RL rewards and the local minimum energy landscape geometry as RL states. We further show that the trained policies improve upon the standard MCMC-based and nonlocal simulated annealing on hard uniform random and scale-free random 4-SAT benchmarks in terms of residual energy, time-to-solution, and diversity of solutions metrics.

\end{abstract}

\maketitle

\section{\label{sec:introduction}Introduction}

Geometry of configuration spaces of combinatorial optimization or inference problems 
has long been recognized as the culprit of algorithmic hardness \cite{mezard2009, zdeborova2016, gamarnik2025}. 
Complex random correlations of variables create energy landscapes full of local minima, 
saddle regions, disjoint basins of attraction with large energy barriers 
making their navigation exponentially difficult 
for solvers. When general-purpose exact algorithms fail, a common approach
is to employ simple heuristics, such as physics-inspired 
Simulated Annealing (SA) \cite{kirkpatrick1983} or Parallel Tempering (PT) \cite{earl2005}, 
and achieve acceptable approximations through sufficient computation.
Designing hardware accelerators implementing such simple algorithms to make them 
more energy efficient and fast is a promising research direction gaining ground in the recent years. 
For example, classical Ising machines \cite{mohseni2022}
or their quantum annealing counterparts \cite{hauke2020} could improve efficiency of optimization 
with native physical implementation of algorithmic operations or unique potentially advantageous  
features such as intrinsic noise \cite{cai2020} or quantum tunneling \cite{albash2018}.

\begin{figure}[ht]
    \includegraphics[trim={19.5cm 2cm 20cm 23cm}, clip, width=\linewidth]{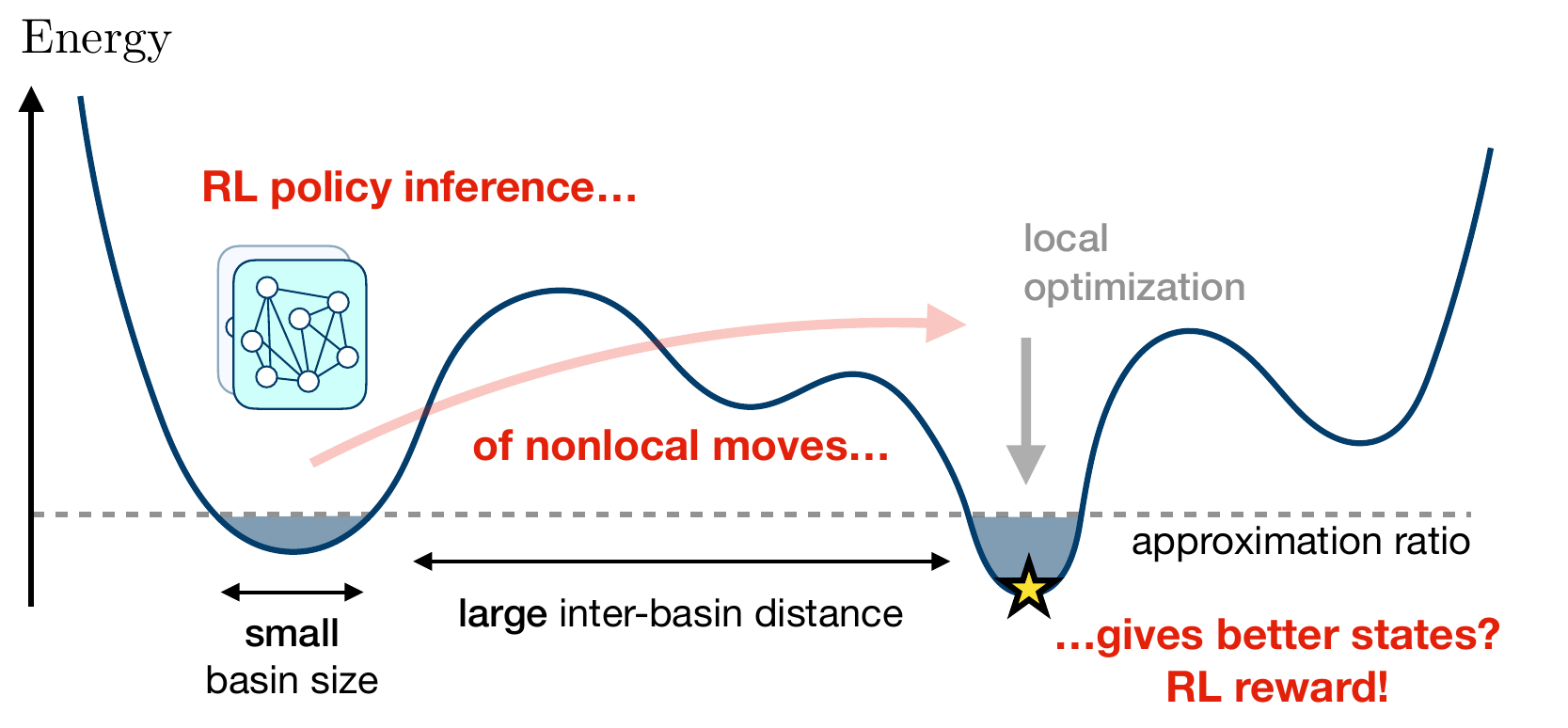}
    \caption{\label{fig:figure_one} 
        Energy landscape of a potentially hard problem featuring large distance between small basins of good solutions:
        nonlocal moves are essential to efficiently explore the configuration space. 
        RLNMC method uses rewards of finding better solutions to train a deep recurrent policy executing nonlocal transitions.}
\end{figure}

Recent developments on the algorithmic barriers for optimizing random structures 
-- the overlap-gap-property \cite{gamarnik2021, gamarnik2025} -- predicts failure to find good solutions for algorithms that are 
``local'' and therefore ``stable'' , regardless of the accelerator used. 
However, many practical heuristics are usually referred to as ``local search'' for they
quickly update intermediate states by small guided steps. The challenge is illustrated in Fig.~\ref{fig:figure_one}. As a result, additional 
heuristics have been introduced over the years to mediate ``nonlocal'' (or cluster) 
moves in the configuration space \cite{swendsen1987, wolff1989, houdayer2001, hamze2004, selby2014, zhu2015, hen2017}. 
However, these methods often  break down for hard problem classes, e.g. below a spin-glass phase 
transition because of frustrations, or become ineffective in higher dimensional problems due to percolation \cite{houdayer2001}. 
To make a step towards energy landscape sensitive cluster moves,
Nonequilibrium Nonlocal Monte Carlo (NMC) family of solvers was introduced in \cite{mohseni2021, mohseni2023}. NMC 
efficiently analyzes the local geometry of the energy landscape to construct nonlocal transitions
from every unique basin of attraction. The nonequilibrium \textit{inhomogeneous} temperature profiles of MC sampling
mediate transitions inaccessible to simple local search routines without typical erasure of information 
from random restarts or homogeneous high temperatures.

A promising direction to accelerate combinatorial optimization leverages deep learning \cite{bengio2020} to design novel algorithms from data.
In this case, a model parametrized by a neural network 
is trained to either become a subroutine in an existing algorithm, or solve problems in an end-to-end fashion. 
Among all standard learning methods, here we propose reinforcement learning (RL) \cite{sutton2018}
to search for nonlocal transitions without supervision. We merge the ideas of deep learning for optimization and physics-inspired 
algorithms by training the NMC-type nonlocal moves with RL, a method we refer to as RLNMC (see Fig.~\ref{fig:figure_one}). 
In this paper, we show promising improvements of RLNMC over the standard MCMC-based Simulated Annealing (MCMC SA), 
as well as the Nonlocal Monte Carlo assisted SA (NMC SA) when tested on different problem classes and using a variety of metrics.
Furthermore, RLNMC shows better generalization than NMC on problem sizes larger than those it was trained on, without additional 
training or hyperparameter tuning.

Sec.~\ref{sec:background} begins with a background on related work and motivation for the RLNMC method. Next, in Sec.~\ref{sec:preliminaries}
the reader is introduced to the problems and methods of interest. We then outline the RLNMC architecture in Sec.~\ref{sec:RLNMC_method}. 
Finally, we show the numerical simulations of time-to-solution, energy, and diversity of solutions in Sec.~\ref{sec:numerical_simulations}. 
Additionally, the methods and supplementary sections provide more details on the implementation of every module, 
and the corresponding hyperparameters. The RLNMC implementation is made available at \cite{rlnmc_code}.

\section{\label{sec:background}Background and motivation}
Deep learning for combinatorial optimization has seen promising developments 
in recent years, featuring a variety of types of 
learning and modes of operation. In particular, supervised, unsupervised, and reinforcement learning 
paradigms have been used to train ML models to solve combinatorial problems ~\cite{bengio2020}.
Furthermore, ML models have been trained to act as standalone solvers \cite{bello2017}, 
as a guides/helpers for a rigorous conventional algorithms \cite{li2018}, 
as imitating subroutines \cite{he2014}, etc. As a result, we would like to clearly 
position our RLNMC method within the existing range of ML methods for combinatorial optimization.

Firstly, RLNMC trainable policy is not designed to find solutions to optimization problems. This task is ``outsourced''
to the Monte-Carlo sampling subroutines of NMC. RLNMC is trained to recognize patterns in the correlations of 
variables/energy landscape, which are used to construct nonlocal transitions in the configuration space. 
One can interpret our approach as discovering generalized adaptive cluster moves 
\cite{swendsen1987, wolff1989, houdayer2001, hamze2004, selby2014, zhu2015, hen2017} 
for the accelerating exploration of configuration space. Thus, RLNMC can act as a trainable ``helper'' for optimization.

Secondly, in RLNMC we use reinforcement learning to train a deep policy.
Notable works on deep end-to-end solvers trained by RL include
\cite{bello2017, kool2019} for permutation problems like TSP, \cite{dai2018, barrett2020, barrett2022, tonshoff2023} 
for combinatorial problems defined on graphs like MAX-CUT
with a review given in \cite{mazyavkina2021}. The works that are closer to this paper in
their goal to \textit{guide} optimization are \cite{li2018, ahn2020, mills2020, kool2021}.
In this regime, one should be careful not to exaggerate the importance of a deep model for the overall success of the solver.
For example, it has been shown that replacing the ML part of certain tree search solver with 
random numbers could have similar effect on performance \cite{bother2022}. 
However, within the same study it has been shown that a deep RL policy indeed could make a net positive impact in solving MIS in \cite{ahn2020}. 

Third, utilizing graph neural networks (GNN) \cite{zhou2020} as a deep model architecture
is chosen for RLNMC in our approach. For a review of previous works on using
GNNs for optimization see~\cite{cappart2023}. 
A variety of interesting practical problems are higher-order, i.e. defined on a hypergraph and not on graphs.
Past works \cite{perdomo2019, valiante2021, dobrynin2024, dobrynin2024b} have investigated the algorithmic penalties 
of using combinatorial optimization problem embeddings such as quadratization that reduces the 
locality of interaction from hypergraphs to graphs.
We aim to avoid quadratization methods and in order for RLNMC to be natively applicable 
to problems of arbitrary order, we will define a deep policy with attention on a factor graph.

Trainable GNNs have been used as standalone solvers in the recent works.
Spin glass models have been optimized with GNNs trained by RL in \cite{fan2023}. 
Similarly, GNNs were employed for solving QUBO (quadratic unconstrained binary optimization) 
problems in \cite{schuetz2022}, and further extended to Potts model \cite{schuetz2022b} 
and higher-order (hypergraph) problems in \cite{heydaribeni2024}. Further positive examples 
were also demonstrated in \cite{pugacheva2024, langedal2024, hu2024}. 
However, the works \cite{angelini2023, boettcher2023, boettcher2023a, gamarnik2023} 
illustrate some of the theoretical challenges and open issues facing GNNs as solvers.
In particular, \cite{gamarnik2023} argues that MAX-CUT on random graphs is a problem that can be efficiently approximated,
and higher-order problems (e.g. MAX-CUT on hypergraphs) could be targeted in the future. Motivated by the 
discussion of the aforementioned works, here we choose to employ GNNs for \textit{augmenting} local solvers 
with nonlocal moves (compared to solving them end-to-end) and target challenging higher-order problems.

In a similar spirit to cluster nonlocal updates aiming to mitigate the locality of MCMC, 
variational autoregressive neural (VAN) methods have been proposed 
to accelerate sampling of the Gibbs-Boltzmann distribution $p(\bm{s}) = \exp{\{-H(\bm{s})\}}/Z$.
VANs train an autoregressive neural network function  $q_{\theta}(\bm{s})$
by minimizing the K-L divergence with $p(\bm{s})$.
The trained $q_{\theta}(\bm{s})$ is either directly used for sampling \cite{wu2019},
or acts as a proposal function of MCMC Metropolis update rule to remove bias \cite{nicoli2020, mcnaughton2020, wu2021}.
The further studies of VANs include their use in a combinatorial optimization setting in
\cite{hibat2021, khandoker2023, sanokowski2023, ma2024}, or advanced architecture suggestions 
in \cite{pan2021, biazzo2023, sanokowski2023, biazzo2024, ma2024, delbono2025a}.
Applicability of VANs for computationally hard tasks and first theoretical studies were carried out 
in \cite{ciarella2023, ghio2024, delbono2025b}. In contrast to the goals of variationally assisted MCMC, 
with RLNMC we aim to predict nonlocal nonequilibrium moves to reach good solutions fast
and do not yet attempt to model the Gibbs-Boltzmann probability distribution. 

\section{\label{sec:preliminaries}Preliminaries}
\subsection{\label{sec:problems_of_interest}Problems of interest}
We are interested in higher-order binary unconstrained combinatorial optimization 
formulated using either the PUBO cost (energy) function,
\begin{equation}
    E_{\rm{PUBO}} = \sum_{i_1 \le \dots \le i_p}^N Q_{i_1, \dots, i_p}x_{i_1}\dots x_{i_p} + \dots + \sum_{i=1}^N b_ix_i + C\,,
    \label{eq:pubo}
\end{equation}
or equivalently the p-spin Ising cost function,
\begin{equation}
    E_{\rm{Ising}} = \sum_{i_1 \le \dots \le i_p}^N J_{i_1, \dots, i_p}\sigma_{i_1}\dots \sigma_{i_p} + \dots + \sum_{i=1}^N h_i\sigma_i + C\,,
    \label{eq:pising}
\end{equation}
where $x\in\{0,1\}$ and $\sigma\in\{-1,1\}$ respectively. Deciding its ground state constitutes an 
NP-Complete problem (NP-Hard in the case of optimization) \cite{garey1990}.
Recent years have seen an increased interest in designing hardware accelerators, Ising machines, 
to reduce the time and energy required to optimize the second and higher-order problem 
\cite{mohseni2022, bybee2023, sharma2023, bhattacharya2024, nikhar2024, pedretti2025}.

Many problems of interest can be readily represented by either Eq.~\ref{eq:pubo} or Eq.~\ref{eq:pising}.
For example, (MAX-)$k$-SAT is described by the conjugate normal form (CNF) of 
$N$ boolean variables and $M$ boolean functions of up to $K$ variables each:
\begin{equation}
    C_1 \wedge \dots \wedge C_M = (\neg x_{i^1_1} \vee x_{i^1_2} \vee \dots \vee x_{i^1_K})\wedge \dots \,,
    \label{eq:K-SAT}
\end{equation}
where $C_m$, $m\in\{1,\dots, M\}$ are called clauses, the indices $i^m_k \in \{1,\dots,N\}$, $k \in \{1,\dots, K\}$, 
and some of the variables $x$ are negated depending on a particular problem instance. The task of maximizing 
the number of satisfied logical clauses in Eq.~\ref{eq:K-SAT} is readily written as the energy minimization 
of Eq.~\ref{eq:pubo} as follows:
\begin{equation}
    E_{k\rm{-SAT}} = x_{i^1_1}(1 - x_{i^1_2})\dots (1 - x_{i^1_K}) + \dots
    \label{eq:ksat_as_pubo}
\end{equation}

Another example that can be easily formulated as Eq.~\ref{eq:pising} 
is the $p$-order hypergraph MAX-CUT optimization \cite{chen2019}:
\begin{equation}
    E_{\rm{max-cut}} = \sum_{i_1 \le \dots \le i_p \in \rm{edges}}^N \sigma_{i_1}\dots \sigma_{i_p}\,.
    \label{eq:hyper_max_cut_as_pising}
\end{equation}
Compared to the standard second-order (graph) version of MAX-CUT, the hypergraph MAX-CUT 
was recently suggested as a suitable testbed for new heuristic solvers because of its OGP 
features for $p \ge 4$, a similar scenario to $k$-SAT with $k \ge 4$ \cite{gamarnik2021, gamarnik2023}. 
If necessary, the weighted versions of MAX-$k$-SAT and MAX-CUT are encoded using 
$Q_{ijk\dots}$ and $J_{ijk\dots}$ coefficients.

For the reinforcement learning training and benchmarking purposes of this study,
we adopt the following optimization problems:
\begin{itemize}
    \item Uniform random 4-SAT of $N = 500,\,1000,\,2000$, and $M = 4942, 9884, 19768$ respectively, which corresponds 
    to the the clause to variable ratio $\alpha = M/N = 9.884$ in the rigidity phase (see App.~\ref{app:4sat_uf}).
    This class is treated as optimization (i.e. MAX-4-SAT) problem, and we are interested in minimizing the average energy across multiple replicas.
    \item Scale-free random 4-SAT (decision version) of $N=250$ variables 
    and $M = 2300$ clauses (see App.~\ref{app:4sat_sf}). This problem class is designed to resemble many 
    structured industrial problems and features a non-uniform distribution of the variables. 
    Therefore some selected variables have a much larger degree of connectivity, in contrast to the uniform random class. 
    Scale-free random 4 SAT is treated as a decision problem and we are 
    looking for the ground state ($0$ violated clauses).
\end{itemize}
For each problem class and size we use $384$ instances: $64$ instances are employed for hyperparameter 
optimization/training, while the remaining $320$ instances are used for the reported benchmarking results 
in the sections below. The details on the generation of benchmarks, their properties, 
and background are provided in App.~\ref{app:benchmarks}.

\subsection{\label{sec:sa}Simulated Annealing}
Simulated annealing (\textbf{SA}) \cite{kirkpatrick1983} is a general purpose heuristic algorithm in which the temperature 
$T = 1/\beta$ of Markov Chain Monte Carlo (MCMC) sampling is gradually reduced from 
initially high values giving a relatively large acceptance rate to a small value which favors exploitation.
MCMC sampling typically follows either the Metropolis-Hastings rule, or the Gibbs (heat bath) rule 
both designed to sample from the Gibbs distribution $p(\mathbf{x}) = \exp{(-\beta E(\mathbf{x}))}/Z$ at a given temperature 
$\beta$, where $Z$ is the partition function.
We use MCMC-based SA as base heuristic in this paper, which will be augmented with nonlocal moves (NMC in Sec.~\ref{sec:NMC})
and later with reinforcement learning (RLNMC in Sec.~\ref{sec:RLNMC_method}).

SA is a stochastic algorithm running for a total $N_{\rm{total\,sw}}$ MCMC sweeps 
that succeeds only with a certain probability of success ($\rm{POS}$), for which cumulative time-to-solution metric needs to be estimated.
Time-to-solution$_p$ ($\mathrm{TTS}_{p}$) is defined as the total number of runs required to succeed with 
the probability $p$ at least once, multiplied by the cost $\tau$ of one individual run. We choose the commonly used $p = 0.99$, which gives
\begin{equation}
    \mathrm{TTS}_{99} = \tau (N_{\rm{total\,sw}}) \times \frac{\log{(1 - 0.99)}}{\log{(1 - \rm{POS}(N_{\rm{total\,sw}}))}},
    \label{eq:tts}
\end{equation}
if $\rm{POS}(N_{\rm{total\,sw}})\le 0.99$, or $\mathrm{TTS}_{99} = \tau (N_{\rm{total\,sw}})$ if $\rm{POS}(N_{\rm{total\,sw}})> 0.99$ 
(at least one run is required).

Because SA is subject to hyperparameter optimization, we describe in detail the SA setup of this paper in App.~\ref{app:sa}. 
In short, for the chosen temperature schedule function, $N_{\rm{total\,sw}}$ 
and the initial $\beta_i$ and final $\beta_f$ temperatures of SA 
are tuned to produce the best TTS$_{99}$ of Eq.~\ref{eq:tts}.
For NMC and RLNMC modifications of SA described below, the schedules of $\beta$, as well as the total 
number of sweeps will remain the same as for SA.

\subsection{\label{sec:nmc}Nonequilibrium Nonlocal Monte Carlo}
Nonequilibrium Nonlocal Monte Carlo is a family of methods suggested in \cite{mohseni2021, mohseni2023}
as means to unfreeze highly correlated/rigid variables in hard combinatorial optimization problems 
which we refer to as \textit{backbones} for the remaining of this paper.
NMC was shown to be competitive with the state-of-the-art specialized K-SAT solver, Backtracking Survey 
Propagation algorithm \cite{marino2016}, and able to robustly sample high quality configurations with a strong 
frozen component in the case of large very hard uniform random 4-SAT problems. 

In \cite{mohseni2021} the construction of backbones can be summarized as follows. 
NMC begins in a local minimum~$\mathbf{\sigma}^*$. 
First, the surrogate Hamiltonian $H^* = H - \lambda\sum_{i = 1}^N{\sigma}^*_i\sigma_i$ is used to localize the state to a particular 
basin of attraction. Second, the absolute correlations $\tilde{J}_{ij\dots} = \frac{1}{\beta}|\mathrm{atanh}\langle\sigma_i \sigma_j\dots \rangle|$
and/or magnetizations $\tilde{h}_{i} = \frac{1}{\beta}|\mathrm{atanh}\langle\sigma_i\rangle|$ are estimated in the given basin.
Third, a hyperparameter optimized cutoff threshold $r$ is defined, and the backbone status is assigned to 
variables having their correlations larger than $r$.

Once identified, the backbones are either sampled at an increased temperature (nonequilibrium stage), or 
a rejection-free transition is made (nonlocal, or infinite $T$ stage), with the 
non-backbone subgraph fixed. The next step is to sample at a base, low temperature 
the non-backbone variables with the backbone variables fixed in their new excited state.
Finally, the ``equilibrium'' sampling of the entire problem is performed obtaining a new local minimum 
for the next NMC jump.

\begin{figure}[ht]
    \centering
    \includegraphics[trim={3.5cm 20.5cm 47cm 5.5cm}, clip, width=0.75\linewidth]{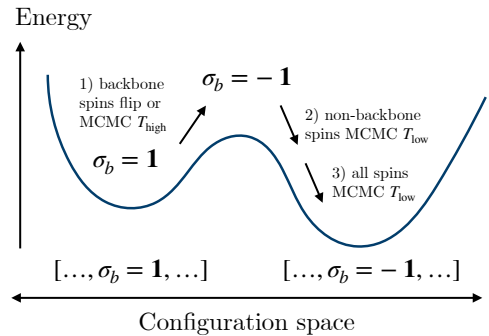}
    \caption{\label{fig:nmc_sketch} 
        Illustration of the NMC stages in \cite{mohseni2021}. 
        1) The ``backbone'' MCMC stage excites the variables, escaping the basin of attraction.
        2) The non-backbone MCMC stage lowers the energy in the new basin so that 
        returning to the original basin is ruled out.
        3) Final all-spins MCMC stage corrects inconsistencies because of possible errors of the backbone inference.
    }
\end{figure}

An intuitive visualization of the NMC is given in Fig.~\ref{fig:nmc_sketch}. 
A backbone variable corresponds to its basin of attraction, while other non-backbone variables are 
easily changed within each basin. NMC raises the temperature from $T_{\rm{low}}$ to 
$T_{\rm{high}}$ for backbones accelerating the transition between the basins. 
If one used $T_{\rm{high}}$ for every variable uniformly, 
then very little energy landscape geometry information acquired at $T_{\rm{low}}$ would be preserved. 
This approximately corresponds to a random restart of optimization. 
Backbones can be understood as variables that remain largely unchanged within a single 
local minimum basin of attraction. As a result, if the backbones are large, 
then escaping local minima becomes a formidable challenge for any ``local'' 
algorithm able to make only small steps in the configuration space at once. 
It was shown that the frozen phase \cite{gamarnik2022} of random combinatorial problems featuring such backbones 
approximately corresponds to the \textit{algorithmic} phase transition of linear time algorithms 
separating the instances that are in principle solvable from those that are not.

NMC is a peculiar algorithm exciting the variables which have strong preference to be in 
their respective states and is biased towards exploration. In contrast to NMC, there exist strategies fixing the variables 
with strong correlations/magnetizations to simplify the problem \cite{karimi2017}. For example, recursive QAOA 
\cite{bravyi2020, bravyi2022, finzgar2024} estimates the two/one-point correlations using the low-energy
quantum states $|\psi\rangle$: $\langle\psi|Z_i Z_j |\psi\rangle$, and removes the 
corresponding nodes with high correlations from the interaction graph. 

\subsubsection{\label{sec:NMC}Nonlocal Simulated Annealing}
Following the intuition of Nonequlibrium Nonlocal Monte Carlo in \cite{mohseni2021} 
we propose a modified version of SA with integrated nonlocal moves (still calling it \textbf{NMC} for simplicity of notation). 
As will be shown below, standard (MCMC) SA is relatively successful at quickly solving/approximating the problems.
However, it tends to slow-down when the temperature of sampling is too low (e.g. below a spin-glass phase transition). Therefore, we suggest that 
the nonlocal moves are carried out exactly when this degradation of performance is observed.

As a result, we start NMC with running SA according to the $\beta$ schedule form $\beta_i$ to 
a hyperparameter optimized $\beta_{\rm{NMC}} = 1/T_{\rm{NMC}} \in (\beta_i, \beta_f)$.
Next, as the temperature is further decreased from $\beta_{\rm{NMC}}$ to $\beta_f$, 
nonlocal transitions are called, which are intended to improve the suboptimal solutions reached by SA. 
$\beta_{\rm{NMC}}$ is estimated from the temperature at which the time-to-solution metric
reaches its minimum, as will be shown below in Sec.~\ref{sec:time_to_solution}.
The other hyperparameter values, such as the frequency of NMC transitions, the backbone 
classification threshold, and others are optimized using the same 64 instances employed 
for hyperparameter optimization of SA.

The construction of backbone clusters subject to the NMC move in this paper is a simplified version of \cite{mohseni2021}. 
First, we define a hyperparameter cutoff threshold $r$.
Next, if the NMC jump is called, the local fields are computed for every variable, defined as
$$H_i \equiv [E(x_i \to \bar{x}_i) - E(x_i)]/2.$$
In the language of K-SAT optimization, $H_i$ is the $\mathrm{``make} - \mathrm{break"}$ value for each variable.
$H_i$ are used instead of the estimated $\tilde{h}_i = \frac{1}{\beta}|\mathrm{atanh\langle \sigma_i\rangle)}|$ 
with the pinned surrogate Hamiltonian of \cite{mohseni2021}. This allows us to avoid the penalty of running sampling
to estimate the marginal probability distributions, i.e., local magnetizations $\langle \sigma_i \rangle$, or higher-order correlation functions, $\langle \sigma_i \ldots \sigma_j\rangle$, which otherwise would necessarily require the computational cost 
equivalent to multiple MCMC sweeps.

When $|H_i| > r$, then the variable $i$ is considered a backbone variable and its state will be randomized
simultaneously in parallel for every $i$. 
Once the backbone is randomized, it is fixed for a duration 
of one MCMC sweep over the non-backbone variables. This sweep is biasing the algorithm to explore a new position in the 
energy landscape instead of returning to the original state. Finally, the problem is optimized in the 
new basin of attraction for $N_{\rm{full\,sw}}$ number of sweeps over all variables. 
Alg.~\ref{alg:nmc} in Sec.~\ref{app:NMC} further elaborates on the NMC move algorithm used in this paper. 

The implementation of SA and NMC in this work is GPU parallelized and implemented in JAX \cite{jax2018github}.
In order to compare the computational efforts of SA and NMC, we use the MC sweeps (MCS) measure, which is hardware-free.
According to the description above, we can assign $N_{\rm{NMC\,sw}} = N_{\rm{full\,sw}} + 2$ MCS computational cost to each NMC transition: 
one MCS for the backbone variables randomization, one MCS for the non-backbone sweep, 
and $N_{\rm{full\,sw}}$ full sweeps for the final stage of the NMC move (details in Sec.~\ref{app:NMC}).

\section{\label{sec:results}Results}
\subsection{\label{sec:RLNMC_method}RLNMC method}
We propose reinforcement learning (RL) \cite{sutton2018} as a natural framework for the discovery of nonlocal moves. Without supervision, 
there is a potential to discover transitions not limited by handcrafted heuristics, equilibrium or stability 
requirements. Furthermore, it opens the possibility for a solver to be adaptive to each individual problem instance, 
conducting search in the algorithmic space simultaneously with the configuration space.
In this work we combine RL with NMC by introducing a general \textbf{RLNMC} method. 

\begin{figure}[ht]
    \centering
    \includegraphics[trim={40cm 17cm 5cm 3.5cm}, clip, width=\linewidth]{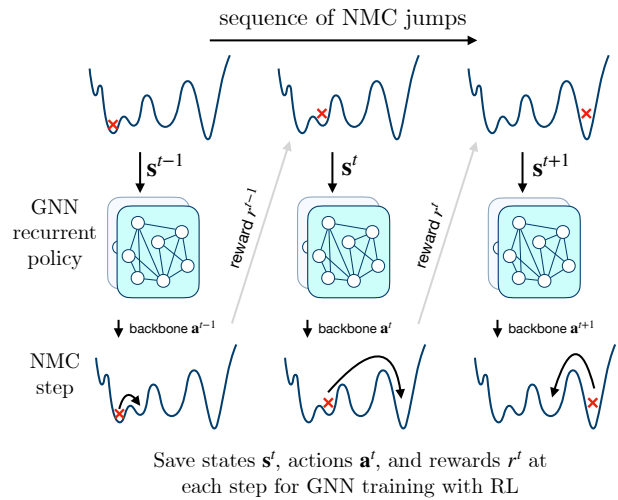}
    \caption{\label{fig:rlnmc}
        Reinforcement Learning Nonlocal Monte Carlo. 
        Each step $t\to t+1$ is given by the NMC jump of Alg.~\ref{alg:nmc}.
        RL policy infers backbones and is specified in Sec.~\ref{sec:rlnmc_details}.
        When needed for training, states, actions, and rewards are collected from multiple 
        instances running in parallel as specified in Alg.~\ref{alg:rlnmc_training}.
    }
\end{figure}

The RLNMC outline is shown in Fig.~\ref{fig:rlnmc}. 
In a local minimum state $\mathbf{s}^t$, a trainable RL policy $\pi_\theta$ with weights $\theta$ makes a prediction (action $\mathbf{a}^t$) 
which variables belong to the backbones that are consequently subject to the NMC jump, i.e. stochastic 
transition $\mathbf{s}^t \to \mathbf{s}^{t+1}$. When states $\mathbf{s}^{t}$ and $\mathbf{s}^{t+1}$ are compared, a reward $r^t$ is issued.
The states, actions, and rewards are collected and used by a RL algorithm of choice to adjust the policy $\pi_\theta$
either during a pre-training phase, or at runtime, or both.

In this work, NMC (which is based on SA itself) is taken as a base algorithm 
in which we substitute the phenomenological thresholding heuristic
of growing backbone clusters with the $\pi_\theta$ policy. The resulting RLNMC algorithm, 
is thus a direct extension of the SA and NMC methods.

The RL policy $\pi_\theta$ architecture is described in detail in Sec.~\ref{sec:rlnmc_details}, with the
sketch of modules in Fig.~\ref{fig:policy_arch}. There are three main components of $\pi_\theta$:
factor graph (hypergraph) message passing, per-variable memory, and a global memory. 
The message passing module is a Graph Neural Network (GNN) defined on the problem factor graph 
and is expected to learn the local state embeddings capturing hidden structures important for nonlocal moves.
The global memory is responsible for the creation of backbone schedules and global graph embeddings.
Finally, the per-variable memory may process basin-to-basin changes and correlations.
For example, if a variable is constantly in the same spin state, while the position in the landscape changes, 
such variable potentially should not be a part of the cluster move \cite{karimi2017}.
In the opposite scenario, if a variable is in different spin states, but its magnetization is always strong, 
it may be a part of the frustrated cluster, which needs to be relaxed to escape the local minimum. 

As a result, the RL policy $\pi_\theta$ receives the following per-variable input at time step $t$: 
the local minimum binary states of all variables $x^t_i$ (or $\sigma^t_i$ if spins are used), all absolute local fields $|H^t_i|$, 
and the recurrent (GRU) memory vector $\mathbf{h}^t_i$.
In addition, we provide the global current SA temperature $\beta^t$, the best energy reached by the algorithm so far $e^t$, 
and the global (GRU) memory $\mathbf{h}^t$.
Thus, the RL state at time $t$ is: $\bm{s}^t \equiv \big([x^t_0, H^t_0], [x^t_1, H^t_1], \dots| e^t, \beta^t\big)$
and $(\mathbf{h}_0^t, \dots | \mathbf{h}^t)$. 
As output, $\pi_\theta$ generates Bernoulli probabilities $p^t_i$ of being a backbone 
for each individual variable $i$ in parallel, and the value function $v^t$ that predicts the RL reward-to-go
at time $t$: $R^t = \sum_{\tau = t}^T r^\tau$.

It is worth paying special attention to the definition of rewards $r^t = R(s^t, a^t, s^{t+1})$.
If an action $a^t$ is followed by good rewards,
such actions will be favored by the RL training algorithm. The goal of this paper is to discover 
nonlocal transitions capable of effectively escaping local minima \textit{and} consequently finding
configurations with better energy, balancing the cost of exploration and exploitation.
As a result, we find the following reward definition appropriate:
\[
r^t = 
\begin{cases}
0, & \text{if $E(\bm{s}^{t+1}) - e^t > 0$} \\
-\left[E(\bm{s}^{t+1}) - e^t\right], & \text{else},
\label{eq:rl_reward}
\end{cases}
\]
where $e^t$ is the best energy seen so far in the RL episode.
This reward encourages global improvement of the energy and is not as restrictive as the 
simpler definition $r^t = -\left[E(\bm{s}^{t+1}) - E(\bm{s}^{t})\right]$ which could immediately penalize 
actions raising the energy, even if future transitions lead to its reduction \cite{barrett2020, barrett2022, tonshoff2023}.
Another possible option is to employ the energy reward that is only issued at the end of annealing \cite{mills2020}. Such definition 
would be the most liberal with respect to the intermediate nonlocal move excitations; however, we found it to be too sparse leading 
to extremely long training times. For simplicity, we use the exact same temperature and NMC hyperparameters of SA/NMC when 
training/testing RLNMC (details in Sec.~\ref{sec:rlnmc_details}).

\subsection{\label{sec:numerical_simulations}Numerical simulations}
In this section we investigate the performance of SA, NMC, and RLNMC with respect to  
the three benchmarking metrics of interest: time-to-solution (Eq.~\ref{eq:tts}), residual energy, and diversity of solutions.
The respective details on metric estimations are given in Secs.~\ref{app:time_to_solution}, \ref{app:residual_energy}, 
\ref{app:diversity_of_solutions}.
\subsubsection{\label{sec:time_to_solution}Time-to-solution}
\begin{figure}[ht]
    \centering
    \includegraphics[width=0.95\linewidth]{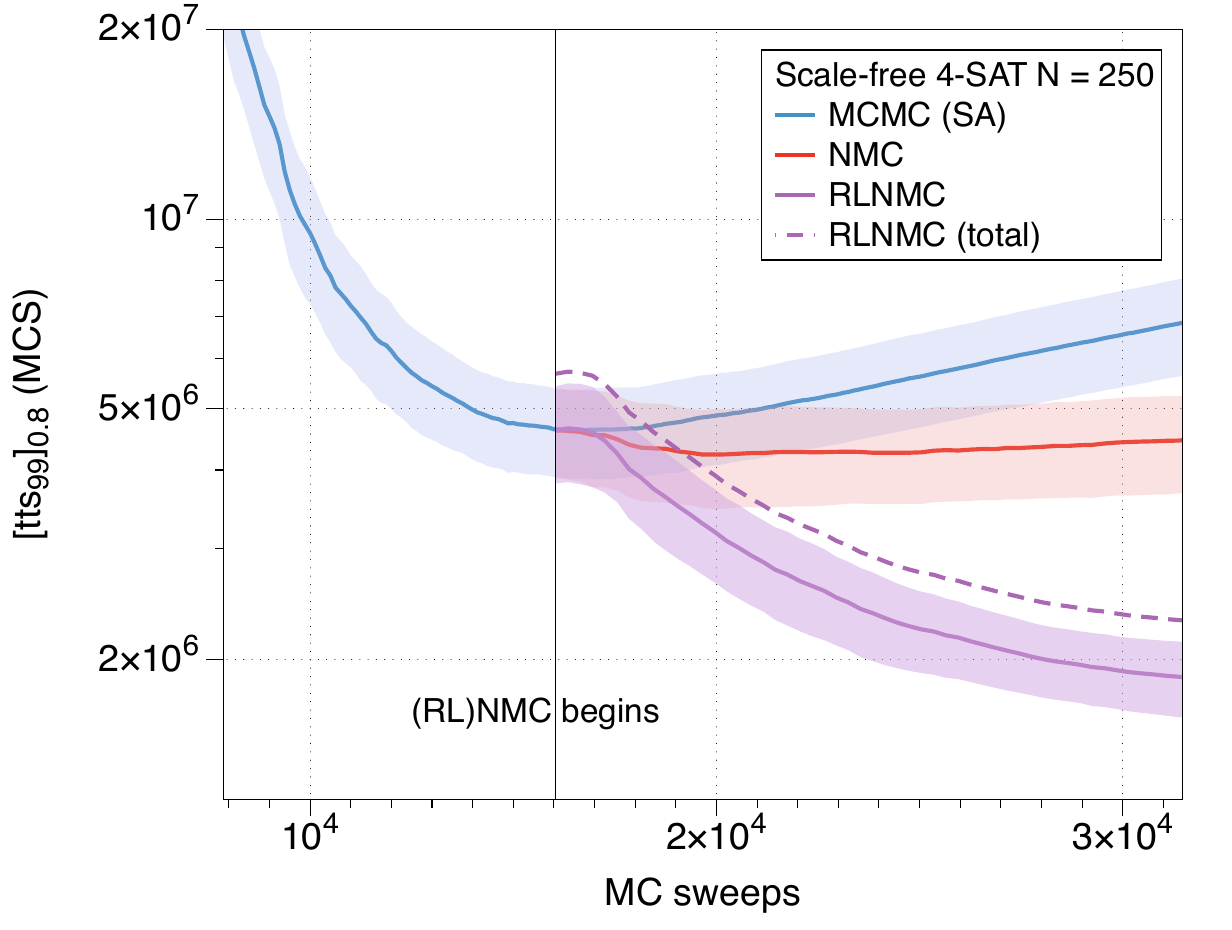}
    \caption{\label{fig:sf250_80_tts} 
        Time-to-solution for the hardest instances of the scale-free 4-SAT ($80$ percentile average and standard deviation)
        as a function of Monte Carlo sweeps in a single run. When Simulated Annealing (SA) saturates,
        NMC and RLNMC improve when increasing runtime. \textit{``RLNMC (total)''} takes into account 
        estimated computational cost of the neural network policy.
    }
\end{figure}
Fig.~\ref{fig:sf250_80_tts} shows the TTS$_{99}$ curve of SA/NMC/RLNMC for the $80$ percentile of the 
scale-free instances (see App.~\ref{app:4sat_sf} for instance description), measured in MC sweeps, as a function of the individual replica runtime. 
NMC/RLNMC algorithms begin at $\beta_{\rm{NMC}} = 5 \in (\beta_i = 2, \beta_f = 8)$ and follow the schedule of SA until $\beta_f$.
This figure illustrates the slow-down of SA, when the minimum of the TTS curve is reached at $\beta_{\rm{NMC}}$.
In comparison to the clear ``freezing'' of SA, NMC plot is almost flat until the end of the run indicating the successful exploration of 
the configuration space: POS in Eq.~\ref{eq:tts} increases quickly enough, justifying the extended runtime.
Beyond the observed advantage in terms of TTS, in later sections we will show that other metrics, such as diversity in 
Sec.~\ref{sec:diversity_of_solutions}, can be improved much more when exploration is effective.

In contrast to NMC, RLNMC significantly reduces TTS$_{99}$ even taking into account the computational cost 
of the relatively heavy recurrent RL policy (see App.~\ref{app:cost_at_test_time}): $\approx\!60\%$ improvement in MC sweeps and $\approx\!50\%$ in real runtime over SA. 
The NMC thresholding heuristic of the backbone inference is clearly not  
the optimal one with the chosen nonlocal move hyperparameters for the scale-free problem class and the RL policy has discovered a new strategy for optimization. 
We emphasize that all the other hyperparameters of RLNMC (except for the backbone inference policy) are identical to those of 
NMC and were not optimized, leaving room for an even further improvement.

\begin{figure}[ht]
    \centering
    \subfloat[\label{fig:sf250_50_energy}]{
        \includegraphics[width=0.9\linewidth]{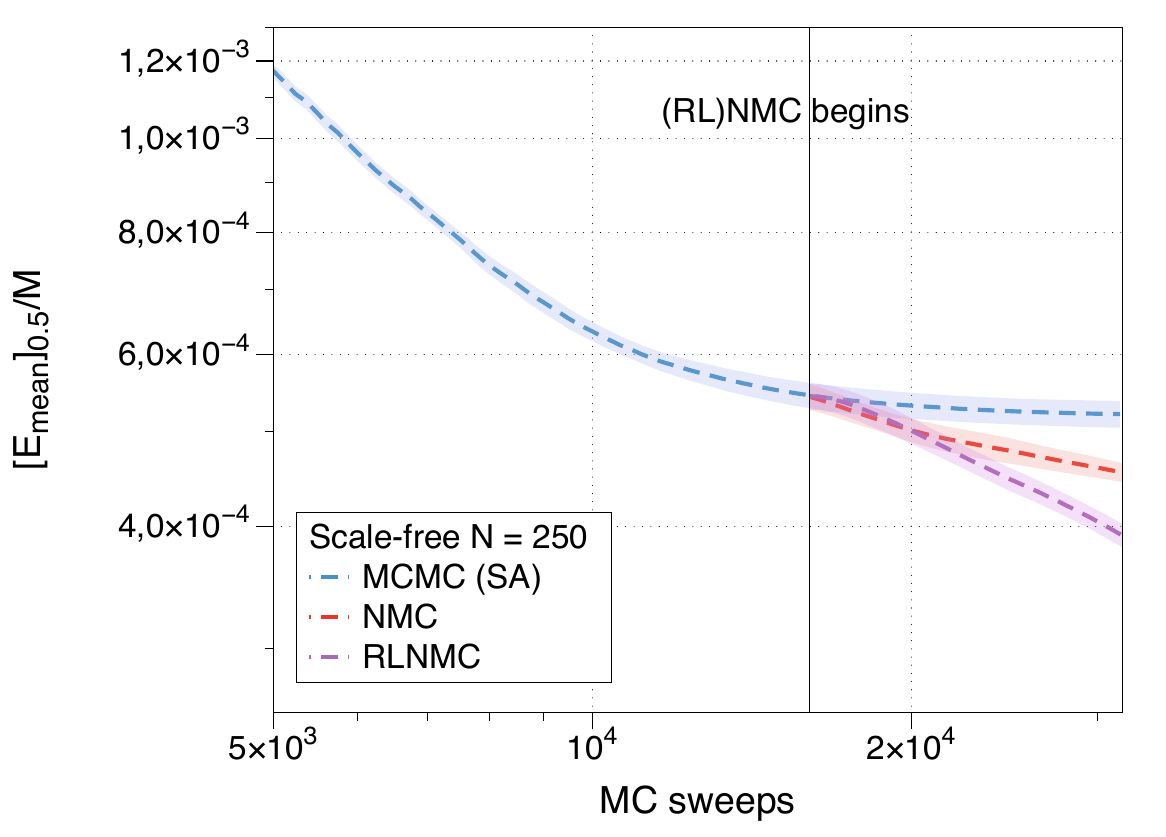}
    }
    \hfill
    \subfloat[\label{fig:uf500_50_energy}]{
        \includegraphics[width=0.9\linewidth]{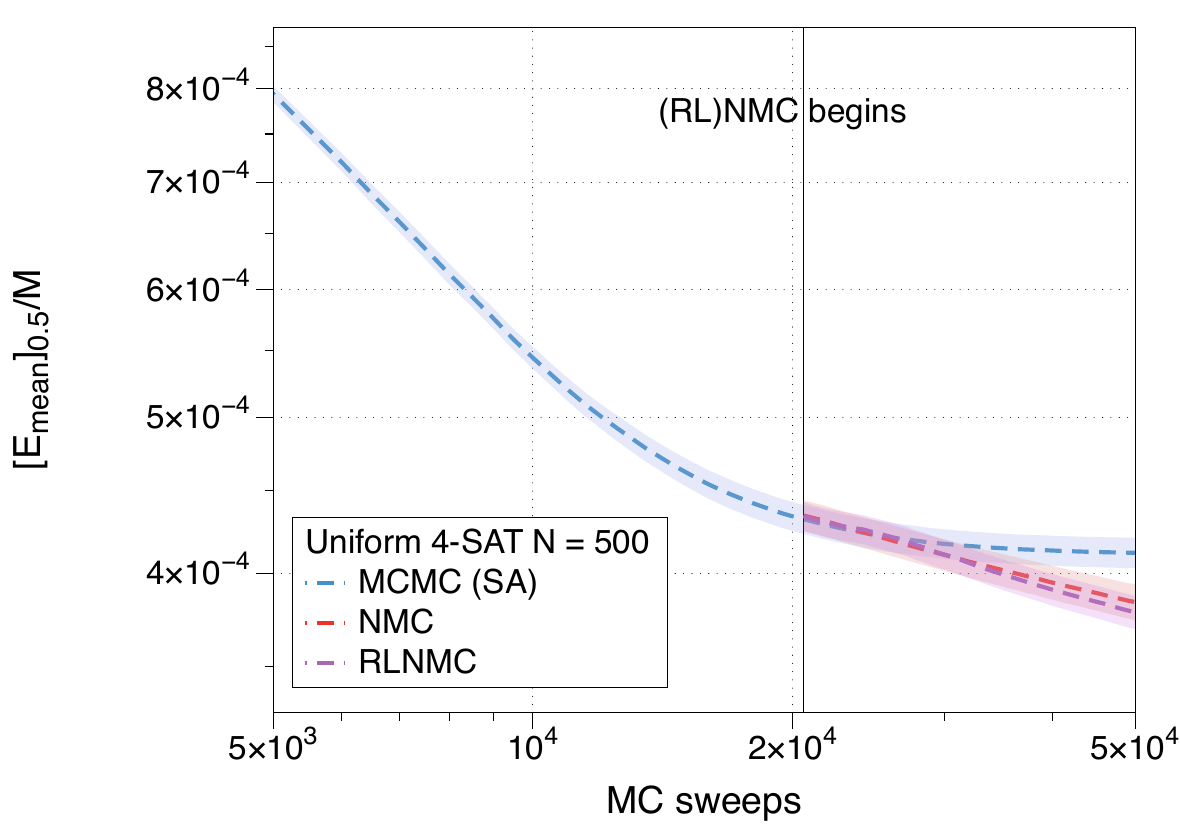}
    }
    \caption{\label{fig:50_energy}
        Median residual energy for (a) scale-free random 4-SAT, 
        (b) uniform random 4-SAT. NMC and RLNMC nonlocal moves begin at the indicated step.  
        For each instance the mean is over 4096 and 2048 replicas respectively.
        The median and its standard deviation are estimated with bootstrap resampling 
        of 320 used instances.
    }
\end{figure}

\subsubsection{\label{sec:residual_energy}Residual energy}
To complement the results in Fig.~\ref{fig:sf250_80_tts} we also show the the average energy 
(number of unsatisfied clauses) for the scale free problems in Fig.~\ref{fig:sf250_50_energy} as a function 
of MC sweeps on the log-log scale. For SA, there are seemingly two phases with distinct slopes of the $E$ vs MC sweeps curve:
an initial steep stage, and the second ``frozen'' stage. We observe that the steeper curve 
of energy reduction is recovered by RLNMC, while NMC is not able to match~it. 

To address the scaling and generalization of RLNMC, we further investigate its performance on the uniform random 4-SAT benchmark 
of $N = 500$, $1000$, and $2000$ at the $M/N = 9.884$ clause-to-variable ratio. 
First, we hyperparameter optimized SA/NMC and trained RLNMC on the $N = 500$ problem size.
At this size, the resulting energy vs MC sweeps benchmarking data is shown in Fig.~\ref{fig:uf500_50_energy}. 
NMC shows a better slope than SA even with the simple nonlocal move heuristic employed in this work (compared to 
the more advanced original version in \cite{mohseni2021}). In turn, RLNMC slightly outperforms NMC at the same number 
of MC sweeps and matches NMC when the policy overhead is taken into account (see App.~\ref{app:cost_at_test_time}).
The bigger advantage of RLNMC over NMC on the scale-free problem class compared to the uniform random could be explained by the more accessible structure of the scale free 4-SAT to the RL algorithm.

\begin{figure}[ht]
    \centering
    \includegraphics[width=0.95\linewidth]{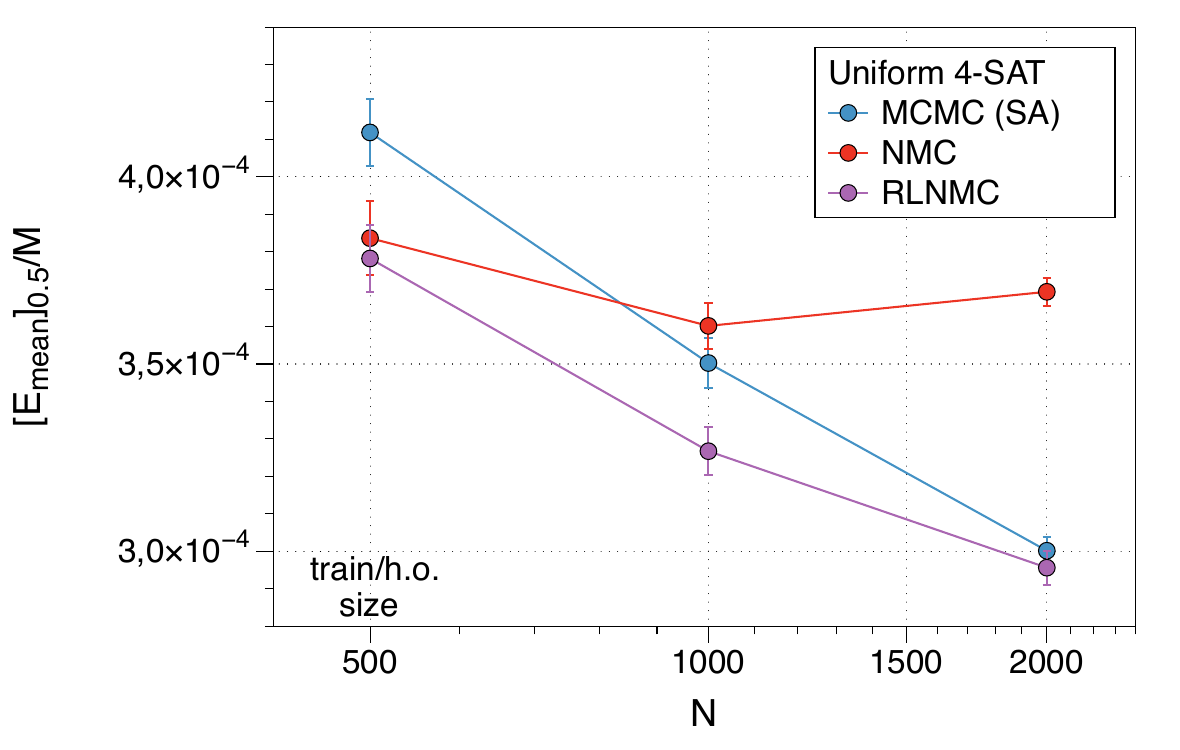}
    \caption{\label{fig:energy_scaling}
        Comparison of heuristics at larger sizes. Hyperparameter optimization (h.o.)/RL training
        is only at $N = 500$. MCMC and RLNMC generalize well, while NMC requires further h.o. 
        Normalized median residual energy for uniform random 4-SAT from 320 instances at each size $N$, $M/N = 9.884$. 
        The $N=500$ data is from Fig.~\ref{fig:uf500_50_energy}. Plots of energy vs MC sweeps for 
        $N = 1000,\,2000$ are shown in Fig.~\ref{fig:uf_50_energy_extra}.
    }
\end{figure}

However, this hierarchy is different when we consider scaling to larger problem sizes. 
The only algorithmic change we make is the increase of the total number of sweeps for every replica from $5\times 10^4$ at 
$N = 500$ to $10^5$ ($2\times 10^5$) at $N = 1000$ ($N = 2000$), i.e. we use $N^2$ runtime scaling 
which is not subject to the current OGP theory of algorithmic limits \cite{angelini2025}.
The number of NMC nonlocal moves (i.e. the number of NMC/RLNMC policy calls) is not changed
but the inter-jump number of sweeps is proportionally scaled.
In Fig.~\ref{fig:energy_scaling} we show average energies normalized by the number of clauses
for SA/NMC/RLNMC at $N = 500,\,1000,\,2000$. When increasing the size, NMC struggles to generalize without additional 
hyperparameter tuning. In contrast, RLNMC continues to perform well and outperforms SA even at $4\times$ problem size
compared to the one it was trained on. 
Furthermore, we have chosen not to scale the number of nonlocal moves with $N$
which greatly reduced the contribution of the policy inference runtime compared to MC sweeps
(see Fig.~\ref{fig:RLNMC_overhead}). There is clearly room for further improvement with additional hyperparameter 
optimization and RL training. Here we wanted to show that the simplest version of nonlocal moves leads to substantial improvement in larger sizes without significant overhead in hyperparameter optimization.

\subsubsection{\label{sec:diversity_of_solutions}Diversity of solutions}
Figs.~\ref{fig:sf250_80_tts} and \ref{fig:sf250_50_energy} 
have shown that we are able to reduce time-to-solution and average energy across 
replicas metrics when improving SA with nonlocal moves (NMC) and reinforcement learning (RLNMC). 
In principle, this can be achieved with either (a) reliably getting the same states within 
the accepted approximation ratio across independent replicas, 
or (b) with finding different solutions that are not necessarily close to each other in the configuration space.
In order to make a case for the latter, here we estimate diversity $D$ of solutions \cite{mohseni2023} reached by the 
SA/NMC/RLNMC solvers.

\begin{figure}[ht]
    \centering
    \includegraphics[width=0.9\linewidth]{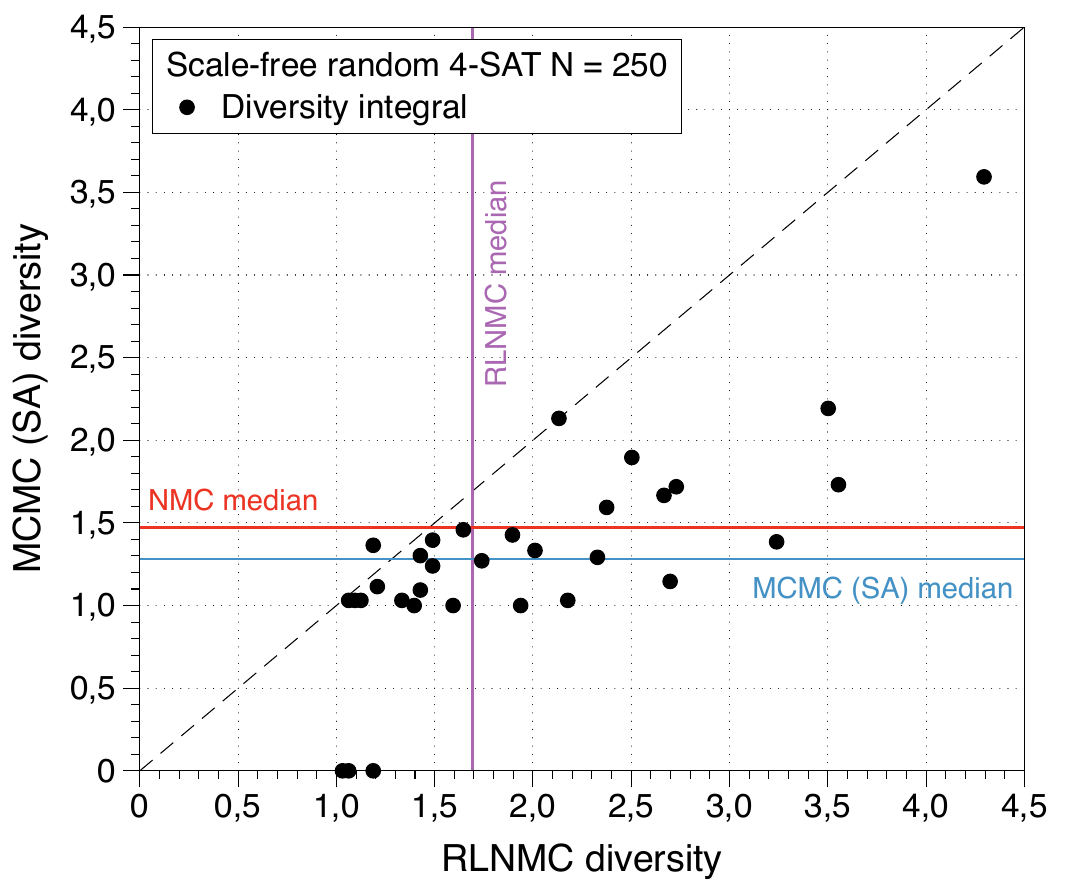}
    \caption{\label{fig:diversity}
        Diversity of configurations for scale-free random 4-SAT at $E = 0$ (solutions). 
        The scatter plot of SA and RLNMC for top $10\%$ hardest instances ($32$ in total); 
        NMC per-instance diversity data is represented by its median.
    }
\end{figure}

To measure diversity of solutions, we compute the following integral:
\begin{equation}
    D \equiv \int_{R_{\rm{min}}}^{R_{\rm{max}}}\frac{D(R)dR}{R_{\rm{max}}-R_{\rm{min}}}\,,
    \label{eq:diversity_definition}
\end{equation}
where $D(R)$ is defined as the Maximum Independent Set (MIS) of the undirected 
graph constructed from the set of solutions. Every solution configuration $\bm{\sigma}_i$ (or $\bm{x}_i$) corresponds 
to a node and every edge $e_{ij}$  connects the nodes if their Hamming distance 
$d(\bm{\sigma}_i, \bm{\sigma}_j)/N$ is less than $R$.
As the name implies, the larger the diversity number is, the more distinct the solutions in independent replicas 
are. The parameter $R$ determines how far in Hamming distance the configurations have to be from each other 
in order to be considered distinct. If $R = 0$, then $D(R = 0)$ equals to the total number of solutions found 
(the graph is not connected); in contrast, if $R = 1$, then $D(R = 1) = 1$ (MIS of the fully connected graph), 
i.\,e. all solutions are considered similar. If no solutions are found at all, then $D \equiv 0$ for any $R$.
In Eq.~\ref{eq:diversity_definition} we used the values $R_{\rm{min}} = 0.02$ and $R_{\rm{max}} = 0.5$, the justification for which is discussed in App.~\ref{app:diversity_of_solutions}

In Fig.~\ref{fig:diversity} we show the diversity results for the hardest $10\%$ instances from the 
scale-free 4-SAT benchmark. RLNMC features a strong advantage in $D$ compared to SA and NMC. 
In particular, there are several instances which feature zero solutions found by SA,
where RLNMC found more than one solution. Also in the case when SA found only one solution ($D \approx 1$), RLNMC 
managed to find several times more (up to $3\times$), without explicitly being trained for the diversity metric. Only for one instance out of $32$ we found SA to slightly outperform RLNMC. When comparing the medians, RLNMC gives $\approx\,32\%$ advantage over SA and $\approx\,15\%$ advantage over NMC.

As a result, we have shown that, when combined with NMC, RL is capable of discovering nonlocal move strategies 
by exploring the energy landscape and observing the energy improvement rewards. The resulting solver 
is faster than SA/NMC, shows generalization and good scaling, and demonstrates improved diversity of solutions.

\subsubsection{RLNMC policy features\label{sec:rlnmc_policy}}
\begin{figure}[ht]
    \centering
    \includegraphics[width=\linewidth]{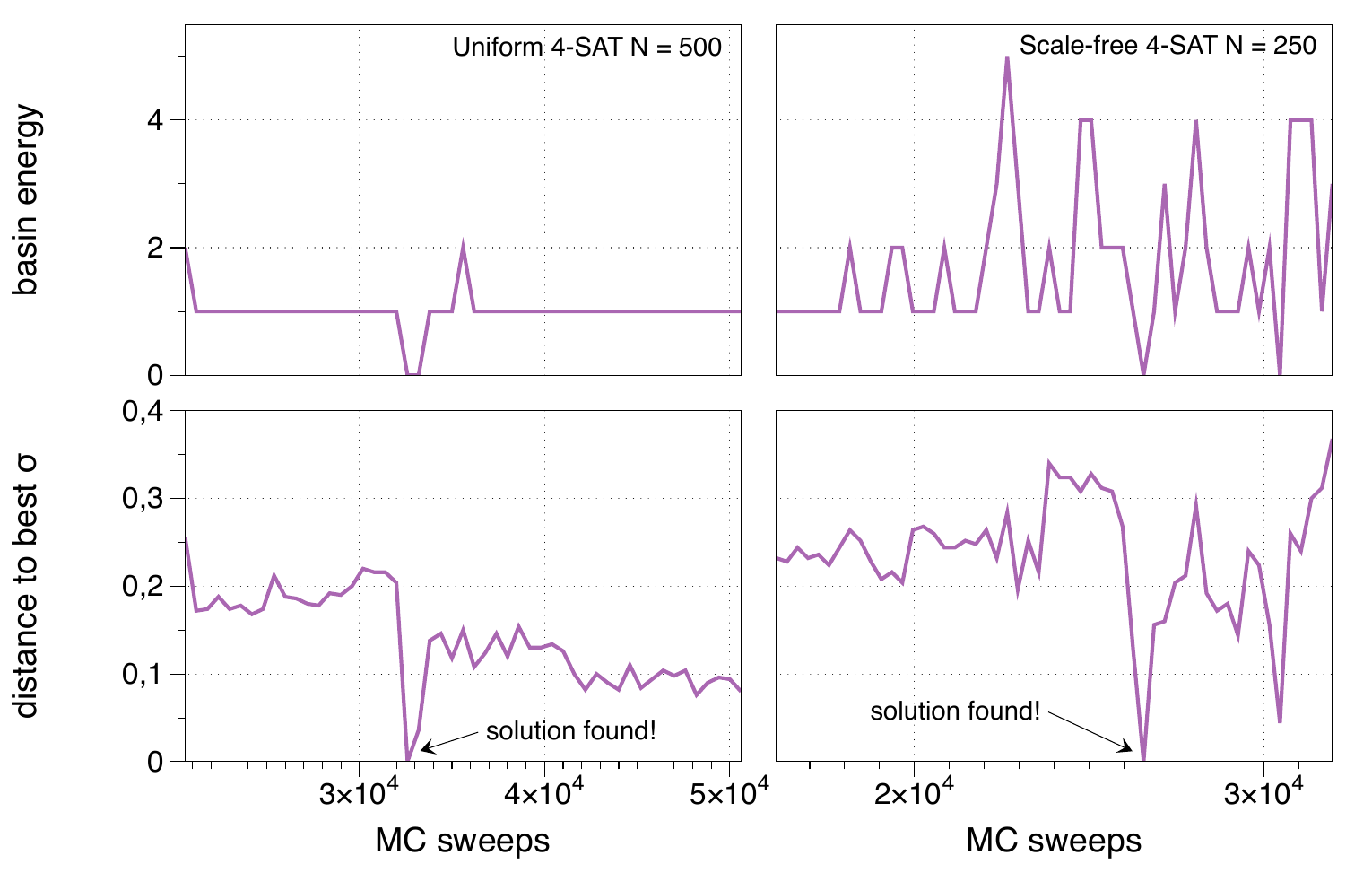}
    \caption{\label{fig:dtobest}
        Energy landscape exploration examples by trained RLNMC policies for (left) uniform random 
        and (right) scale-free random 4-SAT problem instances; (top) trajectory of the minimum 
        energies of basins of attractions; (bottom) distance to the state 
        with the best seen energy so far.
    }
\end{figure}

We would like to gain insights into the features of the trained RLNMC policies of this paper. 
Fig.~\ref{fig:dtobest} shows examples of the energy landscape trajectory for the uniform random $N = 500$
(scale-free random $N = 250$) problems. The basin energy is defined as the minimum energy within every $600$ ($300$) 
MC sweeps. The Hamming ``distance to best $\sigma$'' is defined as the distance (in other words, overlap) of the best state 
of the current basin (i.e. the state of the basin energy) to the best energy state seen so 
far in the configuration space trajectory. This allows us to track when the necessary approximation was found for 
the first time and how far from it the solver typically travels in the energy landscape. 
We observe that in the scale-free problem RLNMC traverses larger distances compared to the uniform problem, 
as well as reaches higher basin energies before converging to a solution.

\begin{figure}[ht]
    \centering
    \subfloat[\label{fig:dtobest_uf}]{
        \includegraphics[width=0.485\linewidth]{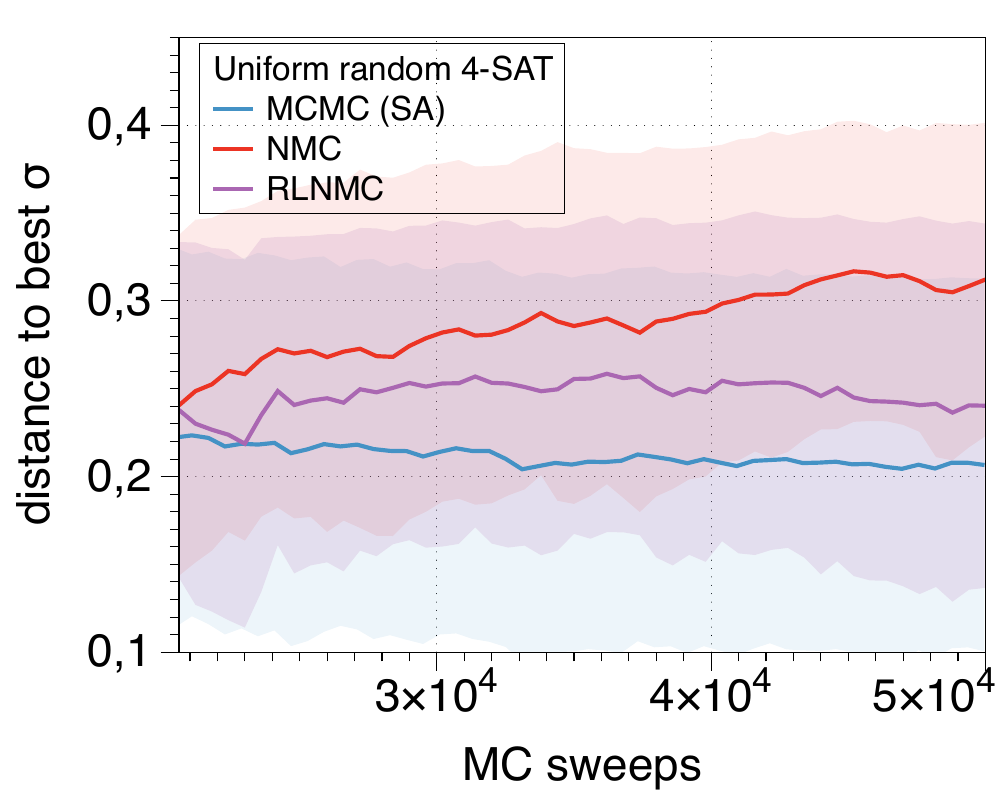}
    }
    \subfloat[\label{fig:dtobest_sf}]{
        \includegraphics[width=0.485\linewidth]{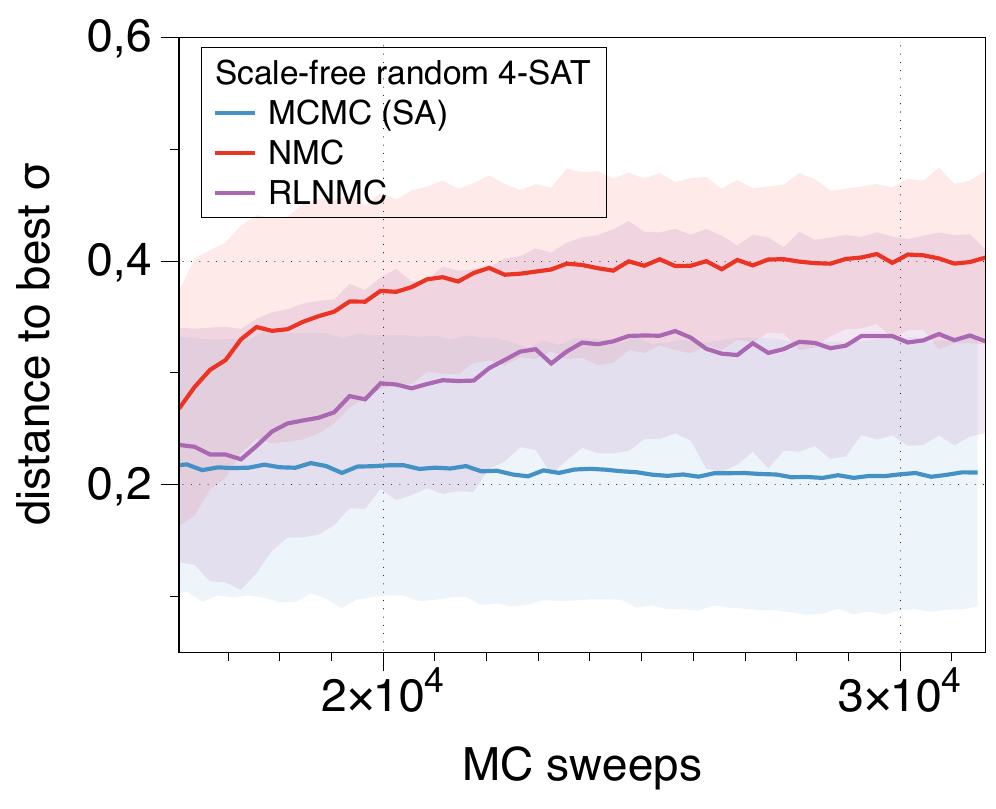}
    }
    \caption{\label{fig:dtobest_avg}
        Distance to the state with the best seen energy so far averaged over multiple (128) replicas 
        for uniform and scale-free instances from Fig.~\ref{fig:dtobest}; means and standard deviations are shown.
    }
\end{figure}

In Fig.~\ref{fig:dtobest_avg} we show again  the ``distance to best'' plot, but averaged over multiple independent
replicas and for all SA/NMC/RLNMC algorithms. We see that RLNMC operates in an intermediate regime between SA and NMC in both problem classes. 
This observation held for all instances we tested. Both NMC and RLNMC show an adaptation to scale-free problems compared to uniform, to make larger distance moves. 
Yet, considering the advantage of RLNMC over NMC in terms of energy, time-to-solution, and diversity metrics, 
the nonlocal strategy of NMC to excite mostly the highly magnetized spins could be too strong and modifications
to this method could be explored to learn from the RLNMC.
Furthermore, schedules of NMC can be learned and successfully employed.  
To support this, in supplementary Sec.~\ref{app:nmc_schedules} we show the 
schedules of the nonlocal moves created by RLNMC of this paper.

\section{\label{sec:conclusion}Outlook}
Here we give an outlook of future improvements for RLNMC and its potential 
for the optimization and sampling applications.

We have used simulated annealing (SA) as a baseline heuristic that was augmented with nonlocal 
moves. In principle, other base algorithm could be augmented 
with RL and NMC. For example, one could take WalkSAT \cite{biere2009}, the heuristic local search method designed for SAT problems, 
and introduce inhomogeneous nonlocal cluster moves for the variables. Another straightforward integration 
of RLNMC can be implemented for parallel tempering (PT) \cite{mohseni2021}, where low-$T$ 
replicas are typically subject to low acceptance rates of sampling. 

In the current study, the hyperparameters of the algorithms were optimized sequentially; first, the SA schedule was adjusted; 
second, NMC moves for NMC were optimized using $T$ schedule of SA; finally, RLNMC was trained using 
the NMC sweeps setting of NMC. This potentially handicapped NMC and RLNMC as these algorithms may require their own 
optimized hyperparameters controlling the NMC sweeps. Joint training of the nonlocal moves and other degrees
of freedom is a future opportunity. For example, the SA schedule itself could be trained 
with RL, as shown in \cite{mills2020}, which could be combined with RLNMC learning of nonlocal moves.

A promising direction is for RLNMC to continue training online to adjust cluster moves in an instance-wise fashion. 
Provided that the costs of online training do not outweigh the benefits, this could make the algorithm adapt 
not only to the problem class, but to individual instances. As a next level of complexity, one could use meta reinforcement learning 
to deliberately pre-train the model to be instance-wise adaptive. 

NMC/RLNMC can be applied to inverse problems. For example, (RL)NMC could be employed for the training of energy-based models
\cite{huembeli2022}, such as Boltzmann machines \cite{niazi2024training}. Furthermore, algorithms introduced here can 
also be incorporated within probabilistic computing paradigm realized with FPGA, custom-design ASIC, or nano-devices
\cite{chowdhury2023fullstack,aadit2023accelerating}. Such probabilistic computing approach could challenge
performance of standalone quantum optimizers \cite{chowdhury2025}.
The NMC family of algorithms can be eventually embedded within Probabilistic Processing Units (PPU) in upcoming 
hybrid high-performance quantum-classical infrastructures. These heterogeneous systems could constitute future quantum 
supercomputers powered by CPUs, GPUs, PPUs, and Quantum Processing Units (QPUs) employing an optimal interplay of 
classical and quantum fluctuations \cite{mohseni2024}.

\section{\label{app:methods}Methods}
\subsection{\label{app:baselines_ho}Baselines description and hyperparameters}
We use a set of $384$ instances for hyperparameter optimization 
and benchmarking of every problem class (uniform random and scale-free random). 
$64$ instances out of $384$ are used for hyperparameter tuning (SA, NMC) and 
training (RLNMC); the remaining $320$ are employed for benchmarking, i.e. results 
reported in this paper.
\subsubsection{\label{app:sa}Simulated Annealing (MCMC SA)}

We limit the total number of MC sweeps $N_{\rm{sw}}$ that SA can run for and optimize 
the initial $\beta_i$ and final $\beta_f$ temperatures to get close to optimal (within the error bars) 
performance of the median $\mathrm{TTS}_{99}$ across the $64$ instances used for hyperparameter optimization.
As it is difficult for the uniform random problem to reach exactly $0$ satisfied clauses, we defined 
success for this problem class when approximation ratio of $2\times 10^{-4}$ was reached (1 or less unsatisfied clauses).
The schedule of $\beta$ is linear (not the linear schedule of $T$). 
Furthermore, we choose to increase the allowed $N_{\rm{sw}}$ values when increasing 
the problem sizes of the benchmarks of this study. As discussed in \cite{angelini2025}, 
the overlap-gap-property is currently not explored for superlinear algorithms, i.e. the case when 
the number of sweeps increases with growing $N$. In principle, such algorithms are ``unstable'' and therefore 
less prone to algorithmic barriers for ``stable'' OGP algorithms, 
the reasoning that aligns with the purposes of this study.

Simulated Annealing (SA) with the linear schedule of $\beta = 1/T$
starts at a hyperparameter optimized small value $\beta_i$ (large $T_i$) and finishes at a large value $\beta_f = 8$ (small $T_f$) that gave 
a small acceptance rate of $\approx 10\%$ for the scale-free random class, and $\approx 5\%$ 
for the uniform random class. The temperature is changed at every sweep with a 
step $\Delta \beta = (\beta_f - \beta_i)/N_{\rm{sw.\,total}}$, where $N_{\rm{sw.\,total}}$ 
is the total number of sweeps allowed for one SA run. The $N_{\rm{sw.\,total}} \approx 3\times 10^4$
for the scale-free problem at $N = 250$, and $N_{\rm{sw.\,total}} = 5\times 10^4,\, 10^5,\,2\times 10^5$
for the uniform random problem at $N = 500,\,1000,\,2000$ variables respectively ($N^2$ scaling of
runtime). The resulting values are $\beta_i = 3$ for 
the uniform random class, and  $\beta_i = 2$ for the scale free class. These relatively small optimized values
of temperature $1/\beta_i$ can be explained by the flatness of the energy landscape which often requires 
exploration even without significant excitations, i.e. features entropic barriers \cite{dobrynin2024}. 
We found that the energy curves of SA begin to approximately saturate
when $\beta_{\rm{NMC}} = 5$ (see Fig.~\ref{fig:50_energy}), which indicates the slow-down of the energy landscape exploration. 
As a result, we aim to improve these suboptimal solutions by calling the nonlocal moves of NMC and RLNMC described below.

\subsubsection{\label{app:NMC}Nonlocal Monte Carlo Simulated Annealing (NMC)}
The Nonlocal Nonequilibrium Monte Carlo (NMC) moves suggested in \cite{mohseni2021} utilize correlations 
$\tilde{J}_{ij\dots} \equiv \mathrm{atanh} |\langle s_i s_j\dots \rangle|/\beta$ and/or magnetizations 
$\tilde{h}_i \equiv \mathrm{atanh}|\langle s_i \rangle|/\beta$ of variables to construct the 
``backbones'' of basins of attraction. The values of $\tilde{J}_{ij\dots}$ and $\tilde{h}_i$ 
are estimated using Loopy Belief Propagation (LBP) on the surrogate (localized) Hamiltonian, 
$H^* = H - \lambda\sum_{i = 1}^N{\sigma}^*_i\sigma_i$, ensuring 
that sampling is carried out in a specific basin, as well as convergence and efficiency. 
Given that the main focus of this paper is testing reinforcement learning ability to learn nonlocal 
transitions having the simplest information about the problem available, we do not use 
the surrogate method of \cite{mohseni2021}. Instead, magnetizations are approximated by the local fields 
$|H_i| = |\sum_{i_2 < \dots < i_p}^N J_{i, i_2, \dots, i_{p}}s_{i_2}\dots s_{i_p} + \dots + h_i|$, which is computationally simple
and provides a sufficient signal for the purposes of this study. However, our implementation \cite{rlnmc_code} does support
estimation of localized correlations with LBP, and its performance combined with RL is an interesting direction for future work.

\begin{algorithm}[H]
    \caption{Nonlocal Monte Carlo (NMC) single step $\bm{t} \to \bm{t+1}$} 
    \label{alg:nmc}
    \begin{algorithmic}[1]
    \renewcommand{\algorithmicrequire}{\textbf{Input:}}
    \renewcommand{\algorithmicensure}{\textbf{Output:}}
    \REQUIRE Local minimum state $\bm{\sigma^t}$, backbones $\{b\}$, $N_{\rm{sw}}$ MC sweeps, $N_{\rm{cycles}}$ hyperparameters
    \ENSURE New state $\bm{\sigma^{t+1}}$ \\
    Set minimum energy state $\bm{\sigma^t}_{\rm{min}} = \bm{\sigma^t}$\\
    \FOR{cycle in $N_{\rm{cycles}}$}
        \STATE \textbf{[Backbone MC stage]:} Randomize $\{b\}$ variables (or sample at $T_{\rm{high}}$ for the nonequilibrium version) with non-backbones $\{nb\}$ fixed:
        $\bm{\sigma^{t}} = [\sigma^t_{i \in \{b\}}, \sigma^t_{i \in \{nb\}}] \to [\sigma^{t+1}_{i \in \{b\}}, \sigma^t_{i \in \{nb\}}] = \bm{\sigma^{t+1}_b}$ \\
        
        \STATE \textbf{[Non-backbone MC stage]:} Monte Carlo sweep over $\{nb\}$ variables at $T_{\rm{low}}$ with backbones $\{b\}$ fixed:
        $\bm{\sigma^{t+1}_b} = [\sigma^{t+1}_{i \in \{b\}}, \sigma^t_{i \in \{nb\}}] \to [\sigma^{t+1}_{i \in \{b\}}, \sigma^{t+1}_{i \in \{nb\}}] = \bm{\sigma^{t+1}_{nb}}$ \\
        
        \STATE \textbf{[Full MC stage]:} $N_{\rm{sw}}-2$ Monte Carlo sweeps over \textit{all} variables at $T_{\rm{low}}$:
        $\bm{\sigma^{t+1}_{nb}} \to \bm{\sigma^{t+1}_{all}}$ \\
        \STATE If $E(\bm{\sigma^{t+1}_{all}}) < E(\bm{\sigma^t}_{\rm{min}})$, then $\bm{\sigma^{t+1}} = \bm{\sigma^{t+1}_{all}}$
    \ENDFOR
    \RETURN $\bm{\sigma^{t+1}}$ [Total: $N_{\rm{sw}}$ MC sweeps]
    \end{algorithmic}
\end{algorithm}

We define a hyperparameter threshold $r$, which controls if a variable $i$ is a backbone subject to the nonlocal move 
by the simple inequality $|H_i| \ge r$. The nonlocal move for the backbone is defined as a randomization of 
spins (infinite $T$ excitation). After hyperparameter optimization we found that the best performing thresholds
are $r = 3$ and $r = 4.5$ for the uniform random and scale-free random problems respectively.
We run the standard SA from $\beta_i$ to $\beta_{\rm{NMC}} = 5$, but NMC from $\beta_{\rm{NMC}} = 5$ to $\beta_{f} = 8$ 
in both cases following approximately the same linear schedule of SA. 
For the scale-free (uniform) problems with $\beta_i = 2$ ($\beta_i = 3$) 
this means that SA is run for the initial $50\%$ ($40\%$) of the total runtime (in MC sweeps). For the remaining 
$50\%$ ($60\%$) of the runtime the NMC jumps are performed with the base (low) temperature decreasing
according to the SA schedule.

The NMC transitions consist of a nonlocal move and the consequent low-$T$ sampling
and are described in detail in Alg.~\ref{alg:nmc}. 
The used hyperparameters are: $N_{\rm{cycles}} = 3$; 
$N_{\rm{NMC\,steps}} = 53$, $N_{\rm{sw}} = 100$ for the scale-free problem at $N = 250$, 
and $N_{\rm{NMC\,steps}} = 50$, $N_{\rm{sw}} = 200,\, 400,\, 800$ 
for the uniform random problems at $N = 500,\,1000,\,2000$ problem sizes respectively.

\subsubsection{\label{sec:rlnmc_details}Reinforcement Learning Nonlocal Monte Carlo  (RLNMC)}
RLNMC is built on top of the MCMC SA/NMC algorithms of Sec.~\ref{app:sa} and Sec.~\ref{app:NMC}.
RLNMC substitutes the thresholding heuristic of NMC with a deep policy trained with RL.
We use the same $N_{\rm{sw}}$, $N_{\rm{cycles}}$ hyperparameters 
controlling the number of MCMC sweeps in the NMC nonlocal move algorithm, and the same 
$\beta_i$, $\beta_{\rm{NMC}}$, $\beta_f$ temperature values used in SA/NMC. 
RLNMC is trained on the same instances that were used for hyperparameter optimization of SA/NMC.

The Proximal Policy Optimization (PPO) training setup is described in supplementary Sec.~\ref{app:training_details}. 
In addition, in Sec.~\ref{app:cost_at_test_time} we discuss the computational cost of RLNMC compared to NMC.

\begin{figure*}[ht]
    \centering
    \includegraphics[width=0.99\linewidth]{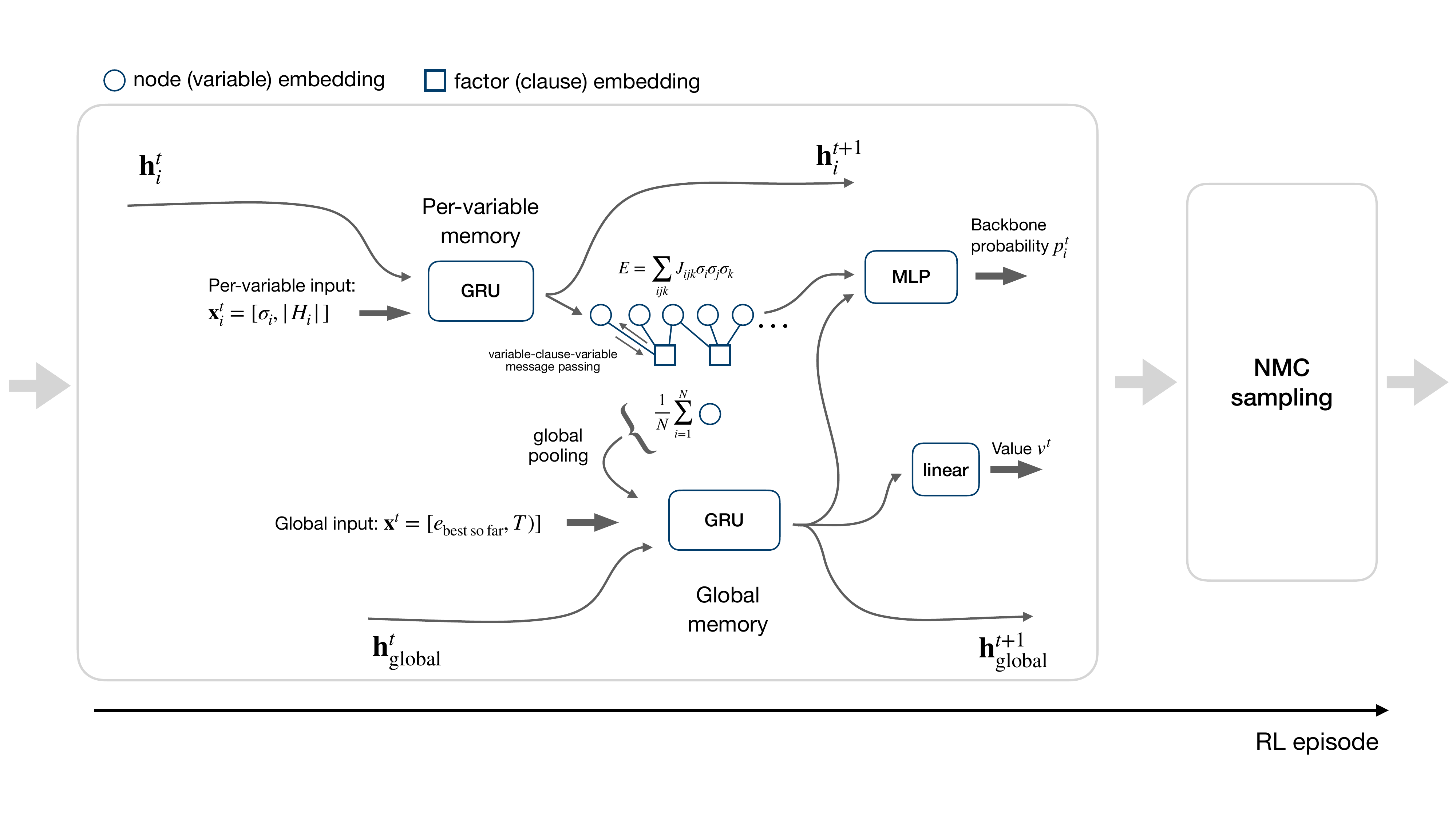}
    \caption{\label{fig:policy_arch} Recurrent policy architecture. It is incorporated into RLNMC as shown in Fig.~\ref{fig:rlnmc}.
        Exact description and hyperparameters of modules are explicitly given in App.~\ref{sec:rlnmc_details}.
    }
\end{figure*}

Here we specify the RLNMC recurrent deep neural network policy, with the architecture visualized in Fig.~\ref{fig:policy_arch}.
We would like to list all the sub-modules, explicitly stating their structure and the chosen hyperparameters.

\begin{itemize}
    \item \textbf{Input.} Local minimum state of every variable $s_i^t$, absolute local fields of variables $|H_i^t|$,
    current best energy seen so far in the episode 
    $e^t_i = E_{\rm{best\,so\,far}}/e_{\rm{scale}}(N)$, current temperature $T^t$.
    The energy scale is chosen to be $e_{\rm{scale}}(N) = N/50$ and used to keep the energy at the same order of magnitude 
    regardless of the problem size.
    \item \textbf{Local GRU.} The local per-variable Gated Recurrent Unit (GRU) memory at every NMC step takes as input its hidden state 
    $\mathbf{h}^t \in \mathbb{R}^{16}$, and the local information $\mathbf{x}^t = [s_i^t, |H_i^t|]$.
    \item \textbf{Factor graph self-attention.} Within each factor we perform self-attention \cite{vaswani2023} message passing 
    using the standard queries $\mathbf{q}_i^t = W^q\mathbf{h}_i^t$, keys $\mathbf{k}_i = W^k\mathbf{h}_i^t$, 
    and values $\mathbf{v}^t_i = W^v\mathbf{h}_i^t$, all $\mathbb{R}^{16}$. As a result, each factor $a$ yields 
    embeddings for variables $i\in \partial a$: $\mathbf{y}^t_{a, i\in \partial a} = \sum_{j\in a} \alpha_{ij}^t \mathbf{v}_{j}^t$, where 
    $\alpha_{ij}^t = \exp{\left(\mathbf{q}_i^t \cdot \mathbf{k}_j^t\right)}/\sum_{j\in \partial a}\exp{\left(\mathbf{q}_i^t \cdot \mathbf{k}_j^t\right)}$.
    \item \textbf{Node aggregation.} To get the node embedding, we average over the factors each node appears in:
    $\mathbf{y}_i = \frac{1}{|\partial i|}\sum_{a\in \partial i} \mathbf{y}^t_{a, i\in \partial a}$.
    \item \textbf{Global GRU.} The global GRU at every NMC step takes as input its hidden state 
    $\mathbf{h}^t \in \mathbb{R}^8$, the global information $\mathbf{x}^t = [e^t_i, T^t]$, and the result 
    of the variables' hidden state mean pooling $\frac{1}{N}\sum_{i = 1}^N \mathbf{y}_i$.
    \item \textbf{Variable Output.} An $24 \to 8 \to 1$ MLP takes as input the concatenated 
    variable embeddings $\mathbf{y}_i^t$ and global GRU state $\mathbf{h}^{t+1}$ and outputs 
    the Bernoulli backbone probability $p_i \in [0, 1]$ which constitutes the stochastic action of the RL policy 
    $\pi_{\theta}$.
    \item \textbf{Global Value Output.} Finally, the PPO value $v^t$ is obtained using the linear $8\to 1$ layer 
    with all other modules shared with $\pi_{\theta}$.
\end{itemize}

\section*{Data availability}
The data that supports the findings of this study is available from 
the corresponding author upon reasonable request.

\section*{Code availability}
The SA, NMC, and RLNMC code for benchmarking and training, as well as the trained models for which the results are reported are available in the repository
\cite{rlnmc_code}.

\bibliography{rlnmc}

\begin{thebibliography}{104}%
\makeatletter
\providecommand \@ifxundefined [1]{%
 \@ifx{#1\undefined}
}%
\providecommand \@ifnum [1]{%
 \ifnum #1\expandafter \@firstoftwo
 \else \expandafter \@secondoftwo
 \fi
}%
\providecommand \@ifx [1]{%
 \ifx #1\expandafter \@firstoftwo
 \else \expandafter \@secondoftwo
 \fi
}%
\providecommand \natexlab [1]{#1}%
\providecommand \enquote  [1]{``#1''}%
\providecommand \bibnamefont  [1]{#1}%
\providecommand \bibfnamefont [1]{#1}%
\providecommand \citenamefont [1]{#1}%
\providecommand \href@noop [0]{\@secondoftwo}%
\providecommand \href [0]{\begingroup \@sanitize@url \@href}%
\providecommand \@href[1]{\@@startlink{#1}\@@href}%
\providecommand \@@href[1]{\endgroup#1\@@endlink}%
\providecommand \@sanitize@url [0]{\catcode `\\12\catcode `\$12\catcode
  `\&12\catcode `\#12\catcode `\^12\catcode `\_12\catcode `\%12\relax}%
\providecommand \@@startlink[1]{}%
\providecommand \@@endlink[0]{}%
\providecommand \url  [0]{\begingroup\@sanitize@url \@url }%
\providecommand \@url [1]{\endgroup\@href {#1}{\urlprefix }}%
\providecommand \urlprefix  [0]{URL }%
\providecommand \Eprint [0]{\href }%
\providecommand \doibase [0]{https://doi.org/}%
\providecommand \selectlanguage [0]{\@gobble}%
\providecommand \bibinfo  [0]{\@secondoftwo}%
\providecommand \bibfield  [0]{\@secondoftwo}%
\providecommand \translation [1]{[#1]}%
\providecommand \BibitemOpen [0]{}%
\providecommand \bibitemStop [0]{}%
\providecommand \bibitemNoStop [0]{.\EOS\space}%
\providecommand \EOS [0]{\spacefactor3000\relax}%
\providecommand \BibitemShut  [1]{\csname bibitem#1\endcsname}%
\let\auto@bib@innerbib\@empty
\bibitem [{\citenamefont {Mezard}\ and\ \citenamefont
  {Montanari}(2009)}]{mezard2009}%
  \BibitemOpen
  \bibfield  {author} {\bibinfo {author} {\bibfnamefont {M.}~\bibnamefont
  {Mezard}}\ and\ \bibinfo {author} {\bibfnamefont {A.}~\bibnamefont
  {Montanari}},\ }\href@noop {} {\emph {\bibinfo {title} {Information, Physics,
  and Computation}}}\ (\bibinfo  {publisher} {Oxford University Press, Inc.},\
  \bibinfo {address} {USA},\ \bibinfo {year} {2009})\BibitemShut {NoStop}%
\bibitem [{\citenamefont {Zdeborov{\'a}}\ and\ \citenamefont
  {Krzakala}(2016)}]{zdeborova2016}%
  \BibitemOpen
  \bibfield  {author} {\bibinfo {author} {\bibfnamefont {L.}~\bibnamefont
  {Zdeborov{\'a}}}\ and\ \bibinfo {author} {\bibfnamefont {F.}~\bibnamefont
  {Krzakala}},\ }\bibfield  {title} {\bibinfo {title} {Statistical physics of
  inference: thresholds and algorithms},\ }\href
  {https://doi.org/10.1080/00018732.2016.1211393} {\bibfield  {journal}
  {\bibinfo  {journal} {Advances in Physics}\ }\textbf {\bibinfo {volume}
  {65}},\ \bibinfo {pages} {453} (\bibinfo {year} {2016})}\BibitemShut
  {NoStop}%
\bibitem [{\citenamefont {Gamarnik}(2025)}]{gamarnik2025}%
  \BibitemOpen
  \bibfield  {author} {\bibinfo {author} {\bibfnamefont {D.}~\bibnamefont
  {Gamarnik}},\ }\href {https://arxiv.org/abs/2501.15312} {\bibinfo {title}
  {Turing in the shadows of nobel and abel: an algorithmic story behind two
  recent prizes}} (\bibinfo {year} {2025}),\ \Eprint
  {https://arxiv.org/abs/2501.15312} {arXiv:2501.15312 [math.PR]} \BibitemShut
  {NoStop}%
\bibitem [{\citenamefont {Kirkpatrick}\ \emph {et~al.}(1983)\citenamefont
  {Kirkpatrick}, \citenamefont {Gelatt},\ and\ \citenamefont
  {Vecchi}}]{kirkpatrick1983}%
  \BibitemOpen
  \bibfield  {author} {\bibinfo {author} {\bibfnamefont {S.}~\bibnamefont
  {Kirkpatrick}}, \bibinfo {author} {\bibfnamefont {C.~D.}\ \bibnamefont
  {Gelatt}},\ and\ \bibinfo {author} {\bibfnamefont {M.~P.}\ \bibnamefont
  {Vecchi}},\ }\bibfield  {title} {\bibinfo {title} {Optimization by simulated
  annealing},\ }\href {https://doi.org/10.1126/science.220.4598.671} {\bibfield
   {journal} {\bibinfo  {journal} {Science}\ }\textbf {\bibinfo {volume}
  {220}},\ \bibinfo {pages} {671} (\bibinfo {year} {1983})}\BibitemShut
  {NoStop}%
\bibitem [{\citenamefont {Earl}\ and\ \citenamefont {Deem}(2005)}]{earl2005}%
  \BibitemOpen
  \bibfield  {author} {\bibinfo {author} {\bibfnamefont {D.~J.}\ \bibnamefont
  {Earl}}\ and\ \bibinfo {author} {\bibfnamefont {M.~W.}\ \bibnamefont
  {Deem}},\ }\bibfield  {title} {\bibinfo {title} {Parallel tempering:
  Theory{,} applications{,} and new perspectives},\ }\href
  {https://doi.org/10.1039/B509983H} {\bibfield  {journal} {\bibinfo  {journal}
  {Phys. Chem. Chem. Phys.}\ }\textbf {\bibinfo {volume} {7}},\ \bibinfo
  {pages} {3910} (\bibinfo {year} {2005})}\BibitemShut {NoStop}%
\bibitem [{\citenamefont {Mohseni}\ \emph {et~al.}(2022)\citenamefont
  {Mohseni}, \citenamefont {McMahon},\ and\ \citenamefont
  {Byrnes}}]{mohseni2022}%
  \BibitemOpen
  \bibfield  {author} {\bibinfo {author} {\bibfnamefont {N.}~\bibnamefont
  {Mohseni}}, \bibinfo {author} {\bibfnamefont {P.~L.}\ \bibnamefont
  {McMahon}},\ and\ \bibinfo {author} {\bibfnamefont {T.}~\bibnamefont
  {Byrnes}},\ }\bibfield  {title} {\bibinfo {title} {Ising machines as hardware
  solvers of combinatorial optimization problems},\ }\href
  {https://doi.org/10.1038/s42254-022-00440-8} {\bibfield  {journal} {\bibinfo
  {journal} {Nature Reviews Physics}\ }\textbf {\bibinfo {volume} {4}},\
  \bibinfo {pages} {363} (\bibinfo {year} {2022})}\BibitemShut {NoStop}%
\bibitem [{\citenamefont {Hauke}\ \emph {et~al.}(2020)\citenamefont {Hauke},
  \citenamefont {Katzgraber}, \citenamefont {Lechner}, \citenamefont
  {Nishimori},\ and\ \citenamefont {Oliver}}]{hauke2020}%
  \BibitemOpen
  \bibfield  {author} {\bibinfo {author} {\bibfnamefont {P.}~\bibnamefont
  {Hauke}}, \bibinfo {author} {\bibfnamefont {H.~G.}\ \bibnamefont
  {Katzgraber}}, \bibinfo {author} {\bibfnamefont {W.}~\bibnamefont {Lechner}},
  \bibinfo {author} {\bibfnamefont {H.}~\bibnamefont {Nishimori}},\ and\
  \bibinfo {author} {\bibfnamefont {W.~D.}\ \bibnamefont {Oliver}},\ }\bibfield
   {title} {\bibinfo {title} {Perspectives of quantum annealing: methods and
  implementations},\ }\href {https://doi.org/10.1088/1361-6633/ab85b8}
  {\bibfield  {journal} {\bibinfo  {journal} {Reports on Progress in Physics}\
  }\textbf {\bibinfo {volume} {83}},\ \bibinfo {pages} {054401} (\bibinfo
  {year} {2020})}\BibitemShut {NoStop}%
\bibitem [{\citenamefont {Cai}\ \emph {et~al.}(2020)\citenamefont {Cai},
  \citenamefont {Kumar}, \citenamefont {Van~Vaerenbergh}, \citenamefont
  {Sheng}, \citenamefont {Liu}, \citenamefont {Li}, \citenamefont {Liu},
  \citenamefont {Foltin}, \citenamefont {Yu}, \citenamefont {Xia},
  \citenamefont {Yang}, \citenamefont {Beausoleil}, \citenamefont {Lu},\ and\
  \citenamefont {Strachan}}]{cai2020}%
  \BibitemOpen
  \bibfield  {author} {\bibinfo {author} {\bibfnamefont {F.}~\bibnamefont
  {Cai}}, \bibinfo {author} {\bibfnamefont {S.}~\bibnamefont {Kumar}}, \bibinfo
  {author} {\bibfnamefont {T.}~\bibnamefont {Van~Vaerenbergh}}, \bibinfo
  {author} {\bibfnamefont {X.}~\bibnamefont {Sheng}}, \bibinfo {author}
  {\bibfnamefont {R.}~\bibnamefont {Liu}}, \bibinfo {author} {\bibfnamefont
  {C.}~\bibnamefont {Li}}, \bibinfo {author} {\bibfnamefont {Z.}~\bibnamefont
  {Liu}}, \bibinfo {author} {\bibfnamefont {M.}~\bibnamefont {Foltin}},
  \bibinfo {author} {\bibfnamefont {S.}~\bibnamefont {Yu}}, \bibinfo {author}
  {\bibfnamefont {Q.}~\bibnamefont {Xia}}, \bibinfo {author} {\bibfnamefont
  {J.~J.}\ \bibnamefont {Yang}}, \bibinfo {author} {\bibfnamefont
  {R.}~\bibnamefont {Beausoleil}}, \bibinfo {author} {\bibfnamefont {W.~D.}\
  \bibnamefont {Lu}},\ and\ \bibinfo {author} {\bibfnamefont {J.~P.}\
  \bibnamefont {Strachan}},\ }\bibfield  {title} {\bibinfo {title}
  {Power-efficient combinatorial optimization using intrinsic noise in
  memristor hopfield neural networks},\ }\href
  {https://doi.org/10.1038/s41928-020-0436-6} {\bibfield  {journal} {\bibinfo
  {journal} {Nature Electronics}\ }\textbf {\bibinfo {volume} {3}},\ \bibinfo
  {pages} {409} (\bibinfo {year} {2020})}\BibitemShut {NoStop}%
\bibitem [{\citenamefont {Albash}\ and\ \citenamefont
  {Lidar}(2018)}]{albash2018}%
  \BibitemOpen
  \bibfield  {author} {\bibinfo {author} {\bibfnamefont {T.}~\bibnamefont
  {Albash}}\ and\ \bibinfo {author} {\bibfnamefont {D.~A.}\ \bibnamefont
  {Lidar}},\ }\bibfield  {title} {\bibinfo {title} {Demonstration of a scaling
  advantage for a quantum annealer over simulated annealing},\ }\href
  {https://doi.org/10.1103/PhysRevX.8.031016} {\bibfield  {journal} {\bibinfo
  {journal} {Phys. Rev. X}\ }\textbf {\bibinfo {volume} {8}},\ \bibinfo {pages}
  {031016} (\bibinfo {year} {2018})}\BibitemShut {NoStop}%
\bibitem [{\citenamefont {Gamarnik}(2021)}]{gamarnik2021}%
  \BibitemOpen
  \bibfield  {author} {\bibinfo {author} {\bibfnamefont {D.}~\bibnamefont
  {Gamarnik}},\ }\bibfield  {title} {\bibinfo {title} {The overlap gap
  property: A topological barrier to optimizing over random structures},\
  }\href {https://doi.org/10.1073/pnas.2108492118} {\bibfield  {journal}
  {\bibinfo  {journal} {Proceedings of the National Academy of Sciences}\
  }\textbf {\bibinfo {volume} {118}},\ \bibinfo {pages} {e2108492118} (\bibinfo
  {year} {2021})},\ \Eprint
  {https://arxiv.org/abs/https://www.pnas.org/doi/pdf/10.1073/pnas.2108492118}
  {https://www.pnas.org/doi/pdf/10.1073/pnas.2108492118} \BibitemShut {NoStop}%
\bibitem [{\citenamefont {Swendsen}\ and\ \citenamefont
  {Wang}(1987)}]{swendsen1987}%
  \BibitemOpen
  \bibfield  {author} {\bibinfo {author} {\bibfnamefont {R.~H.}\ \bibnamefont
  {Swendsen}}\ and\ \bibinfo {author} {\bibfnamefont {J.-S.}\ \bibnamefont
  {Wang}},\ }\bibfield  {title} {\bibinfo {title} {Nonuniversal critical
  dynamics in monte carlo simulations},\ }\href
  {https://doi.org/10.1103/PhysRevLett.58.86} {\bibfield  {journal} {\bibinfo
  {journal} {Phys. Rev. Lett.}\ }\textbf {\bibinfo {volume} {58}},\ \bibinfo
  {pages} {86} (\bibinfo {year} {1987})}\BibitemShut {NoStop}%
\bibitem [{\citenamefont {Wolff}(1989)}]{wolff1989}%
  \BibitemOpen
  \bibfield  {author} {\bibinfo {author} {\bibfnamefont {U.}~\bibnamefont
  {Wolff}},\ }\bibfield  {title} {\bibinfo {title} {Collective monte carlo
  updating for spin systems},\ }\href
  {https://doi.org/10.1103/PhysRevLett.62.361} {\bibfield  {journal} {\bibinfo
  {journal} {Phys. Rev. Lett.}\ }\textbf {\bibinfo {volume} {62}},\ \bibinfo
  {pages} {361} (\bibinfo {year} {1989})}\BibitemShut {NoStop}%
\bibitem [{\citenamefont {Houdayer}(2001)}]{houdayer2001}%
  \BibitemOpen
  \bibfield  {author} {\bibinfo {author} {\bibfnamefont {J.}~\bibnamefont
  {Houdayer}},\ }\bibfield  {title} {\bibinfo {title} {A cluster monte carlo
  algorithm for 2-dimensional spin glasses},\ }\href
  {https://doi.org/10.1007/PL00011151} {\bibfield  {journal} {\bibinfo
  {journal} {The European Physical Journal B - Condensed Matter and Complex
  Systems}\ }\textbf {\bibinfo {volume} {22}},\ \bibinfo {pages} {479}
  (\bibinfo {year} {2001})}\BibitemShut {NoStop}%
\bibitem [{\citenamefont {Hamze}\ and\ \citenamefont
  {de~Freitas}(2004)}]{hamze2004}%
  \BibitemOpen
  \bibfield  {author} {\bibinfo {author} {\bibfnamefont {F.}~\bibnamefont
  {Hamze}}\ and\ \bibinfo {author} {\bibfnamefont {N.}~\bibnamefont
  {de~Freitas}},\ }\bibfield  {title} {\bibinfo {title} {From fields to
  trees},\ }in\ \href@noop {} {\emph {\bibinfo {booktitle} {Proceedings of the
  20th Conference on Uncertainty in Artificial Intelligence}}},\ \bibinfo
  {series and number} {UAI '04}\ (\bibinfo  {publisher} {AUAI Press},\ \bibinfo
  {address} {Arlington, Virginia, USA},\ \bibinfo {year} {2004})\ pp.\ \bibinfo
  {pages} {243--250}\BibitemShut {NoStop}%
\bibitem [{\citenamefont {Selby}(2014)}]{selby2014}%
  \BibitemOpen
  \bibfield  {author} {\bibinfo {author} {\bibfnamefont {A.}~\bibnamefont
  {Selby}},\ }\href {https://arxiv.org/abs/1409.3934} {\bibinfo {title}
  {Efficient subgraph-based sampling of ising-type models with frustration}}
  (\bibinfo {year} {2014}),\ \Eprint {https://arxiv.org/abs/1409.3934}
  {arXiv:1409.3934 [cond-mat.stat-mech]} \BibitemShut {NoStop}%
\bibitem [{\citenamefont {Zhu}\ \emph {et~al.}(2015)\citenamefont {Zhu},
  \citenamefont {Ochoa},\ and\ \citenamefont {Katzgraber}}]{zhu2015}%
  \BibitemOpen
  \bibfield  {author} {\bibinfo {author} {\bibfnamefont {Z.}~\bibnamefont
  {Zhu}}, \bibinfo {author} {\bibfnamefont {A.~J.}\ \bibnamefont {Ochoa}},\
  and\ \bibinfo {author} {\bibfnamefont {H.~G.}\ \bibnamefont {Katzgraber}},\
  }\bibfield  {title} {\bibinfo {title} {Efficient cluster algorithm for spin
  glasses in any space dimension},\ }\href
  {https://doi.org/10.1103/PhysRevLett.115.077201} {\bibfield  {journal}
  {\bibinfo  {journal} {Phys. Rev. Lett.}\ }\textbf {\bibinfo {volume} {115}},\
  \bibinfo {pages} {077201} (\bibinfo {year} {2015})}\BibitemShut {NoStop}%
\bibitem [{\citenamefont {Hen}(2017)}]{hen2017}%
  \BibitemOpen
  \bibfield  {author} {\bibinfo {author} {\bibfnamefont {I.}~\bibnamefont
  {Hen}},\ }\bibfield  {title} {\bibinfo {title} {Solving spin glasses with
  optimized trees of clustered spins},\ }\href
  {https://doi.org/10.1103/PhysRevE.96.022105} {\bibfield  {journal} {\bibinfo
  {journal} {Phys. Rev. E}\ }\textbf {\bibinfo {volume} {96}},\ \bibinfo
  {pages} {022105} (\bibinfo {year} {2017})}\BibitemShut {NoStop}%
\bibitem [{\citenamefont {Mohseni}\ \emph {et~al.}(2021)\citenamefont
  {Mohseni}, \citenamefont {Eppens}, \citenamefont {Strumpfer}, \citenamefont
  {Marino}, \citenamefont {Denchev}, \citenamefont {Ho}, \citenamefont
  {Isakov}, \citenamefont {Boixo}, \citenamefont {Ricci-Tersenghi},\ and\
  \citenamefont {Neven}}]{mohseni2021}%
  \BibitemOpen
  \bibfield  {author} {\bibinfo {author} {\bibfnamefont {M.}~\bibnamefont
  {Mohseni}}, \bibinfo {author} {\bibfnamefont {D.}~\bibnamefont {Eppens}},
  \bibinfo {author} {\bibfnamefont {J.}~\bibnamefont {Strumpfer}}, \bibinfo
  {author} {\bibfnamefont {R.}~\bibnamefont {Marino}}, \bibinfo {author}
  {\bibfnamefont {V.}~\bibnamefont {Denchev}}, \bibinfo {author} {\bibfnamefont
  {A.~K.}\ \bibnamefont {Ho}}, \bibinfo {author} {\bibfnamefont {S.~V.}\
  \bibnamefont {Isakov}}, \bibinfo {author} {\bibfnamefont {S.}~\bibnamefont
  {Boixo}}, \bibinfo {author} {\bibfnamefont {F.}~\bibnamefont
  {Ricci-Tersenghi}},\ and\ \bibinfo {author} {\bibfnamefont {H.}~\bibnamefont
  {Neven}},\ }\href {https://arxiv.org/abs/2111.13628} {\bibinfo {title}
  {Nonequilibrium monte carlo for unfreezing variables in hard combinatorial
  optimization}} (\bibinfo {year} {2021}),\ \Eprint
  {https://arxiv.org/abs/2111.13628} {arXiv:2111.13628 [cond-mat.dis-nn]}
  \BibitemShut {NoStop}%
\bibitem [{\citenamefont {Mohseni}\ \emph
  {et~al.}(2023{\natexlab{a}})\citenamefont {Mohseni}, \citenamefont {Rams},
  \citenamefont {Isakov}, \citenamefont {Eppens}, \citenamefont {Pielawa},
  \citenamefont {Strumpfer}, \citenamefont {Boixo},\ and\ \citenamefont
  {Neven}}]{mohseni2023}%
  \BibitemOpen
  \bibfield  {author} {\bibinfo {author} {\bibfnamefont {M.}~\bibnamefont
  {Mohseni}}, \bibinfo {author} {\bibfnamefont {M.~M.}\ \bibnamefont {Rams}},
  \bibinfo {author} {\bibfnamefont {S.~V.}\ \bibnamefont {Isakov}}, \bibinfo
  {author} {\bibfnamefont {D.}~\bibnamefont {Eppens}}, \bibinfo {author}
  {\bibfnamefont {S.}~\bibnamefont {Pielawa}}, \bibinfo {author} {\bibfnamefont
  {J.}~\bibnamefont {Strumpfer}}, \bibinfo {author} {\bibfnamefont
  {S.}~\bibnamefont {Boixo}},\ and\ \bibinfo {author} {\bibfnamefont
  {H.}~\bibnamefont {Neven}},\ }\bibfield  {title} {\bibinfo {title} {Sampling
  diverse near-optimal solutions via algorithmic quantum annealing},\ }\href
  {https://doi.org/10.1103/PhysRevE.108.065303} {\bibfield  {journal} {\bibinfo
   {journal} {Phys. Rev. E}\ }\textbf {\bibinfo {volume} {108}},\ \bibinfo
  {pages} {065303} (\bibinfo {year} {2023}{\natexlab{a}})}\BibitemShut
  {NoStop}%
\bibitem [{\citenamefont {Bengio}\ \emph {et~al.}(2020)\citenamefont {Bengio},
  \citenamefont {Lodi},\ and\ \citenamefont {Prouvost}}]{bengio2020}%
  \BibitemOpen
  \bibfield  {author} {\bibinfo {author} {\bibfnamefont {Y.}~\bibnamefont
  {Bengio}}, \bibinfo {author} {\bibfnamefont {A.}~\bibnamefont {Lodi}},\ and\
  \bibinfo {author} {\bibfnamefont {A.}~\bibnamefont {Prouvost}},\ }\href
  {https://arxiv.org/abs/1811.06128} {\bibinfo {title} {Machine learning for
  combinatorial optimization: a methodological tour d'horizon}} (\bibinfo
  {year} {2020}),\ \Eprint {https://arxiv.org/abs/1811.06128} {arXiv:1811.06128
  [cs.LG]} \BibitemShut {NoStop}%
\bibitem [{\citenamefont {Sutton}\ and\ \citenamefont
  {Barto}(2018)}]{sutton2018}%
  \BibitemOpen
  \bibfield  {author} {\bibinfo {author} {\bibfnamefont {R.~S.}\ \bibnamefont
  {Sutton}}\ and\ \bibinfo {author} {\bibfnamefont {A.~G.}\ \bibnamefont
  {Barto}},\ }\href@noop {} {\emph {\bibinfo {title} {Reinforcement Learning:
  An Introduction}}}\ (\bibinfo  {publisher} {A Bradford Book},\ \bibinfo
  {address} {Cambridge, MA, USA},\ \bibinfo {year} {2018})\BibitemShut
  {NoStop}%
\bibitem [{rln(2025)}]{rlnmc_code}%
  \BibitemOpen
  \href {https://github.com/dumdob/rlnmc.git} {\bibinfo {title} {Reinforcement
  learning nonlocal monte carlo, github.com/dumdob/rlnmc.git}} (\bibinfo {year}
  {2025})\BibitemShut {NoStop}%
\bibitem [{\citenamefont {Bello}\ \emph {et~al.}(2017)\citenamefont {Bello},
  \citenamefont {Pham}, \citenamefont {Le}, \citenamefont {Norouzi},\ and\
  \citenamefont {Bengio}}]{bello2017}%
  \BibitemOpen
  \bibfield  {author} {\bibinfo {author} {\bibfnamefont {I.}~\bibnamefont
  {Bello}}, \bibinfo {author} {\bibfnamefont {H.}~\bibnamefont {Pham}},
  \bibinfo {author} {\bibfnamefont {Q.~V.}\ \bibnamefont {Le}}, \bibinfo
  {author} {\bibfnamefont {M.}~\bibnamefont {Norouzi}},\ and\ \bibinfo {author}
  {\bibfnamefont {S.}~\bibnamefont {Bengio}},\ }\href
  {https://arxiv.org/abs/1611.09940} {\bibinfo {title} {Neural combinatorial
  optimization with reinforcement learning}} (\bibinfo {year} {2017}),\ \Eprint
  {https://arxiv.org/abs/1611.09940} {arXiv:1611.09940 [cs.AI]} \BibitemShut
  {NoStop}%
\bibitem [{\citenamefont {Li}\ \emph {et~al.}(2018)\citenamefont {Li},
  \citenamefont {Chen},\ and\ \citenamefont {Koltun}}]{li2018}%
  \BibitemOpen
  \bibfield  {author} {\bibinfo {author} {\bibfnamefont {Z.}~\bibnamefont
  {Li}}, \bibinfo {author} {\bibfnamefont {Q.}~\bibnamefont {Chen}},\ and\
  \bibinfo {author} {\bibfnamefont {V.}~\bibnamefont {Koltun}},\ }\href
  {https://arxiv.org/abs/1810.10659} {\bibinfo {title} {Combinatorial
  optimization with graph convolutional networks and guided tree search}}
  (\bibinfo {year} {2018}),\ \Eprint {https://arxiv.org/abs/1810.10659}
  {arXiv:1810.10659 [cs.LG]} \BibitemShut {NoStop}%
\bibitem [{\citenamefont {He}\ \emph {et~al.}(2014)\citenamefont {He},
  \citenamefont {Daume~III},\ and\ \citenamefont {Eisner}}]{he2014}%
  \BibitemOpen
  \bibfield  {author} {\bibinfo {author} {\bibfnamefont {H.}~\bibnamefont
  {He}}, \bibinfo {author} {\bibfnamefont {H.}~\bibnamefont {Daume~III}},\ and\
  \bibinfo {author} {\bibfnamefont {J.~M.}\ \bibnamefont {Eisner}},\ }\bibfield
   {title} {\bibinfo {title} {Learning to search in branch and bound
  algorithms},\ }in\ \href
  {https://proceedings.neurips.cc/paper_files/paper/2014/file/757f843a169cc678064d9530d12a1881-Paper.pdf}
  {\emph {\bibinfo {booktitle} {Advances in Neural Information Processing
  Systems}}},\ Vol.~\bibinfo {volume} {27},\ \bibinfo {editor} {edited by\
  \bibinfo {editor} {\bibfnamefont {Z.}~\bibnamefont {Ghahramani}}, \bibinfo
  {editor} {\bibfnamefont {M.}~\bibnamefont {Welling}}, \bibinfo {editor}
  {\bibfnamefont {C.}~\bibnamefont {Cortes}}, \bibinfo {editor} {\bibfnamefont
  {N.}~\bibnamefont {Lawrence}},\ and\ \bibinfo {editor} {\bibfnamefont
  {K.}~\bibnamefont {Weinberger}}}\ (\bibinfo  {publisher} {Curran Associates,
  Inc.},\ \bibinfo {year} {2014})\BibitemShut {NoStop}%
\bibitem [{\citenamefont {Kool}\ \emph {et~al.}(2019)\citenamefont {Kool},
  \citenamefont {van Hoof},\ and\ \citenamefont {Welling}}]{kool2019}%
  \BibitemOpen
  \bibfield  {author} {\bibinfo {author} {\bibfnamefont {W.}~\bibnamefont
  {Kool}}, \bibinfo {author} {\bibfnamefont {H.}~\bibnamefont {van Hoof}},\
  and\ \bibinfo {author} {\bibfnamefont {M.}~\bibnamefont {Welling}},\ }\href
  {https://arxiv.org/abs/1803.08475} {\bibinfo {title} {Attention, learn to
  solve routing problems!}} (\bibinfo {year} {2019}),\ \Eprint
  {https://arxiv.org/abs/1803.08475} {arXiv:1803.08475 [stat.ML]} \BibitemShut
  {NoStop}%
\bibitem [{\citenamefont {Dai}\ \emph {et~al.}(2018)\citenamefont {Dai},
  \citenamefont {Khalil}, \citenamefont {Zhang}, \citenamefont {Dilkina},\ and\
  \citenamefont {Song}}]{dai2018}%
  \BibitemOpen
  \bibfield  {author} {\bibinfo {author} {\bibfnamefont {H.}~\bibnamefont
  {Dai}}, \bibinfo {author} {\bibfnamefont {E.~B.}\ \bibnamefont {Khalil}},
  \bibinfo {author} {\bibfnamefont {Y.}~\bibnamefont {Zhang}}, \bibinfo
  {author} {\bibfnamefont {B.}~\bibnamefont {Dilkina}},\ and\ \bibinfo {author}
  {\bibfnamefont {L.}~\bibnamefont {Song}},\ }\href
  {https://arxiv.org/abs/1704.01665} {\bibinfo {title} {Learning combinatorial
  optimization algorithms over graphs}} (\bibinfo {year} {2018}),\ \Eprint
  {https://arxiv.org/abs/1704.01665} {arXiv:1704.01665 [cs.LG]} \BibitemShut
  {NoStop}%
\bibitem [{\citenamefont {Barrett}\ \emph {et~al.}(2020)\citenamefont
  {Barrett}, \citenamefont {Clements}, \citenamefont {Foerster},\ and\
  \citenamefont {Lvovsky}}]{barrett2020}%
  \BibitemOpen
  \bibfield  {author} {\bibinfo {author} {\bibfnamefont {T.~D.}\ \bibnamefont
  {Barrett}}, \bibinfo {author} {\bibfnamefont {W.~R.}\ \bibnamefont
  {Clements}}, \bibinfo {author} {\bibfnamefont {J.~N.}\ \bibnamefont
  {Foerster}},\ and\ \bibinfo {author} {\bibfnamefont {A.~I.}\ \bibnamefont
  {Lvovsky}},\ }\href {https://arxiv.org/abs/1909.04063} {\bibinfo {title}
  {Exploratory combinatorial optimization with reinforcement learning}}
  (\bibinfo {year} {2020}),\ \Eprint {https://arxiv.org/abs/1909.04063}
  {arXiv:1909.04063 [cs.LG]} \BibitemShut {NoStop}%
\bibitem [{\citenamefont {Barrett}\ \emph {et~al.}(2022)\citenamefont
  {Barrett}, \citenamefont {Parsonson},\ and\ \citenamefont
  {Laterre}}]{barrett2022}%
  \BibitemOpen
  \bibfield  {author} {\bibinfo {author} {\bibfnamefont {T.~D.}\ \bibnamefont
  {Barrett}}, \bibinfo {author} {\bibfnamefont {C.~W.~F.}\ \bibnamefont
  {Parsonson}},\ and\ \bibinfo {author} {\bibfnamefont {A.}~\bibnamefont
  {Laterre}},\ }\href {https://arxiv.org/abs/2205.14105} {\bibinfo {title}
  {Learning to solve combinatorial graph partitioning problems via efficient
  exploration}} (\bibinfo {year} {2022}),\ \Eprint
  {https://arxiv.org/abs/2205.14105} {arXiv:2205.14105 [cs.LG]} \BibitemShut
  {NoStop}%
\bibitem [{\citenamefont {T{\"o}nshoff}\ \emph {et~al.}(2023)\citenamefont
  {T{\"o}nshoff}, \citenamefont {Kisin}, \citenamefont {Lindner},\ and\
  \citenamefont {Grohe}}]{tonshoff2023}%
  \BibitemOpen
  \bibfield  {author} {\bibinfo {author} {\bibfnamefont {J.}~\bibnamefont
  {T{\"o}nshoff}}, \bibinfo {author} {\bibfnamefont {B.}~\bibnamefont {Kisin}},
  \bibinfo {author} {\bibfnamefont {J.}~\bibnamefont {Lindner}},\ and\ \bibinfo
  {author} {\bibfnamefont {M.}~\bibnamefont {Grohe}},\ }\bibfield  {title}
  {\bibinfo {title} {One model, any csp: Graph neural networks as fast global
  search heuristics for constraint satisfaction},\ }in\ \href
  {https://doi.org/10.24963/ijcai.2023/476} {\emph {\bibinfo {booktitle}
  {Proceedings of the Thirty-Second International Joint Conference on
  Artificial Intelligence, {IJCAI-23}}}},\ \bibinfo {editor} {edited by\
  \bibinfo {editor} {\bibfnamefont {E.}~\bibnamefont {Elkind}}}\ (\bibinfo
  {publisher} {International Joint Conferences on Artificial Intelligence
  Organization},\ \bibinfo {year} {2023})\ pp.\ \bibinfo {pages} {4280--4288},\
  \bibinfo {note} {main Track}\BibitemShut {NoStop}%
\bibitem [{\citenamefont {Mazyavkina}\ \emph {et~al.}(2021)\citenamefont
  {Mazyavkina}, \citenamefont {Sviridov}, \citenamefont {Ivanov},\ and\
  \citenamefont {Burnaev}}]{mazyavkina2021}%
  \BibitemOpen
  \bibfield  {author} {\bibinfo {author} {\bibfnamefont {N.}~\bibnamefont
  {Mazyavkina}}, \bibinfo {author} {\bibfnamefont {S.}~\bibnamefont
  {Sviridov}}, \bibinfo {author} {\bibfnamefont {S.}~\bibnamefont {Ivanov}},\
  and\ \bibinfo {author} {\bibfnamefont {E.}~\bibnamefont {Burnaev}},\
  }\bibfield  {title} {\bibinfo {title} {Reinforcement learning for
  combinatorial optimization: A survey},\ }\href
  {https://doi.org/https://doi.org/10.1016/j.cor.2021.105400} {\bibfield
  {journal} {\bibinfo  {journal} {Computers \& Operations Research}\ }\textbf
  {\bibinfo {volume} {134}},\ \bibinfo {pages} {105400} (\bibinfo {year}
  {2021})}\BibitemShut {NoStop}%
\bibitem [{\citenamefont {Ahn}\ \emph {et~al.}(2020)\citenamefont {Ahn},
  \citenamefont {Seo},\ and\ \citenamefont {Shin}}]{ahn2020}%
  \BibitemOpen
  \bibfield  {author} {\bibinfo {author} {\bibfnamefont {S.}~\bibnamefont
  {Ahn}}, \bibinfo {author} {\bibfnamefont {Y.}~\bibnamefont {Seo}},\ and\
  \bibinfo {author} {\bibfnamefont {J.}~\bibnamefont {Shin}},\ }\href
  {https://arxiv.org/abs/2006.09607} {\bibinfo {title} {Learning what to defer
  for maximum independent sets}} (\bibinfo {year} {2020}),\ \Eprint
  {https://arxiv.org/abs/2006.09607} {arXiv:2006.09607 [cs.LG]} \BibitemShut
  {NoStop}%
\bibitem [{\citenamefont {Mills}\ \emph {et~al.}(2020)\citenamefont {Mills},
  \citenamefont {Ronagh},\ and\ \citenamefont {Tamblyn}}]{mills2020}%
  \BibitemOpen
  \bibfield  {author} {\bibinfo {author} {\bibfnamefont {K.}~\bibnamefont
  {Mills}}, \bibinfo {author} {\bibfnamefont {P.}~\bibnamefont {Ronagh}},\ and\
  \bibinfo {author} {\bibfnamefont {I.}~\bibnamefont {Tamblyn}},\ }\bibfield
  {title} {\bibinfo {title} {Finding the ground state of spin hamiltonians with
  reinforcement learning},\ }\href {https://doi.org/10.1038/s42256-020-0226-x}
  {\bibfield  {journal} {\bibinfo  {journal} {Nature Machine Intelligence}\
  }\textbf {\bibinfo {volume} {2}},\ \bibinfo {pages} {509} (\bibinfo {year}
  {2020})}\BibitemShut {NoStop}%
\bibitem [{\citenamefont {Kool}\ \emph {et~al.}(2021)\citenamefont {Kool},
  \citenamefont {van Hoof}, \citenamefont {Gromicho},\ and\ \citenamefont
  {Welling}}]{kool2021}%
  \BibitemOpen
  \bibfield  {author} {\bibinfo {author} {\bibfnamefont {W.}~\bibnamefont
  {Kool}}, \bibinfo {author} {\bibfnamefont {H.}~\bibnamefont {van Hoof}},
  \bibinfo {author} {\bibfnamefont {J.}~\bibnamefont {Gromicho}},\ and\
  \bibinfo {author} {\bibfnamefont {M.}~\bibnamefont {Welling}},\ }\href
  {https://arxiv.org/abs/2102.11756} {\bibinfo {title} {Deep policy dynamic
  programming for vehicle routing problems}} (\bibinfo {year} {2021}),\ \Eprint
  {https://arxiv.org/abs/2102.11756} {arXiv:2102.11756 [cs.LG]} \BibitemShut
  {NoStop}%
\bibitem [{\citenamefont {B\"{o}ther}\ \emph {et~al.}(2022)\citenamefont
  {B\"{o}ther}, \citenamefont {Ki\ss{}ig}, \citenamefont {Taraz}, \citenamefont
  {Cohen}, \citenamefont {Seidel},\ and\ \citenamefont
  {Friedrich}}]{bother2022}%
  \BibitemOpen
  \bibfield  {author} {\bibinfo {author} {\bibfnamefont {M.}~\bibnamefont
  {B\"{o}ther}}, \bibinfo {author} {\bibfnamefont {O.}~\bibnamefont
  {Ki\ss{}ig}}, \bibinfo {author} {\bibfnamefont {M.}~\bibnamefont {Taraz}},
  \bibinfo {author} {\bibfnamefont {S.}~\bibnamefont {Cohen}}, \bibinfo
  {author} {\bibfnamefont {K.}~\bibnamefont {Seidel}},\ and\ \bibinfo {author}
  {\bibfnamefont {T.}~\bibnamefont {Friedrich}},\ }\href
  {https://arxiv.org/abs/2201.10494} {\bibinfo {title} {What's wrong with deep
  learning in tree search for combinatorial optimization}} (\bibinfo {year}
  {2022}),\ \Eprint {https://arxiv.org/abs/2201.10494} {arXiv:2201.10494
  [cs.LG]} \BibitemShut {NoStop}%
\bibitem [{\citenamefont {Zhou}\ \emph {et~al.}(2020)\citenamefont {Zhou},
  \citenamefont {Cui}, \citenamefont {Hu}, \citenamefont {Zhang}, \citenamefont
  {Yang}, \citenamefont {Liu}, \citenamefont {Wang}, \citenamefont {Li},\ and\
  \citenamefont {Sun}}]{zhou2020}%
  \BibitemOpen
  \bibfield  {author} {\bibinfo {author} {\bibfnamefont {J.}~\bibnamefont
  {Zhou}}, \bibinfo {author} {\bibfnamefont {G.}~\bibnamefont {Cui}}, \bibinfo
  {author} {\bibfnamefont {S.}~\bibnamefont {Hu}}, \bibinfo {author}
  {\bibfnamefont {Z.}~\bibnamefont {Zhang}}, \bibinfo {author} {\bibfnamefont
  {C.}~\bibnamefont {Yang}}, \bibinfo {author} {\bibfnamefont {Z.}~\bibnamefont
  {Liu}}, \bibinfo {author} {\bibfnamefont {L.}~\bibnamefont {Wang}}, \bibinfo
  {author} {\bibfnamefont {C.}~\bibnamefont {Li}},\ and\ \bibinfo {author}
  {\bibfnamefont {M.}~\bibnamefont {Sun}},\ }\bibfield  {title} {\bibinfo
  {title} {Graph neural networks: A review of methods and applications},\
  }\href {https://doi.org/https://doi.org/10.1016/j.aiopen.2021.01.001}
  {\bibfield  {journal} {\bibinfo  {journal} {AI Open}\ }\textbf {\bibinfo
  {volume} {1}},\ \bibinfo {pages} {57} (\bibinfo {year} {2020})}\BibitemShut
  {NoStop}%
\bibitem [{\citenamefont {Cappart}\ \emph {et~al.}(2023)\citenamefont
  {Cappart}, \citenamefont {Ch\'{e}telat}, \citenamefont {Khalil},
  \citenamefont {Lodi}, \citenamefont {Morris},\ and\ \citenamefont
  {Veli\v{c}kovi\'{c}}}]{cappart2023}%
  \BibitemOpen
  \bibfield  {author} {\bibinfo {author} {\bibfnamefont {Q.}~\bibnamefont
  {Cappart}}, \bibinfo {author} {\bibfnamefont {D.}~\bibnamefont
  {Ch\'{e}telat}}, \bibinfo {author} {\bibfnamefont {E.}~\bibnamefont
  {Khalil}}, \bibinfo {author} {\bibfnamefont {A.}~\bibnamefont {Lodi}},
  \bibinfo {author} {\bibfnamefont {C.}~\bibnamefont {Morris}},\ and\ \bibinfo
  {author} {\bibfnamefont {P.}~\bibnamefont {Veli\v{c}kovi\'{c}}},\ }\bibfield
  {title} {\bibinfo {title} {Combinatorial optimization and reasoning with
  graph neural networks},\ }\href {http://jmlr.org/papers/v24/21-0449.html}
  {\bibfield  {journal} {\bibinfo  {journal} {Journal of Machine Learning
  Research}\ }\textbf {\bibinfo {volume} {24}},\ \bibinfo {pages} {1} (\bibinfo
  {year} {2023})}\BibitemShut {NoStop}%
\bibitem [{\citenamefont {Perdomo-Ortiz}\ \emph {et~al.}(2019)\citenamefont
  {Perdomo-Ortiz}, \citenamefont {Feldman}, \citenamefont {Ozaeta},
  \citenamefont {Isakov}, \citenamefont {Zhu}, \citenamefont {O'Gorman},
  \citenamefont {Katzgraber}, \citenamefont {Diedrich}, \citenamefont {Neven},
  \citenamefont {de~Kleer}, \citenamefont {Lackey},\ and\ \citenamefont
  {Biswas}}]{perdomo2019}%
  \BibitemOpen
  \bibfield  {author} {\bibinfo {author} {\bibfnamefont {A.}~\bibnamefont
  {Perdomo-Ortiz}}, \bibinfo {author} {\bibfnamefont {A.}~\bibnamefont
  {Feldman}}, \bibinfo {author} {\bibfnamefont {A.}~\bibnamefont {Ozaeta}},
  \bibinfo {author} {\bibfnamefont {S.~V.}\ \bibnamefont {Isakov}}, \bibinfo
  {author} {\bibfnamefont {Z.}~\bibnamefont {Zhu}}, \bibinfo {author}
  {\bibfnamefont {B.}~\bibnamefont {O'Gorman}}, \bibinfo {author}
  {\bibfnamefont {H.~G.}\ \bibnamefont {Katzgraber}}, \bibinfo {author}
  {\bibfnamefont {A.}~\bibnamefont {Diedrich}}, \bibinfo {author}
  {\bibfnamefont {H.}~\bibnamefont {Neven}}, \bibinfo {author} {\bibfnamefont
  {J.}~\bibnamefont {de~Kleer}}, \bibinfo {author} {\bibfnamefont
  {B.}~\bibnamefont {Lackey}},\ and\ \bibinfo {author} {\bibfnamefont
  {R.}~\bibnamefont {Biswas}},\ }\bibfield  {title} {\bibinfo {title}
  {Readiness of quantum optimization machines for industrial applications},\
  }\href {https://doi.org/10.1103/PhysRevApplied.12.014004} {\bibfield
  {journal} {\bibinfo  {journal} {Phys. Rev. Appl.}\ }\textbf {\bibinfo
  {volume} {12}},\ \bibinfo {pages} {014004} (\bibinfo {year}
  {2019})}\BibitemShut {NoStop}%
\bibitem [{\citenamefont {Valiante}\ \emph {et~al.}(2021)\citenamefont
  {Valiante}, \citenamefont {Hernandez}, \citenamefont {Barzegar},\ and\
  \citenamefont {Katzgraber}}]{valiante2021}%
  \BibitemOpen
  \bibfield  {author} {\bibinfo {author} {\bibfnamefont {E.}~\bibnamefont
  {Valiante}}, \bibinfo {author} {\bibfnamefont {M.}~\bibnamefont {Hernandez}},
  \bibinfo {author} {\bibfnamefont {A.}~\bibnamefont {Barzegar}},\ and\
  \bibinfo {author} {\bibfnamefont {H.~G.}\ \bibnamefont {Katzgraber}},\
  }\bibfield  {title} {\bibinfo {title} {Computational overhead of locality
  reduction in binary optimization problems},\ }\href
  {https://doi.org/https://doi.org/10.1016/j.cpc.2021.108102} {\bibfield
  {journal} {\bibinfo  {journal} {Computer Physics Communications}\ }\textbf
  {\bibinfo {volume} {269}},\ \bibinfo {pages} {108102} (\bibinfo {year}
  {2021})}\BibitemShut {NoStop}%
\bibitem [{\citenamefont {Dobrynin}\ \emph
  {et~al.}(2024{\natexlab{a}})\citenamefont {Dobrynin}, \citenamefont
  {Renaudineau}, \citenamefont {Hizzani}, \citenamefont {Strukov},
  \citenamefont {Mohseni},\ and\ \citenamefont {Strachan}}]{dobrynin2024}%
  \BibitemOpen
  \bibfield  {author} {\bibinfo {author} {\bibfnamefont {D.}~\bibnamefont
  {Dobrynin}}, \bibinfo {author} {\bibfnamefont {A.}~\bibnamefont
  {Renaudineau}}, \bibinfo {author} {\bibfnamefont {M.}~\bibnamefont
  {Hizzani}}, \bibinfo {author} {\bibfnamefont {D.}~\bibnamefont {Strukov}},
  \bibinfo {author} {\bibfnamefont {M.}~\bibnamefont {Mohseni}},\ and\ \bibinfo
  {author} {\bibfnamefont {J.~P.}\ \bibnamefont {Strachan}},\ }\bibfield
  {title} {\bibinfo {title} {Energy landscapes of combinatorial optimization in
  ising machines},\ }\href {https://doi.org/10.1103/PhysRevE.110.045308}
  {\bibfield  {journal} {\bibinfo  {journal} {Phys. Rev. E}\ }\textbf {\bibinfo
  {volume} {110}},\ \bibinfo {pages} {045308} (\bibinfo {year}
  {2024}{\natexlab{a}})}\BibitemShut {NoStop}%
\bibitem [{\citenamefont {Dobrynin}\ \emph
  {et~al.}(2024{\natexlab{b}})\citenamefont {Dobrynin}, \citenamefont
  {Tedeschi}, \citenamefont {Heittmann},\ and\ \citenamefont
  {Strachan}}]{dobrynin2024b}%
  \BibitemOpen
  \bibfield  {author} {\bibinfo {author} {\bibfnamefont {D.}~\bibnamefont
  {Dobrynin}}, \bibinfo {author} {\bibfnamefont {M.}~\bibnamefont {Tedeschi}},
  \bibinfo {author} {\bibfnamefont {A.}~\bibnamefont {Heittmann}},\ and\
  \bibinfo {author} {\bibfnamefont {J.~P.}\ \bibnamefont {Strachan}},\
  }\bibfield  {title} {\bibinfo {title} {Gradient matching of higher order
  combinatorial optimization in quadratic ising machines},\ }in\ \href
  {https://doi.org/10.1109/ICRC64395.2024.10937013} {\emph {\bibinfo
  {booktitle} {2024 IEEE International Conference on Rebooting Computing
  (ICRC)}}}\ (\bibinfo {year} {2024})\ pp.\ \bibinfo {pages}
  {1--10}\BibitemShut {NoStop}%
\bibitem [{\citenamefont {Fan}\ \emph {et~al.}(2023)\citenamefont {Fan},
  \citenamefont {Shen}, \citenamefont {Nussinov}, \citenamefont {Liu},
  \citenamefont {Sun},\ and\ \citenamefont {Liu}}]{fan2023}%
  \BibitemOpen
  \bibfield  {author} {\bibinfo {author} {\bibfnamefont {C.}~\bibnamefont
  {Fan}}, \bibinfo {author} {\bibfnamefont {M.}~\bibnamefont {Shen}}, \bibinfo
  {author} {\bibfnamefont {Z.}~\bibnamefont {Nussinov}}, \bibinfo {author}
  {\bibfnamefont {Z.}~\bibnamefont {Liu}}, \bibinfo {author} {\bibfnamefont
  {Y.}~\bibnamefont {Sun}},\ and\ \bibinfo {author} {\bibfnamefont {Y.-Y.}\
  \bibnamefont {Liu}},\ }\bibfield  {title} {\bibinfo {title} {Searching for
  spin glass ground states through deep reinforcement learning},\ }\href
  {https://doi.org/10.1038/s41467-023-36363-w} {\bibfield  {journal} {\bibinfo
  {journal} {Nature Communications}\ }\textbf {\bibinfo {volume} {14}},\
  \bibinfo {pages} {725} (\bibinfo {year} {2023})}\BibitemShut {NoStop}%
\bibitem [{\citenamefont {Schuetz}\ \emph
  {et~al.}(2022{\natexlab{a}})\citenamefont {Schuetz}, \citenamefont
  {Brubaker},\ and\ \citenamefont {Katzgraber}}]{schuetz2022}%
  \BibitemOpen
  \bibfield  {author} {\bibinfo {author} {\bibfnamefont {M.~J.~A.}\
  \bibnamefont {Schuetz}}, \bibinfo {author} {\bibfnamefont {J.~K.}\
  \bibnamefont {Brubaker}},\ and\ \bibinfo {author} {\bibfnamefont {H.~G.}\
  \bibnamefont {Katzgraber}},\ }\bibfield  {title} {\bibinfo {title}
  {Combinatorial optimization with physics-inspired graph neural networks},\
  }\href {https://doi.org/10.1038/s42256-022-00468-6} {\bibfield  {journal}
  {\bibinfo  {journal} {Nature Machine Intelligence}\ }\textbf {\bibinfo
  {volume} {4}},\ \bibinfo {pages} {367} (\bibinfo {year}
  {2022}{\natexlab{a}})}\BibitemShut {NoStop}%
\bibitem [{\citenamefont {Schuetz}\ \emph
  {et~al.}(2022{\natexlab{b}})\citenamefont {Schuetz}, \citenamefont
  {Brubaker}, \citenamefont {Zhu},\ and\ \citenamefont
  {Katzgraber}}]{schuetz2022b}%
  \BibitemOpen
  \bibfield  {author} {\bibinfo {author} {\bibfnamefont {M.~J.~A.}\
  \bibnamefont {Schuetz}}, \bibinfo {author} {\bibfnamefont {J.~K.}\
  \bibnamefont {Brubaker}}, \bibinfo {author} {\bibfnamefont {Z.}~\bibnamefont
  {Zhu}},\ and\ \bibinfo {author} {\bibfnamefont {H.~G.}\ \bibnamefont
  {Katzgraber}},\ }\bibfield  {title} {\bibinfo {title} {Graph coloring with
  physics-inspired graph neural networks},\ }\href
  {https://doi.org/10.1103/PhysRevResearch.4.043131} {\bibfield  {journal}
  {\bibinfo  {journal} {Phys. Rev. Res.}\ }\textbf {\bibinfo {volume} {4}},\
  \bibinfo {pages} {043131} (\bibinfo {year} {2022}{\natexlab{b}})}\BibitemShut
  {NoStop}%
\bibitem [{\citenamefont {Heydaribeni}\ \emph {et~al.}(2024)\citenamefont
  {Heydaribeni}, \citenamefont {Zhan}, \citenamefont {Zhang}, \citenamefont
  {Eliassi-Rad},\ and\ \citenamefont {Koushanfar}}]{heydaribeni2024}%
  \BibitemOpen
  \bibfield  {author} {\bibinfo {author} {\bibfnamefont {N.}~\bibnamefont
  {Heydaribeni}}, \bibinfo {author} {\bibfnamefont {X.}~\bibnamefont {Zhan}},
  \bibinfo {author} {\bibfnamefont {R.}~\bibnamefont {Zhang}}, \bibinfo
  {author} {\bibfnamefont {T.}~\bibnamefont {Eliassi-Rad}},\ and\ \bibinfo
  {author} {\bibfnamefont {F.}~\bibnamefont {Koushanfar}},\ }\bibfield  {title}
  {\bibinfo {title} {Distributed constrained combinatorial optimization
  leveraging hypergraph neural networks},\ }\href
  {https://doi.org/10.1038/s42256-024-00833-7} {\bibfield  {journal} {\bibinfo
  {journal} {Nature Machine Intelligence}\ }\textbf {\bibinfo {volume} {6}},\
  \bibinfo {pages} {664} (\bibinfo {year} {2024})}\BibitemShut {NoStop}%
\bibitem [{\citenamefont {Pugacheva}\ \emph {et~al.}(2024)\citenamefont
  {Pugacheva}, \citenamefont {Ermakov}, \citenamefont {Lyskov}, \citenamefont
  {Makarov},\ and\ \citenamefont {Zotov}}]{pugacheva2024}%
  \BibitemOpen
  \bibfield  {author} {\bibinfo {author} {\bibfnamefont {D.}~\bibnamefont
  {Pugacheva}}, \bibinfo {author} {\bibfnamefont {A.}~\bibnamefont {Ermakov}},
  \bibinfo {author} {\bibfnamefont {I.}~\bibnamefont {Lyskov}}, \bibinfo
  {author} {\bibfnamefont {I.}~\bibnamefont {Makarov}},\ and\ \bibinfo {author}
  {\bibfnamefont {Y.}~\bibnamefont {Zotov}},\ }\href
  {https://arxiv.org/abs/2407.16468} {\bibinfo {title} {Enhancing gnns
  performance on combinatorial optimization by recurrent feature update}}
  (\bibinfo {year} {2024}),\ \Eprint {https://arxiv.org/abs/2407.16468}
  {arXiv:2407.16468 [cs.LG]} \BibitemShut {NoStop}%
\bibitem [{\citenamefont {Langedal}\ and\ \citenamefont
  {Manne}(2024)}]{langedal2024}%
  \BibitemOpen
  \bibfield  {author} {\bibinfo {author} {\bibfnamefont {K.}~\bibnamefont
  {Langedal}}\ and\ \bibinfo {author} {\bibfnamefont {F.}~\bibnamefont
  {Manne}},\ }\href {https://arxiv.org/abs/2408.05054} {\bibinfo {title} {Graph
  neural networks as ordering heuristics for parallel graph coloring}}
  (\bibinfo {year} {2024}),\ \Eprint {https://arxiv.org/abs/2408.05054}
  {arXiv:2408.05054 [cs.LG]} \BibitemShut {NoStop}%
\bibitem [{\citenamefont {Hu}(2024)}]{hu2024}%
  \BibitemOpen
  \bibfield  {author} {\bibinfo {author} {\bibfnamefont {C.}~\bibnamefont
  {Hu}},\ }\href {https://arxiv.org/abs/2411.05834} {\bibinfo {title}
  {Assessing and enhancing graph neural networks for combinatorial
  optimization: Novel approaches and application in maximum independent set
  problems}} (\bibinfo {year} {2024}),\ \Eprint
  {https://arxiv.org/abs/2411.05834} {arXiv:2411.05834 [math.OC]} \BibitemShut
  {NoStop}%
\bibitem [{\citenamefont {Angelini}\ and\ \citenamefont
  {Ricci-Tersenghi}(2023)}]{angelini2023}%
  \BibitemOpen
  \bibfield  {author} {\bibinfo {author} {\bibfnamefont {M.~C.}\ \bibnamefont
  {Angelini}}\ and\ \bibinfo {author} {\bibfnamefont {F.}~\bibnamefont
  {Ricci-Tersenghi}},\ }\bibfield  {title} {\bibinfo {title} {Modern graph
  neural networks do worse than classical greedy algorithms in solving
  combinatorial optimization problems like maximum independent set},\ }\href
  {https://doi.org/10.1038/s42256-022-00589-y} {\bibfield  {journal} {\bibinfo
  {journal} {Nature Machine Intelligence}\ }\textbf {\bibinfo {volume} {5}},\
  \bibinfo {pages} {29} (\bibinfo {year} {2023})}\BibitemShut {NoStop}%
\bibitem [{\citenamefont {Boettcher}(2023{\natexlab{a}})}]{boettcher2023}%
  \BibitemOpen
  \bibfield  {author} {\bibinfo {author} {\bibfnamefont {S.}~\bibnamefont
  {Boettcher}},\ }\bibfield  {title} {\bibinfo {title} {Inability of a graph
  neural network heuristic to outperform greedy algorithms in solving
  combinatorial optimization problems},\ }\href
  {https://doi.org/10.1038/s42256-022-00587-0} {\bibfield  {journal} {\bibinfo
  {journal} {Nature Machine Intelligence}\ }\textbf {\bibinfo {volume} {5}},\
  \bibinfo {pages} {24} (\bibinfo {year} {2023}{\natexlab{a}})}\BibitemShut
  {NoStop}%
\bibitem [{\citenamefont {Boettcher}(2023{\natexlab{b}})}]{boettcher2023a}%
  \BibitemOpen
  \bibfield  {author} {\bibinfo {author} {\bibfnamefont {S.}~\bibnamefont
  {Boettcher}},\ }\bibfield  {title} {\bibinfo {title} {Deep reinforced
  learning heuristic tested on spin-glass ground states: The larger picture},\
  }\href {https://doi.org/10.1038/s41467-023-41106-y} {\bibfield  {journal}
  {\bibinfo  {journal} {Nature Communications}\ }\textbf {\bibinfo {volume}
  {14}},\ \bibinfo {pages} {5658} (\bibinfo {year}
  {2023}{\natexlab{b}})}\BibitemShut {NoStop}%
\bibitem [{\citenamefont {Gamarnik}(2023)}]{gamarnik2023}%
  \BibitemOpen
  \bibfield  {author} {\bibinfo {author} {\bibfnamefont {D.}~\bibnamefont
  {Gamarnik}},\ }\bibfield  {title} {\bibinfo {title} {Barriers for the
  performance of graph neural networks (gnn) in discrete random structures},\
  }\href {https://doi.org/10.1073/pnas.2314092120} {\bibfield  {journal}
  {\bibinfo  {journal} {Proceedings of the National Academy of Sciences}\
  }\textbf {\bibinfo {volume} {120}},\ \bibinfo {pages} {e2314092120} (\bibinfo
  {year} {2023})},\ \Eprint
  {https://arxiv.org/abs/https://www.pnas.org/doi/pdf/10.1073/pnas.2314092120}
  {https://www.pnas.org/doi/pdf/10.1073/pnas.2314092120} \BibitemShut {NoStop}%
\bibitem [{\citenamefont {Wu}\ \emph {et~al.}(2019)\citenamefont {Wu},
  \citenamefont {Wang},\ and\ \citenamefont {Zhang}}]{wu2019}%
  \BibitemOpen
  \bibfield  {author} {\bibinfo {author} {\bibfnamefont {D.}~\bibnamefont
  {Wu}}, \bibinfo {author} {\bibfnamefont {L.}~\bibnamefont {Wang}},\ and\
  \bibinfo {author} {\bibfnamefont {P.}~\bibnamefont {Zhang}},\ }\bibfield
  {title} {\bibinfo {title} {Solving statistical mechanics using variational
  autoregressive networks},\ }\href
  {https://doi.org/10.1103/PhysRevLett.122.080602} {\bibfield  {journal}
  {\bibinfo  {journal} {Phys. Rev. Lett.}\ }\textbf {\bibinfo {volume} {122}},\
  \bibinfo {pages} {080602} (\bibinfo {year} {2019})}\BibitemShut {NoStop}%
\bibitem [{\citenamefont {Nicoli}\ \emph {et~al.}(2020)\citenamefont {Nicoli},
  \citenamefont {Nakajima}, \citenamefont {Strodthoff}, \citenamefont {Samek},
  \citenamefont {M\"uller},\ and\ \citenamefont {Kessel}}]{nicoli2020}%
  \BibitemOpen
  \bibfield  {author} {\bibinfo {author} {\bibfnamefont {K.~A.}\ \bibnamefont
  {Nicoli}}, \bibinfo {author} {\bibfnamefont {S.}~\bibnamefont {Nakajima}},
  \bibinfo {author} {\bibfnamefont {N.}~\bibnamefont {Strodthoff}}, \bibinfo
  {author} {\bibfnamefont {W.}~\bibnamefont {Samek}}, \bibinfo {author}
  {\bibfnamefont {K.-R.}\ \bibnamefont {M\"uller}},\ and\ \bibinfo {author}
  {\bibfnamefont {P.}~\bibnamefont {Kessel}},\ }\bibfield  {title} {\bibinfo
  {title} {Asymptotically unbiased estimation of physical observables with
  neural samplers},\ }\href {https://doi.org/10.1103/PhysRevE.101.023304}
  {\bibfield  {journal} {\bibinfo  {journal} {Phys. Rev. E}\ }\textbf {\bibinfo
  {volume} {101}},\ \bibinfo {pages} {023304} (\bibinfo {year}
  {2020})}\BibitemShut {NoStop}%
\bibitem [{\citenamefont {McNaughton}\ \emph {et~al.}(2020)\citenamefont
  {McNaughton}, \citenamefont {Milo\ifmmode \check{s}\else
  \v{s}\fi{}evi\ifmmode~\acute{c}\else \'{c}\fi{}}, \citenamefont {Perali},\
  and\ \citenamefont {Pilati}}]{mcnaughton2020}%
  \BibitemOpen
  \bibfield  {author} {\bibinfo {author} {\bibfnamefont {B.}~\bibnamefont
  {McNaughton}}, \bibinfo {author} {\bibfnamefont {M.~V.}\ \bibnamefont
  {Milo\ifmmode \check{s}\else \v{s}\fi{}evi\ifmmode~\acute{c}\else
  \'{c}\fi{}}}, \bibinfo {author} {\bibfnamefont {A.}~\bibnamefont {Perali}},\
  and\ \bibinfo {author} {\bibfnamefont {S.}~\bibnamefont {Pilati}},\
  }\bibfield  {title} {\bibinfo {title} {Boosting monte carlo simulations of
  spin glasses using autoregressive neural networks},\ }\href
  {https://doi.org/10.1103/PhysRevE.101.053312} {\bibfield  {journal} {\bibinfo
   {journal} {Phys. Rev. E}\ }\textbf {\bibinfo {volume} {101}},\ \bibinfo
  {pages} {053312} (\bibinfo {year} {2020})}\BibitemShut {NoStop}%
\bibitem [{\citenamefont {Wu}\ \emph {et~al.}(2021)\citenamefont {Wu},
  \citenamefont {Rossi},\ and\ \citenamefont {Carleo}}]{wu2021}%
  \BibitemOpen
  \bibfield  {author} {\bibinfo {author} {\bibfnamefont {D.}~\bibnamefont
  {Wu}}, \bibinfo {author} {\bibfnamefont {R.}~\bibnamefont {Rossi}},\ and\
  \bibinfo {author} {\bibfnamefont {G.}~\bibnamefont {Carleo}},\ }\bibfield
  {title} {\bibinfo {title} {Unbiased monte carlo cluster updates with
  autoregressive neural networks},\ }\href
  {https://doi.org/10.1103/PhysRevResearch.3.L042024} {\bibfield  {journal}
  {\bibinfo  {journal} {Phys. Rev. Res.}\ }\textbf {\bibinfo {volume} {3}},\
  \bibinfo {pages} {L042024} (\bibinfo {year} {2021})}\BibitemShut {NoStop}%
\bibitem [{\citenamefont {Hibat-Allah}\ \emph {et~al.}(2021)\citenamefont
  {Hibat-Allah}, \citenamefont {Inack}, \citenamefont {Wiersema}, \citenamefont
  {Melko},\ and\ \citenamefont {Carrasquilla}}]{hibat2021}%
  \BibitemOpen
  \bibfield  {author} {\bibinfo {author} {\bibfnamefont {M.}~\bibnamefont
  {Hibat-Allah}}, \bibinfo {author} {\bibfnamefont {E.~M.}\ \bibnamefont
  {Inack}}, \bibinfo {author} {\bibfnamefont {R.}~\bibnamefont {Wiersema}},
  \bibinfo {author} {\bibfnamefont {R.~G.}\ \bibnamefont {Melko}},\ and\
  \bibinfo {author} {\bibfnamefont {J.}~\bibnamefont {Carrasquilla}},\
  }\bibfield  {title} {\bibinfo {title} {Variational neural annealing},\ }\href
  {https://doi.org/10.1038/s42256-021-00401-3} {\bibfield  {journal} {\bibinfo
  {journal} {Nature Machine Intelligence}\ }\textbf {\bibinfo {volume} {3}},\
  \bibinfo {pages} {952} (\bibinfo {year} {2021})}\BibitemShut {NoStop}%
\bibitem [{\citenamefont {Ahsan~Khandoker}\ \emph {et~al.}(2023)\citenamefont
  {Ahsan~Khandoker}, \citenamefont {Munshad~Abedin},\ and\ \citenamefont
  {Hibat-Allah}}]{khandoker2023}%
  \BibitemOpen
  \bibfield  {author} {\bibinfo {author} {\bibfnamefont {S.}~\bibnamefont
  {Ahsan~Khandoker}}, \bibinfo {author} {\bibfnamefont {J.}~\bibnamefont
  {Munshad~Abedin}},\ and\ \bibinfo {author} {\bibfnamefont {M.}~\bibnamefont
  {Hibat-Allah}},\ }\bibfield  {title} {\bibinfo {title} {Supplementing
  recurrent neural networks with annealing to solve combinatorial optimization
  problems},\ }\href {https://doi.org/10.1088/2632-2153/acb895} {\bibfield
  {journal} {\bibinfo  {journal} {Machine Learning: Science and Technology}\
  }\textbf {\bibinfo {volume} {4}},\ \bibinfo {pages} {015026} (\bibinfo {year}
  {2023})}\BibitemShut {NoStop}%
\bibitem [{\citenamefont {Sanokowski}\ \emph {et~al.}(2023)\citenamefont
  {Sanokowski}, \citenamefont {Berghammer}, \citenamefont {Hochreiter},\ and\
  \citenamefont {Lehner}}]{sanokowski2023}%
  \BibitemOpen
  \bibfield  {author} {\bibinfo {author} {\bibfnamefont {S.}~\bibnamefont
  {Sanokowski}}, \bibinfo {author} {\bibfnamefont {W.}~\bibnamefont
  {Berghammer}}, \bibinfo {author} {\bibfnamefont {S.}~\bibnamefont
  {Hochreiter}},\ and\ \bibinfo {author} {\bibfnamefont {S.}~\bibnamefont
  {Lehner}},\ }\bibfield  {title} {\bibinfo {title} {Variational annealing on
  graphs for combinatorial optimization},\ }in\ \href
  {https://proceedings.neurips.cc/paper_files/paper/2023/file/c9c54ac0dd5e942b99b2b51c297544fd-Paper-Conference.pdf}
  {\emph {\bibinfo {booktitle} {Advances in Neural Information Processing
  Systems}}},\ Vol.~\bibinfo {volume} {36},\ \bibinfo {editor} {edited by\
  \bibinfo {editor} {\bibfnamefont {A.}~\bibnamefont {Oh}}, \bibinfo {editor}
  {\bibfnamefont {T.}~\bibnamefont {Naumann}}, \bibinfo {editor} {\bibfnamefont
  {A.}~\bibnamefont {Globerson}}, \bibinfo {editor} {\bibfnamefont
  {K.}~\bibnamefont {Saenko}}, \bibinfo {editor} {\bibfnamefont
  {M.}~\bibnamefont {Hardt}},\ and\ \bibinfo {editor} {\bibfnamefont
  {S.}~\bibnamefont {Levine}}}\ (\bibinfo  {publisher} {Curran Associates,
  Inc.},\ \bibinfo {year} {2023})\ pp.\ \bibinfo {pages}
  {63907--63930}\BibitemShut {NoStop}%
\bibitem [{\citenamefont {Ma}\ \emph {et~al.}(2024)\citenamefont {Ma},
  \citenamefont {Ma}, \citenamefont {Xu}, \citenamefont {Zhang},\ and\
  \citenamefont {Gao}}]{ma2024}%
  \BibitemOpen
  \bibfield  {author} {\bibinfo {author} {\bibfnamefont {Q.}~\bibnamefont
  {Ma}}, \bibinfo {author} {\bibfnamefont {Z.}~\bibnamefont {Ma}}, \bibinfo
  {author} {\bibfnamefont {J.}~\bibnamefont {Xu}}, \bibinfo {author}
  {\bibfnamefont {H.}~\bibnamefont {Zhang}},\ and\ \bibinfo {author}
  {\bibfnamefont {M.}~\bibnamefont {Gao}},\ }\bibfield  {title} {\bibinfo
  {title} {Message passing variational autoregressive network for solving
  intractable ising models},\ }\href
  {https://doi.org/10.1038/s42005-024-01711-9} {\bibfield  {journal} {\bibinfo
  {journal} {Communications Physics}\ }\textbf {\bibinfo {volume} {7}},\
  \bibinfo {pages} {236} (\bibinfo {year} {2024})}\BibitemShut {NoStop}%
\bibitem [{\citenamefont {Pan}\ \emph {et~al.}(2021)\citenamefont {Pan},
  \citenamefont {Zhou}, \citenamefont {Zhou},\ and\ \citenamefont
  {Zhang}}]{pan2021}%
  \BibitemOpen
  \bibfield  {author} {\bibinfo {author} {\bibfnamefont {F.}~\bibnamefont
  {Pan}}, \bibinfo {author} {\bibfnamefont {P.}~\bibnamefont {Zhou}}, \bibinfo
  {author} {\bibfnamefont {H.-J.}\ \bibnamefont {Zhou}},\ and\ \bibinfo
  {author} {\bibfnamefont {P.}~\bibnamefont {Zhang}},\ }\bibfield  {title}
  {\bibinfo {title} {Solving statistical mechanics on sparse graphs with
  feedback-set variational autoregressive networks},\ }\href
  {https://doi.org/10.1103/PhysRevE.103.012103} {\bibfield  {journal} {\bibinfo
   {journal} {Phys. Rev. E}\ }\textbf {\bibinfo {volume} {103}},\ \bibinfo
  {pages} {012103} (\bibinfo {year} {2021})}\BibitemShut {NoStop}%
\bibitem [{\citenamefont {Biazzo}(2023)}]{biazzo2023}%
  \BibitemOpen
  \bibfield  {author} {\bibinfo {author} {\bibfnamefont {I.}~\bibnamefont
  {Biazzo}},\ }\bibfield  {title} {\bibinfo {title} {The autoregressive neural
  network architecture of the boltzmann distribution of pairwise interacting
  spins systems},\ }\href {https://doi.org/10.1038/s42005-023-01416-5}
  {\bibfield  {journal} {\bibinfo  {journal} {Communications Physics}\ }\textbf
  {\bibinfo {volume} {6}},\ \bibinfo {pages} {296} (\bibinfo {year}
  {2023})}\BibitemShut {NoStop}%
\bibitem [{\citenamefont {Biazzo}\ \emph {et~al.}(2024)\citenamefont {Biazzo},
  \citenamefont {Wu},\ and\ \citenamefont {Carleo}}]{biazzo2024}%
  \BibitemOpen
  \bibfield  {author} {\bibinfo {author} {\bibfnamefont {I.}~\bibnamefont
  {Biazzo}}, \bibinfo {author} {\bibfnamefont {D.}~\bibnamefont {Wu}},\ and\
  \bibinfo {author} {\bibfnamefont {G.}~\bibnamefont {Carleo}},\ }\bibfield
  {title} {\bibinfo {title} {Sparse autoregressive neural networks for
  classical spin systems},\ }\href {https://doi.org/10.1088/2632-2153/ad5783}
  {\bibfield  {journal} {\bibinfo  {journal} {Machine Learning: Science and
  Technology}\ }\textbf {\bibinfo {volume} {5}},\ \bibinfo {pages} {025074}
  (\bibinfo {year} {2024})}\BibitemShut {NoStop}%
\bibitem [{\citenamefont {Del~Bono}\ \emph {et~al.}(2025)\citenamefont
  {Del~Bono}, \citenamefont {Ricci-Tersenghi},\ and\ \citenamefont
  {Zamponi}}]{delbono2025a}%
  \BibitemOpen
  \bibfield  {author} {\bibinfo {author} {\bibfnamefont {L.~M.}\ \bibnamefont
  {Del~Bono}}, \bibinfo {author} {\bibfnamefont {F.}~\bibnamefont
  {Ricci-Tersenghi}},\ and\ \bibinfo {author} {\bibfnamefont {F.}~\bibnamefont
  {Zamponi}},\ }\bibfield  {title} {\bibinfo {title} {Nearest-neighbors neural
  network architecture for efficient sampling of statistical physics models},\
  }\href {https://doi.org/10.1088/2632-2153/adcdc1} {\bibfield  {journal}
  {\bibinfo  {journal} {Machine Learning: Science and Technology}\ }\textbf
  {\bibinfo {volume} {6}},\ \bibinfo {pages} {025029} (\bibinfo {year}
  {2025})}\BibitemShut {NoStop}%
\bibitem [{\citenamefont {Ciarella}\ \emph {et~al.}(2023)\citenamefont
  {Ciarella}, \citenamefont {Trinquier}, \citenamefont {Weigt},\ and\
  \citenamefont {Zamponi}}]{ciarella2023}%
  \BibitemOpen
  \bibfield  {author} {\bibinfo {author} {\bibfnamefont {S.}~\bibnamefont
  {Ciarella}}, \bibinfo {author} {\bibfnamefont {J.}~\bibnamefont {Trinquier}},
  \bibinfo {author} {\bibfnamefont {M.}~\bibnamefont {Weigt}},\ and\ \bibinfo
  {author} {\bibfnamefont {F.}~\bibnamefont {Zamponi}},\ }\bibfield  {title}
  {\bibinfo {title} {Machine-learning-assisted monte carlo fails at sampling
  computationally hard problems},\ }\href
  {https://doi.org/10.1088/2632-2153/acbe91} {\bibfield  {journal} {\bibinfo
  {journal} {Machine Learning: Science and Technology}\ }\textbf {\bibinfo
  {volume} {4}},\ \bibinfo {pages} {010501} (\bibinfo {year}
  {2023})}\BibitemShut {NoStop}%
\bibitem [{\citenamefont {Ghio}\ \emph {et~al.}(2024)\citenamefont {Ghio},
  \citenamefont {Dandi}, \citenamefont {Krzakala},\ and\ \citenamefont
  {Zdeborov{\'a}}}]{ghio2024}%
  \BibitemOpen
  \bibfield  {author} {\bibinfo {author} {\bibfnamefont {D.}~\bibnamefont
  {Ghio}}, \bibinfo {author} {\bibfnamefont {Y.}~\bibnamefont {Dandi}},
  \bibinfo {author} {\bibfnamefont {F.}~\bibnamefont {Krzakala}},\ and\
  \bibinfo {author} {\bibfnamefont {L.}~\bibnamefont {Zdeborov{\'a}}},\
  }\bibfield  {title} {\bibinfo {title} {Sampling with flows, diffusion, and
  autoregressive neural networks from a spin-glass perspective},\ }\href
  {https://doi.org/10.1073/pnas.2311810121} {\bibfield  {journal} {\bibinfo
  {journal} {Proceedings of the National Academy of Sciences}\ }\textbf
  {\bibinfo {volume} {121}},\ \bibinfo {pages} {e2311810121} (\bibinfo {year}
  {2024})},\ \Eprint
  {https://arxiv.org/abs/https://www.pnas.org/doi/pdf/10.1073/pnas.2311810121}
  {https://www.pnas.org/doi/pdf/10.1073/pnas.2311810121} \BibitemShut {NoStop}%
\bibitem [{\citenamefont {Bono}\ \emph {et~al.}(2025)\citenamefont {Bono},
  \citenamefont {Ricci-Tersenghi},\ and\ \citenamefont
  {Zamponi}}]{delbono2025b}%
  \BibitemOpen
  \bibfield  {author} {\bibinfo {author} {\bibfnamefont {L.~M.~D.}\
  \bibnamefont {Bono}}, \bibinfo {author} {\bibfnamefont {F.}~\bibnamefont
  {Ricci-Tersenghi}},\ and\ \bibinfo {author} {\bibfnamefont {F.}~\bibnamefont
  {Zamponi}},\ }\href {https://arxiv.org/abs/2505.22598} {\bibinfo {title} {On
  the performance of machine-learning-assisted monte carlo in sampling from
  simple statistical physics models}} (\bibinfo {year} {2025}),\ \Eprint
  {https://arxiv.org/abs/2505.22598} {arXiv:2505.22598 [cond-mat.dis-nn]}
  \BibitemShut {NoStop}%
\bibitem [{\citenamefont {Garey}\ and\ \citenamefont
  {Johnson}(1990)}]{garey1990}%
  \BibitemOpen
  \bibfield  {author} {\bibinfo {author} {\bibfnamefont {M.~R.}\ \bibnamefont
  {Garey}}\ and\ \bibinfo {author} {\bibfnamefont {D.~S.}\ \bibnamefont
  {Johnson}},\ }\href@noop {} {\emph {\bibinfo {title} {Computers and
  Intractability; A Guide to the Theory of NP-Completeness}}}\ (\bibinfo
  {publisher} {W. H. Freeman \& Co.},\ \bibinfo {address} {USA},\ \bibinfo
  {year} {1990})\BibitemShut {NoStop}%
\bibitem [{\citenamefont {Bybee}\ \emph {et~al.}(2023)\citenamefont {Bybee},
  \citenamefont {Kleyko}, \citenamefont {Nikonov}, \citenamefont
  {Khosrowshahi}, \citenamefont {Olshausen},\ and\ \citenamefont
  {Sommer}}]{bybee2023}%
  \BibitemOpen
  \bibfield  {author} {\bibinfo {author} {\bibfnamefont {C.}~\bibnamefont
  {Bybee}}, \bibinfo {author} {\bibfnamefont {D.}~\bibnamefont {Kleyko}},
  \bibinfo {author} {\bibfnamefont {D.~E.}\ \bibnamefont {Nikonov}}, \bibinfo
  {author} {\bibfnamefont {A.}~\bibnamefont {Khosrowshahi}}, \bibinfo {author}
  {\bibfnamefont {B.~A.}\ \bibnamefont {Olshausen}},\ and\ \bibinfo {author}
  {\bibfnamefont {F.~T.}\ \bibnamefont {Sommer}},\ }\bibfield  {title}
  {\bibinfo {title} {Efficient optimization with higher-order ising machines},\
  }\href {https://doi.org/10.1038/s41467-023-41214-9} {\bibfield  {journal}
  {\bibinfo  {journal} {Nature Communications}\ }\textbf {\bibinfo {volume}
  {14}},\ \bibinfo {pages} {6033} (\bibinfo {year} {2023})}\BibitemShut
  {NoStop}%
\bibitem [{\citenamefont {Sharma}\ \emph {et~al.}(2023)\citenamefont {Sharma},
  \citenamefont {Burns}, \citenamefont {Hahn},\ and\ \citenamefont
  {Huang}}]{sharma2023}%
  \BibitemOpen
  \bibfield  {author} {\bibinfo {author} {\bibfnamefont {A.}~\bibnamefont
  {Sharma}}, \bibinfo {author} {\bibfnamefont {M.}~\bibnamefont {Burns}},
  \bibinfo {author} {\bibfnamefont {A.}~\bibnamefont {Hahn}},\ and\ \bibinfo
  {author} {\bibfnamefont {M.}~\bibnamefont {Huang}},\ }\bibfield  {title}
  {\bibinfo {title} {Augmenting an electronic ising machine to effectively
  solve boolean satisfiability},\ }\href
  {https://doi.org/10.1038/s41598-023-49966-6} {\bibfield  {journal} {\bibinfo
  {journal} {Scientific Reports}\ }\textbf {\bibinfo {volume} {13}},\ \bibinfo
  {pages} {22858} (\bibinfo {year} {2023})}\BibitemShut {NoStop}%
\bibitem [{\citenamefont {Bhattacharya}\ \emph {et~al.}(2024)\citenamefont
  {Bhattacharya}, \citenamefont {Hutchinson}, \citenamefont {Pedretti},
  \citenamefont {Sheng}, \citenamefont {Ignowski}, \citenamefont
  {Van~Vaerenbergh}, \citenamefont {Beausoleil}, \citenamefont {Strachan},\
  and\ \citenamefont {Strukov}}]{bhattacharya2024}%
  \BibitemOpen
  \bibfield  {author} {\bibinfo {author} {\bibfnamefont {T.}~\bibnamefont
  {Bhattacharya}}, \bibinfo {author} {\bibfnamefont {G.~H.}\ \bibnamefont
  {Hutchinson}}, \bibinfo {author} {\bibfnamefont {G.}~\bibnamefont
  {Pedretti}}, \bibinfo {author} {\bibfnamefont {X.}~\bibnamefont {Sheng}},
  \bibinfo {author} {\bibfnamefont {J.}~\bibnamefont {Ignowski}}, \bibinfo
  {author} {\bibfnamefont {T.}~\bibnamefont {Van~Vaerenbergh}}, \bibinfo
  {author} {\bibfnamefont {R.}~\bibnamefont {Beausoleil}}, \bibinfo {author}
  {\bibfnamefont {J.~P.}\ \bibnamefont {Strachan}},\ and\ \bibinfo {author}
  {\bibfnamefont {D.~B.}\ \bibnamefont {Strukov}},\ }\bibfield  {title}
  {\bibinfo {title} {Computing high-degree polynomial gradients in memory},\
  }\href {https://doi.org/10.1038/s41467-024-52488-y} {\bibfield  {journal}
  {\bibinfo  {journal} {Nature Communications}\ }\textbf {\bibinfo {volume}
  {15}},\ \bibinfo {pages} {8211} (\bibinfo {year} {2024})}\BibitemShut
  {NoStop}%
\bibitem [{\citenamefont {Nikhar}\ \emph {et~al.}(2024)\citenamefont {Nikhar},
  \citenamefont {Kannan}, \citenamefont {Aadit}, \citenamefont {Chowdhury},\
  and\ \citenamefont {Camsari}}]{nikhar2024}%
  \BibitemOpen
  \bibfield  {author} {\bibinfo {author} {\bibfnamefont {S.}~\bibnamefont
  {Nikhar}}, \bibinfo {author} {\bibfnamefont {S.}~\bibnamefont {Kannan}},
  \bibinfo {author} {\bibfnamefont {N.~A.}\ \bibnamefont {Aadit}}, \bibinfo
  {author} {\bibfnamefont {S.}~\bibnamefont {Chowdhury}},\ and\ \bibinfo
  {author} {\bibfnamefont {K.~Y.}\ \bibnamefont {Camsari}},\ }\bibfield
  {title} {\bibinfo {title} {All-to-all reconfigurability with sparse and
  higher-order ising machines},\ }\href
  {https://doi.org/10.1038/s41467-024-53270-w} {\bibfield  {journal} {\bibinfo
  {journal} {Nature Communications}\ }\textbf {\bibinfo {volume} {15}},\
  \bibinfo {pages} {8977} (\bibinfo {year} {2024})}\BibitemShut {NoStop}%
\bibitem [{\citenamefont {Pedretti}\ \emph {et~al.}(2025)\citenamefont
  {Pedretti}, \citenamefont {B{\"o}hm}, \citenamefont {Bhattacharya},
  \citenamefont {Heittmann}, \citenamefont {Zhang}, \citenamefont {Hizzani},
  \citenamefont {Hutchinson}, \citenamefont {Kwon}, \citenamefont {Moon},
  \citenamefont {Valiante}, \citenamefont {Rozada}, \citenamefont {Graves},
  \citenamefont {Ignowski}, \citenamefont {Mohseni}, \citenamefont {Strachan},
  \citenamefont {Strukov}, \citenamefont {Beausoleil},\ and\ \citenamefont
  {Van~Vaerenbergh}}]{pedretti2025}%
  \BibitemOpen
  \bibfield  {author} {\bibinfo {author} {\bibfnamefont {G.}~\bibnamefont
  {Pedretti}}, \bibinfo {author} {\bibfnamefont {F.}~\bibnamefont {B{\"o}hm}},
  \bibinfo {author} {\bibfnamefont {T.}~\bibnamefont {Bhattacharya}}, \bibinfo
  {author} {\bibfnamefont {A.}~\bibnamefont {Heittmann}}, \bibinfo {author}
  {\bibfnamefont {X.}~\bibnamefont {Zhang}}, \bibinfo {author} {\bibfnamefont
  {M.}~\bibnamefont {Hizzani}}, \bibinfo {author} {\bibfnamefont
  {G.}~\bibnamefont {Hutchinson}}, \bibinfo {author} {\bibfnamefont
  {D.}~\bibnamefont {Kwon}}, \bibinfo {author} {\bibfnamefont {J.}~\bibnamefont
  {Moon}}, \bibinfo {author} {\bibfnamefont {E.}~\bibnamefont {Valiante}},
  \bibinfo {author} {\bibfnamefont {I.}~\bibnamefont {Rozada}}, \bibinfo
  {author} {\bibfnamefont {C.~E.}\ \bibnamefont {Graves}}, \bibinfo {author}
  {\bibfnamefont {J.}~\bibnamefont {Ignowski}}, \bibinfo {author}
  {\bibfnamefont {M.}~\bibnamefont {Mohseni}}, \bibinfo {author} {\bibfnamefont
  {J.~P.}\ \bibnamefont {Strachan}}, \bibinfo {author} {\bibfnamefont
  {D.}~\bibnamefont {Strukov}}, \bibinfo {author} {\bibfnamefont
  {R.}~\bibnamefont {Beausoleil}},\ and\ \bibinfo {author} {\bibfnamefont
  {T.}~\bibnamefont {Van~Vaerenbergh}},\ }\bibfield  {title} {\bibinfo {title}
  {Solving boolean satisfiability problems with resistive content addressable
  memories},\ }\href {https://doi.org/10.1038/s44335-025-00020-w} {\bibfield
  {journal} {\bibinfo  {journal} {npj Unconventional Computing}\ }\textbf
  {\bibinfo {volume} {2}},\ \bibinfo {pages} {7} (\bibinfo {year}
  {2025})}\BibitemShut {NoStop}%
\bibitem [{\citenamefont {Chen}\ \emph {et~al.}(2019)\citenamefont {Chen},
  \citenamefont {Gamarnik}, \citenamefont {Panchenko},\ and\ \citenamefont
  {Rahman}}]{chen2019}%
  \BibitemOpen
  \bibfield  {author} {\bibinfo {author} {\bibfnamefont {W.-K.}\ \bibnamefont
  {Chen}}, \bibinfo {author} {\bibfnamefont {D.}~\bibnamefont {Gamarnik}},
  \bibinfo {author} {\bibfnamefont {D.}~\bibnamefont {Panchenko}},\ and\
  \bibinfo {author} {\bibfnamefont {M.}~\bibnamefont {Rahman}},\ }\bibfield
  {title} {\bibinfo {title} {Suboptimality of local algorithms for a class of
  max-cut problems},\ }\href {https://doi.org/10.1214/18-AOP1291} {\bibfield
  {journal} {\bibinfo  {journal} {The Annals of Probability}\ }\textbf
  {\bibinfo {volume} {47}},\ \bibinfo {pages} {1587} (\bibinfo {year}
  {2019})}\BibitemShut {NoStop}%
\bibitem [{\citenamefont {Marino}\ \emph {et~al.}(2016)\citenamefont {Marino},
  \citenamefont {Parisi},\ and\ \citenamefont {Ricci-Tersenghi}}]{marino2016}%
  \BibitemOpen
  \bibfield  {author} {\bibinfo {author} {\bibfnamefont {R.}~\bibnamefont
  {Marino}}, \bibinfo {author} {\bibfnamefont {G.}~\bibnamefont {Parisi}},\
  and\ \bibinfo {author} {\bibfnamefont {F.}~\bibnamefont {Ricci-Tersenghi}},\
  }\bibfield  {title} {\bibinfo {title} {The backtracking survey propagation
  algorithm for solving random k-sat problems},\ }\href
  {https://doi.org/10.1038/ncomms12996} {\bibfield  {journal} {\bibinfo
  {journal} {Nature Communications}\ }\textbf {\bibinfo {volume} {7}},\
  \bibinfo {pages} {12996} (\bibinfo {year} {2016})}\BibitemShut {NoStop}%
\bibitem [{\citenamefont {Gamarnik}\ \emph {et~al.}(2022)\citenamefont
  {Gamarnik}, \citenamefont {Moore},\ and\ \citenamefont
  {Zdeborov{\'a}}}]{gamarnik2022}%
  \BibitemOpen
  \bibfield  {author} {\bibinfo {author} {\bibfnamefont {D.}~\bibnamefont
  {Gamarnik}}, \bibinfo {author} {\bibfnamefont {C.}~\bibnamefont {Moore}},\
  and\ \bibinfo {author} {\bibfnamefont {L.}~\bibnamefont {Zdeborov{\'a}}},\
  }\bibfield  {title} {\bibinfo {title} {Disordered systems insights on
  computational hardness},\ }\href {https://doi.org/10.1088/1742-5468/ac9cc8}
  {\bibfield  {journal} {\bibinfo  {journal} {Journal of Statistical Mechanics:
  Theory and Experiment}\ }\textbf {\bibinfo {volume} {2022}},\ \bibinfo
  {pages} {114015} (\bibinfo {year} {2022})}\BibitemShut {NoStop}%
\bibitem [{\citenamefont {Karimi}\ \emph {et~al.}(2017)\citenamefont {Karimi},
  \citenamefont {Rosenberg},\ and\ \citenamefont {Katzgraber}}]{karimi2017}%
  \BibitemOpen
  \bibfield  {author} {\bibinfo {author} {\bibfnamefont {H.}~\bibnamefont
  {Karimi}}, \bibinfo {author} {\bibfnamefont {G.}~\bibnamefont {Rosenberg}},\
  and\ \bibinfo {author} {\bibfnamefont {H.~G.}\ \bibnamefont {Katzgraber}},\
  }\bibfield  {title} {\bibinfo {title} {Effective optimization using sample
  persistence: A case study on quantum annealers and various monte carlo
  optimization methods},\ }\href {https://doi.org/10.1103/PhysRevE.96.043312}
  {\bibfield  {journal} {\bibinfo  {journal} {Phys. Rev. E}\ }\textbf {\bibinfo
  {volume} {96}},\ \bibinfo {pages} {043312} (\bibinfo {year}
  {2017})}\BibitemShut {NoStop}%
\bibitem [{\citenamefont {Bravyi}\ \emph {et~al.}(2020)\citenamefont {Bravyi},
  \citenamefont {Kliesch}, \citenamefont {Koenig},\ and\ \citenamefont
  {Tang}}]{bravyi2020}%
  \BibitemOpen
  \bibfield  {author} {\bibinfo {author} {\bibfnamefont {S.}~\bibnamefont
  {Bravyi}}, \bibinfo {author} {\bibfnamefont {A.}~\bibnamefont {Kliesch}},
  \bibinfo {author} {\bibfnamefont {R.}~\bibnamefont {Koenig}},\ and\ \bibinfo
  {author} {\bibfnamefont {E.}~\bibnamefont {Tang}},\ }\bibfield  {title}
  {\bibinfo {title} {Obstacles to variational quantum optimization from
  symmetry protection},\ }\href
  {https://doi.org/10.1103/PhysRevLett.125.260505} {\bibfield  {journal}
  {\bibinfo  {journal} {Phys. Rev. Lett.}\ }\textbf {\bibinfo {volume} {125}},\
  \bibinfo {pages} {260505} (\bibinfo {year} {2020})}\BibitemShut {NoStop}%
\bibitem [{\citenamefont {Bravyi}\ \emph {et~al.}(2022)\citenamefont {Bravyi},
  \citenamefont {Kliesch}, \citenamefont {Koenig},\ and\ \citenamefont
  {Tang}}]{bravyi2022}%
  \BibitemOpen
  \bibfield  {author} {\bibinfo {author} {\bibfnamefont {S.}~\bibnamefont
  {Bravyi}}, \bibinfo {author} {\bibfnamefont {A.}~\bibnamefont {Kliesch}},
  \bibinfo {author} {\bibfnamefont {R.}~\bibnamefont {Koenig}},\ and\ \bibinfo
  {author} {\bibfnamefont {E.}~\bibnamefont {Tang}},\ }\bibfield  {title}
  {\bibinfo {title} {Hybrid quantum-classical algorithms for approximate graph
  coloring},\ }\href {https://doi.org/10.22331/q-2022-03-30-678} {\bibfield
  {journal} {\bibinfo  {journal} {{Quantum}}\ }\textbf {\bibinfo {volume}
  {6}},\ \bibinfo {pages} {678} (\bibinfo {year} {2022})}\BibitemShut {NoStop}%
\bibitem [{\citenamefont {Fin\ifmmode~\check{z}\else \v{z}\fi{}gar}\ \emph
  {et~al.}(2024)\citenamefont {Fin\ifmmode~\check{z}\else \v{z}\fi{}gar},
  \citenamefont {Kerschbaumer}, \citenamefont {Schuetz}, \citenamefont
  {Mendl},\ and\ \citenamefont {Katzgraber}}]{finzgar2024}%
  \BibitemOpen
  \bibfield  {author} {\bibinfo {author} {\bibfnamefont {J.~R.}\ \bibnamefont
  {Fin\ifmmode~\check{z}\else \v{z}\fi{}gar}}, \bibinfo {author} {\bibfnamefont
  {A.}~\bibnamefont {Kerschbaumer}}, \bibinfo {author} {\bibfnamefont {M.~J.}\
  \bibnamefont {Schuetz}}, \bibinfo {author} {\bibfnamefont {C.~B.}\
  \bibnamefont {Mendl}},\ and\ \bibinfo {author} {\bibfnamefont {H.~G.}\
  \bibnamefont {Katzgraber}},\ }\bibfield  {title} {\bibinfo {title}
  {Quantum-informed recursive optimization algorithms},\ }\href
  {https://doi.org/10.1103/PRXQuantum.5.020327} {\bibfield  {journal} {\bibinfo
   {journal} {PRX Quantum}\ }\textbf {\bibinfo {volume} {5}},\ \bibinfo {pages}
  {020327} (\bibinfo {year} {2024})}\BibitemShut {NoStop}%
\bibitem [{\citenamefont {Bradbury}\ \emph {et~al.}(2018)\citenamefont
  {Bradbury}, \citenamefont {Frostig}, \citenamefont {Hawkins}, \citenamefont
  {Johnson}, \citenamefont {Leary}, \citenamefont {Maclaurin}, \citenamefont
  {Necula}, \citenamefont {Paszke}, \citenamefont {Vander{P}las}, \citenamefont
  {Wanderman-{M}ilne},\ and\ \citenamefont {Zhang}}]{jax2018github}%
  \BibitemOpen
  \bibfield  {author} {\bibinfo {author} {\bibfnamefont {J.}~\bibnamefont
  {Bradbury}}, \bibinfo {author} {\bibfnamefont {R.}~\bibnamefont {Frostig}},
  \bibinfo {author} {\bibfnamefont {P.}~\bibnamefont {Hawkins}}, \bibinfo
  {author} {\bibfnamefont {M.~J.}\ \bibnamefont {Johnson}}, \bibinfo {author}
  {\bibfnamefont {C.}~\bibnamefont {Leary}}, \bibinfo {author} {\bibfnamefont
  {D.}~\bibnamefont {Maclaurin}}, \bibinfo {author} {\bibfnamefont
  {G.}~\bibnamefont {Necula}}, \bibinfo {author} {\bibfnamefont
  {A.}~\bibnamefont {Paszke}}, \bibinfo {author} {\bibfnamefont
  {J.}~\bibnamefont {Vander{P}las}}, \bibinfo {author} {\bibfnamefont
  {S.}~\bibnamefont {Wanderman-{M}ilne}},\ and\ \bibinfo {author}
  {\bibfnamefont {Q.}~\bibnamefont {Zhang}},\ }\href
  {http://github.com/jax-ml/jax} {\bibinfo {title} {{JAX}: composable
  transformations of {P}ython+{N}um{P}y programs}} (\bibinfo {year}
  {2018})\BibitemShut {NoStop}%
\bibitem [{\citenamefont {Angelini}\ \emph {et~al.}(2025)\citenamefont
  {Angelini}, \citenamefont {Avila-Gonz{\'a}lez}, \citenamefont {D'Amico},
  \citenamefont {Machado}, \citenamefont {Mulet},\ and\ \citenamefont
  {Ricci-Tersenghi}}]{angelini2025}%
  \BibitemOpen
  \bibfield  {author} {\bibinfo {author} {\bibfnamefont {M.~C.}\ \bibnamefont
  {Angelini}}, \bibinfo {author} {\bibfnamefont {M.}~\bibnamefont
  {Avila-Gonz{\'a}lez}}, \bibinfo {author} {\bibfnamefont {F.}~\bibnamefont
  {D'Amico}}, \bibinfo {author} {\bibfnamefont {D.}~\bibnamefont {Machado}},
  \bibinfo {author} {\bibfnamefont {R.}~\bibnamefont {Mulet}},\ and\ \bibinfo
  {author} {\bibfnamefont {F.}~\bibnamefont {Ricci-Tersenghi}},\ }\href
  {https://arxiv.org/abs/2504.11174} {\bibinfo {title} {Algorithmic thresholds
  in combinatorial optimization depend on the time scaling}} (\bibinfo {year}
  {2025}),\ \Eprint {https://arxiv.org/abs/2504.11174} {arXiv:2504.11174
  [cond-mat.dis-nn]} \BibitemShut {NoStop}%
\bibitem [{\citenamefont {Biere}\ \emph {et~al.}(2009)\citenamefont {Biere},
  \citenamefont {Biere}, \citenamefont {Heule}, \citenamefont {van Maaren},\
  and\ \citenamefont {Walsh}}]{biere2009}%
  \BibitemOpen
  \bibfield  {author} {\bibinfo {author} {\bibfnamefont {A.}~\bibnamefont
  {Biere}}, \bibinfo {author} {\bibfnamefont {A.}~\bibnamefont {Biere}},
  \bibinfo {author} {\bibfnamefont {M.}~\bibnamefont {Heule}}, \bibinfo
  {author} {\bibfnamefont {H.}~\bibnamefont {van Maaren}},\ and\ \bibinfo
  {author} {\bibfnamefont {T.}~\bibnamefont {Walsh}},\ }\href@noop {} {\emph
  {\bibinfo {title} {Handbook of Satisfiability: Volume 185 Frontiers in
  Artificial Intelligence and Applications}}}\ (\bibinfo  {publisher} {IOS
  Press},\ \bibinfo {address} {NLD},\ \bibinfo {year} {2009})\BibitemShut
  {NoStop}%
\bibitem [{\citenamefont {Huembeli}\ \emph {et~al.}(2022)\citenamefont
  {Huembeli}, \citenamefont {Arrazola}, \citenamefont {Killoran}, \citenamefont
  {Mohseni},\ and\ \citenamefont {Wittek}}]{huembeli2022}%
  \BibitemOpen
  \bibfield  {author} {\bibinfo {author} {\bibfnamefont {P.}~\bibnamefont
  {Huembeli}}, \bibinfo {author} {\bibfnamefont {J.~M.}\ \bibnamefont
  {Arrazola}}, \bibinfo {author} {\bibfnamefont {N.}~\bibnamefont {Killoran}},
  \bibinfo {author} {\bibfnamefont {M.}~\bibnamefont {Mohseni}},\ and\ \bibinfo
  {author} {\bibfnamefont {P.}~\bibnamefont {Wittek}},\ }\bibfield  {title}
  {\bibinfo {title} {The physics of energy-based models},\ }\href
  {https://doi.org/10.1007/s42484-021-00057-7} {\bibfield  {journal} {\bibinfo
  {journal} {Quantum Machine Intelligence}\ }\textbf {\bibinfo {volume} {4}},\
  \bibinfo {pages} {1} (\bibinfo {year} {2022})}\BibitemShut {NoStop}%
\bibitem [{\citenamefont {Niazi}\ \emph {et~al.}(2024)\citenamefont {Niazi},
  \citenamefont {Chowdhury}, \citenamefont {Aadit}, \citenamefont {Mohseni},
  \citenamefont {Qin},\ and\ \citenamefont {Camsari}}]{niazi2024training}%
  \BibitemOpen
  \bibfield  {author} {\bibinfo {author} {\bibfnamefont {S.}~\bibnamefont
  {Niazi}}, \bibinfo {author} {\bibfnamefont {S.}~\bibnamefont {Chowdhury}},
  \bibinfo {author} {\bibfnamefont {N.~A.}\ \bibnamefont {Aadit}}, \bibinfo
  {author} {\bibfnamefont {M.}~\bibnamefont {Mohseni}}, \bibinfo {author}
  {\bibfnamefont {Y.}~\bibnamefont {Qin}},\ and\ \bibinfo {author}
  {\bibfnamefont {K.~Y.}\ \bibnamefont {Camsari}},\ }\bibfield  {title}
  {\bibinfo {title} {Training deep boltzmann networks with sparse ising
  machines},\ }\href {https://doi.org/10.1038/s41928-024-01182-4} {\bibfield
  {journal} {\bibinfo  {journal} {Nature Electronics}\ }\textbf {\bibinfo
  {volume} {7}},\ \bibinfo {pages} {610} (\bibinfo {year} {2024})}\BibitemShut
  {NoStop}%
\bibitem [{\citenamefont {Chowdhury}\ \emph {et~al.}(2023)\citenamefont
  {Chowdhury}, \citenamefont {Grimaldi}, \citenamefont {Aadit}, \citenamefont
  {Niazi}, \citenamefont {Mohseni}, \citenamefont {Kanai}, \citenamefont
  {Ohno}, \citenamefont {Fukami}, \citenamefont {Theogarajan}, \citenamefont
  {Finocchio}, \citenamefont {Datta},\ and\ \citenamefont
  {Camsari}}]{chowdhury2023fullstack}%
  \BibitemOpen
  \bibfield  {author} {\bibinfo {author} {\bibfnamefont {S.}~\bibnamefont
  {Chowdhury}}, \bibinfo {author} {\bibfnamefont {A.}~\bibnamefont {Grimaldi}},
  \bibinfo {author} {\bibfnamefont {N.~A.}\ \bibnamefont {Aadit}}, \bibinfo
  {author} {\bibfnamefont {S.}~\bibnamefont {Niazi}}, \bibinfo {author}
  {\bibfnamefont {M.}~\bibnamefont {Mohseni}}, \bibinfo {author} {\bibfnamefont
  {S.}~\bibnamefont {Kanai}}, \bibinfo {author} {\bibfnamefont
  {H.}~\bibnamefont {Ohno}}, \bibinfo {author} {\bibfnamefont {S.}~\bibnamefont
  {Fukami}}, \bibinfo {author} {\bibfnamefont {L.}~\bibnamefont {Theogarajan}},
  \bibinfo {author} {\bibfnamefont {G.}~\bibnamefont {Finocchio}}, \bibinfo
  {author} {\bibfnamefont {S.}~\bibnamefont {Datta}},\ and\ \bibinfo {author}
  {\bibfnamefont {K.~Y.}\ \bibnamefont {Camsari}},\ }\bibfield  {title}
  {\bibinfo {title} {A full-stack view of probabilistic computing with p-bits:
  Devices, architectures, and algorithms},\ }\href
  {https://doi.org/10.1109/JXCDC.2023.3256981} {\bibfield  {journal} {\bibinfo
  {journal} {IEEE Journal on Exploratory Solid-State Computational Devices and
  Circuits}\ }\textbf {\bibinfo {volume} {9}},\ \bibinfo {pages} {1} (\bibinfo
  {year} {2023})}\BibitemShut {NoStop}%
\bibitem [{\citenamefont {Aadit}\ \emph {et~al.}(2023)\citenamefont {Aadit},
  \citenamefont {Mohseni},\ and\ \citenamefont
  {Camsari}}]{aadit2023accelerating}%
  \BibitemOpen
  \bibfield  {author} {\bibinfo {author} {\bibfnamefont {N.~A.}\ \bibnamefont
  {Aadit}}, \bibinfo {author} {\bibfnamefont {M.}~\bibnamefont {Mohseni}},\
  and\ \bibinfo {author} {\bibfnamefont {K.~Y.}\ \bibnamefont {Camsari}},\
  }\bibfield  {title} {\bibinfo {title} {Accelerating adaptive parallel
  tempering with fpga-based p-bits},\ }in\ \href
  {https://doi.org/10.23919/VLSITechnologyandCir57934.2023.10185207} {\emph
  {\bibinfo {booktitle} {2023 IEEE Symposium on VLSI Technology and Circuits
  (VLSI Technology and Circuits)}}}\ (\bibinfo {year} {2023})\ pp.\ \bibinfo
  {pages} {1--2}\BibitemShut {NoStop}%
\bibitem [{\citenamefont {Chowdhury}\ \emph {et~al.}(2025)\citenamefont
  {Chowdhury}, \citenamefont {Aadit}, \citenamefont {Grimaldi}, \citenamefont
  {Raimondo}, \citenamefont {Raut}, \citenamefont {Lott}, \citenamefont
  {Mentink}, \citenamefont {Rams}, \citenamefont {Ricci-Tersenghi},
  \citenamefont {Chiappini}, \citenamefont {Theogarajan}, \citenamefont
  {Srimani}, \citenamefont {Finocchio}, \citenamefont {Mohseni},\ and\
  \citenamefont {Camsari}}]{chowdhury2025}%
  \BibitemOpen
  \bibfield  {author} {\bibinfo {author} {\bibfnamefont {S.}~\bibnamefont
  {Chowdhury}}, \bibinfo {author} {\bibfnamefont {N.~A.}\ \bibnamefont
  {Aadit}}, \bibinfo {author} {\bibfnamefont {A.}~\bibnamefont {Grimaldi}},
  \bibinfo {author} {\bibfnamefont {E.}~\bibnamefont {Raimondo}}, \bibinfo
  {author} {\bibfnamefont {A.}~\bibnamefont {Raut}}, \bibinfo {author}
  {\bibfnamefont {P.~A.}\ \bibnamefont {Lott}}, \bibinfo {author}
  {\bibfnamefont {J.~H.}\ \bibnamefont {Mentink}}, \bibinfo {author}
  {\bibfnamefont {M.~M.}\ \bibnamefont {Rams}}, \bibinfo {author}
  {\bibfnamefont {F.}~\bibnamefont {Ricci-Tersenghi}}, \bibinfo {author}
  {\bibfnamefont {M.}~\bibnamefont {Chiappini}}, \bibinfo {author}
  {\bibfnamefont {L.~S.}\ \bibnamefont {Theogarajan}}, \bibinfo {author}
  {\bibfnamefont {T.}~\bibnamefont {Srimani}}, \bibinfo {author} {\bibfnamefont
  {G.}~\bibnamefont {Finocchio}}, \bibinfo {author} {\bibfnamefont
  {M.}~\bibnamefont {Mohseni}},\ and\ \bibinfo {author} {\bibfnamefont {K.~Y.}\
  \bibnamefont {Camsari}},\ }\href {https://arxiv.org/abs/2503.10302} {\bibinfo
  {title} {Pushing the boundary of quantum advantage in hard combinatorial
  optimization with probabilistic computers}} (\bibinfo {year} {2025}),\
  \Eprint {https://arxiv.org/abs/2503.10302} {arXiv:2503.10302 [quant-ph]}
  \BibitemShut {NoStop}%
\bibitem [{\citenamefont {Mohseni}\ \emph {et~al.}(2025)\citenamefont
  {Mohseni}, \citenamefont {Scherer}, \citenamefont {Johnson}, \citenamefont
  {Wertheim}, \citenamefont {Otten}, \citenamefont {Aadit}, \citenamefont
  {Alexeev}, \citenamefont {Bresniker}, \citenamefont {Camsari}, \citenamefont
  {Chapman}, \citenamefont {Chatterjee}, \citenamefont {Dagnew}, \citenamefont
  {Esposito}, \citenamefont {Fahim}, \citenamefont {Fiorentino}, \citenamefont
  {Gajjar}, \citenamefont {Khalid}, \citenamefont {Kong}, \citenamefont
  {Kulchytskyy}, \citenamefont {Kyoseva}, \citenamefont {Li}, \citenamefont
  {Lott}, \citenamefont {Markov}, \citenamefont {McDermott}, \citenamefont
  {Pedretti}, \citenamefont {Rao}, \citenamefont {Rieffel}, \citenamefont
  {Silva}, \citenamefont {Sorebo}, \citenamefont {Spentzouris}, \citenamefont
  {Steiner}, \citenamefont {Torosov}, \citenamefont {Venturelli}, \citenamefont
  {Visser}, \citenamefont {Webb}, \citenamefont {Zhan}, \citenamefont {Cohen},
  \citenamefont {Ronagh}, \citenamefont {Ho}, \citenamefont {Beausoleil},\ and\
  \citenamefont {Martinis}}]{mohseni2024}%
  \BibitemOpen
  \bibfield  {author} {\bibinfo {author} {\bibfnamefont {M.}~\bibnamefont
  {Mohseni}}, \bibinfo {author} {\bibfnamefont {A.}~\bibnamefont {Scherer}},
  \bibinfo {author} {\bibfnamefont {K.~G.}\ \bibnamefont {Johnson}}, \bibinfo
  {author} {\bibfnamefont {O.}~\bibnamefont {Wertheim}}, \bibinfo {author}
  {\bibfnamefont {M.}~\bibnamefont {Otten}}, \bibinfo {author} {\bibfnamefont
  {N.~A.}\ \bibnamefont {Aadit}}, \bibinfo {author} {\bibfnamefont
  {Y.}~\bibnamefont {Alexeev}}, \bibinfo {author} {\bibfnamefont {K.~M.}\
  \bibnamefont {Bresniker}}, \bibinfo {author} {\bibfnamefont {K.~Y.}\
  \bibnamefont {Camsari}}, \bibinfo {author} {\bibfnamefont {B.}~\bibnamefont
  {Chapman}}, \bibinfo {author} {\bibfnamefont {S.}~\bibnamefont {Chatterjee}},
  \bibinfo {author} {\bibfnamefont {G.~A.}\ \bibnamefont {Dagnew}}, \bibinfo
  {author} {\bibfnamefont {A.}~\bibnamefont {Esposito}}, \bibinfo {author}
  {\bibfnamefont {F.}~\bibnamefont {Fahim}}, \bibinfo {author} {\bibfnamefont
  {M.}~\bibnamefont {Fiorentino}}, \bibinfo {author} {\bibfnamefont
  {A.}~\bibnamefont {Gajjar}}, \bibinfo {author} {\bibfnamefont
  {A.}~\bibnamefont {Khalid}}, \bibinfo {author} {\bibfnamefont
  {X.}~\bibnamefont {Kong}}, \bibinfo {author} {\bibfnamefont {B.}~\bibnamefont
  {Kulchytskyy}}, \bibinfo {author} {\bibfnamefont {E.}~\bibnamefont
  {Kyoseva}}, \bibinfo {author} {\bibfnamefont {R.}~\bibnamefont {Li}},
  \bibinfo {author} {\bibfnamefont {P.~A.}\ \bibnamefont {Lott}}, \bibinfo
  {author} {\bibfnamefont {I.~L.}\ \bibnamefont {Markov}}, \bibinfo {author}
  {\bibfnamefont {R.~F.}\ \bibnamefont {McDermott}}, \bibinfo {author}
  {\bibfnamefont {G.}~\bibnamefont {Pedretti}}, \bibinfo {author}
  {\bibfnamefont {P.}~\bibnamefont {Rao}}, \bibinfo {author} {\bibfnamefont
  {E.}~\bibnamefont {Rieffel}}, \bibinfo {author} {\bibfnamefont
  {A.}~\bibnamefont {Silva}}, \bibinfo {author} {\bibfnamefont
  {J.}~\bibnamefont {Sorebo}}, \bibinfo {author} {\bibfnamefont
  {P.}~\bibnamefont {Spentzouris}}, \bibinfo {author} {\bibfnamefont
  {Z.}~\bibnamefont {Steiner}}, \bibinfo {author} {\bibfnamefont
  {B.}~\bibnamefont {Torosov}}, \bibinfo {author} {\bibfnamefont
  {D.}~\bibnamefont {Venturelli}}, \bibinfo {author} {\bibfnamefont {R.~J.}\
  \bibnamefont {Visser}}, \bibinfo {author} {\bibfnamefont {Z.}~\bibnamefont
  {Webb}}, \bibinfo {author} {\bibfnamefont {X.}~\bibnamefont {Zhan}}, \bibinfo
  {author} {\bibfnamefont {Y.}~\bibnamefont {Cohen}}, \bibinfo {author}
  {\bibfnamefont {P.}~\bibnamefont {Ronagh}}, \bibinfo {author} {\bibfnamefont
  {A.}~\bibnamefont {Ho}}, \bibinfo {author} {\bibfnamefont {R.~G.}\
  \bibnamefont {Beausoleil}},\ and\ \bibinfo {author} {\bibfnamefont {J.~M.}\
  \bibnamefont {Martinis}},\ }\href {https://arxiv.org/abs/2411.10406}
  {\bibinfo {title} {How to build a quantum supercomputer: Scaling from
  hundreds to millions of qubits}} (\bibinfo {year} {2025}),\ \Eprint
  {https://arxiv.org/abs/2411.10406} {arXiv:2411.10406 [quant-ph]} \BibitemShut
  {NoStop}%
\bibitem [{\citenamefont {Vaswani}\ \emph {et~al.}(2023)\citenamefont
  {Vaswani}, \citenamefont {Shazeer}, \citenamefont {Parmar}, \citenamefont
  {Uszkoreit}, \citenamefont {Jones}, \citenamefont {Gomez}, \citenamefont
  {Kaiser},\ and\ \citenamefont {Polosukhin}}]{vaswani2023}%
  \BibitemOpen
  \bibfield  {author} {\bibinfo {author} {\bibfnamefont {A.}~\bibnamefont
  {Vaswani}}, \bibinfo {author} {\bibfnamefont {N.}~\bibnamefont {Shazeer}},
  \bibinfo {author} {\bibfnamefont {N.}~\bibnamefont {Parmar}}, \bibinfo
  {author} {\bibfnamefont {J.}~\bibnamefont {Uszkoreit}}, \bibinfo {author}
  {\bibfnamefont {L.}~\bibnamefont {Jones}}, \bibinfo {author} {\bibfnamefont
  {A.~N.}\ \bibnamefont {Gomez}}, \bibinfo {author} {\bibfnamefont
  {L.}~\bibnamefont {Kaiser}},\ and\ \bibinfo {author} {\bibfnamefont
  {I.}~\bibnamefont {Polosukhin}},\ }\href {https://arxiv.org/abs/1706.03762}
  {\bibinfo {title} {Attention is all you need}} (\bibinfo {year} {2023}),\
  \Eprint {https://arxiv.org/abs/1706.03762} {arXiv:1706.03762 [cs.CL]}
  \BibitemShut {NoStop}%
\bibitem [{\citenamefont {Mohseni}\ \emph
  {et~al.}(2023{\natexlab{b}})\citenamefont {Mohseni}, \citenamefont {Aadit},\
  and\ \citenamefont {Lott}}]{aadit2023}%
  \BibitemOpen
  \bibfield  {author} {\bibinfo {author} {\bibfnamefont {M.}~\bibnamefont
  {Mohseni}}, \bibinfo {author} {\bibfnamefont {N.~A.}\ \bibnamefont {Aadit}},\
  and\ \bibinfo {author} {\bibfnamefont {A.}~\bibnamefont {Lott}},\ }\href
  {https://github.com/usra-riacs/Nonlocal-Monte-Carlo} {\bibinfo {title}
  {Nonlocal monte carlo, github.com/usra-riacs/nonlocal-monte-carlo}} (\bibinfo
  {year} {2023}{\natexlab{b}})\BibitemShut {NoStop}%
\bibitem [{\citenamefont {Montanari}\ \emph {et~al.}(2008)\citenamefont
  {Montanari}, \citenamefont {Ricci-Tersenghi},\ and\ \citenamefont
  {Semerjian}}]{montanari2008}%
  \BibitemOpen
  \bibfield  {author} {\bibinfo {author} {\bibfnamefont {A.}~\bibnamefont
  {Montanari}}, \bibinfo {author} {\bibfnamefont {F.}~\bibnamefont
  {Ricci-Tersenghi}},\ and\ \bibinfo {author} {\bibfnamefont {G.}~\bibnamefont
  {Semerjian}},\ }\bibfield  {title} {\bibinfo {title} {Clusters of solutions
  and replica symmetry breaking in random k-satisfiability},\ }\href
  {https://doi.org/10.1088/1742-5468/2008/04/P04004} {\bibfield  {journal}
  {\bibinfo  {journal} {Journal of Statistical Mechanics: Theory and
  Experiment}\ }\textbf {\bibinfo {volume} {2008}},\ \bibinfo {pages} {P04004}
  (\bibinfo {year} {2008})}\BibitemShut {NoStop}%
\bibitem [{\citenamefont {Ans\'{o}tegui}\ \emph {et~al.}(2009)\citenamefont
  {Ans\'{o}tegui}, \citenamefont {Bonet},\ and\ \citenamefont
  {Levy}}]{ansotegui2009a}%
  \BibitemOpen
  \bibfield  {author} {\bibinfo {author} {\bibfnamefont {C.}~\bibnamefont
  {Ans\'{o}tegui}}, \bibinfo {author} {\bibfnamefont {M.~L.}\ \bibnamefont
  {Bonet}},\ and\ \bibinfo {author} {\bibfnamefont {J.}~\bibnamefont {Levy}},\
  }\bibfield  {title} {\bibinfo {title} {Towards industrial-like random sat
  instances},\ }in\ \href@noop {} {\emph {\bibinfo {booktitle} {Proceedings of
  the 21st International Joint Conference on Artificial Intelligence}}},\
  \bibinfo {series and number} {IJCAI'09}\ (\bibinfo  {publisher} {Morgan
  Kaufmann Publishers Inc.},\ \bibinfo {address} {San Francisco, CA, USA},\
  \bibinfo {year} {2009})\ pp.\ \bibinfo {pages} {387--392}\BibitemShut
  {NoStop}%
\bibitem [{\citenamefont {Friedrich}\ \emph {et~al.}(2017)\citenamefont
  {Friedrich}, \citenamefont {Krohmer}, \citenamefont {Rothenberger},\ and\
  \citenamefont {Sutton}}]{friedrich2017}%
  \BibitemOpen
  \bibfield  {author} {\bibinfo {author} {\bibfnamefont {T.}~\bibnamefont
  {Friedrich}}, \bibinfo {author} {\bibfnamefont {A.}~\bibnamefont {Krohmer}},
  \bibinfo {author} {\bibfnamefont {R.}~\bibnamefont {Rothenberger}},\ and\
  \bibinfo {author} {\bibfnamefont {A.~M.}\ \bibnamefont {Sutton}},\ }\bibfield
   {title} {\bibinfo {title} {Phase transitions for scale-free sat formulas},\
  }in\ \href@noop {} {\emph {\bibinfo {booktitle} {Proceedings of the
  Thirty-First AAAI Conference on Artificial Intelligence}}},\ \bibinfo {series
  and number} {AAAI'17}\ (\bibinfo  {publisher} {AAAI Press},\ \bibinfo {year}
  {2017})\ pp.\ \bibinfo {pages} {3893--3899}\BibitemShut {NoStop}%
\bibitem [{\citenamefont {Ignatiev}\ \emph {et~al.}(2018)\citenamefont
  {Ignatiev}, \citenamefont {Morgado},\ and\ \citenamefont
  {Marques{-}Silva}}]{ignatiev2018}%
  \BibitemOpen
  \bibfield  {author} {\bibinfo {author} {\bibfnamefont {A.}~\bibnamefont
  {Ignatiev}}, \bibinfo {author} {\bibfnamefont {A.}~\bibnamefont {Morgado}},\
  and\ \bibinfo {author} {\bibfnamefont {J.}~\bibnamefont {Marques{-}Silva}},\
  }\bibfield  {title} {\bibinfo {title} {{PySAT:} {A} {Python} toolkit for
  prototyping with {SAT} oracles},\ }in\ \href
  {https://doi.org/10.1007/978-3-319-94144-8_26} {\emph {\bibinfo {booktitle}
  {SAT}}}\ (\bibinfo {year} {2018})\ pp.\ \bibinfo {pages}
  {428--437}\BibitemShut {NoStop}%
\bibitem [{\citenamefont {R{\o}nnow}\ \emph {et~al.}(2014)\citenamefont
  {R{\o}nnow}, \citenamefont {Wang}, \citenamefont {Job}, \citenamefont
  {Boixo}, \citenamefont {Isakov}, \citenamefont {Wecker}, \citenamefont
  {Martinis}, \citenamefont {Lidar},\ and\ \citenamefont
  {Troyer}}]{troels2014}%
  \BibitemOpen
  \bibfield  {author} {\bibinfo {author} {\bibfnamefont {T.~F.}\ \bibnamefont
  {R{\o}nnow}}, \bibinfo {author} {\bibfnamefont {Z.}~\bibnamefont {Wang}},
  \bibinfo {author} {\bibfnamefont {J.}~\bibnamefont {Job}}, \bibinfo {author}
  {\bibfnamefont {S.}~\bibnamefont {Boixo}}, \bibinfo {author} {\bibfnamefont
  {S.~V.}\ \bibnamefont {Isakov}}, \bibinfo {author} {\bibfnamefont
  {D.}~\bibnamefont {Wecker}}, \bibinfo {author} {\bibfnamefont {J.~M.}\
  \bibnamefont {Martinis}}, \bibinfo {author} {\bibfnamefont {D.~A.}\
  \bibnamefont {Lidar}},\ and\ \bibinfo {author} {\bibfnamefont
  {M.}~\bibnamefont {Troyer}},\ }\bibfield  {title} {\bibinfo {title} {Defining
  and detecting quantum speedup},\ }\href
  {https://doi.org/10.1126/science.1252319} {\bibfield  {journal} {\bibinfo
  {journal} {Science}\ }\textbf {\bibinfo {volume} {345}},\ \bibinfo {pages}
  {420} (\bibinfo {year} {2014})}\BibitemShut {NoStop}%
\bibitem [{\citenamefont {{Gurobi Optimization, LLC}}(2025)}]{gurobi}%
  \BibitemOpen
  \bibfield  {author} {\bibinfo {author} {\bibnamefont {{Gurobi Optimization,
  LLC}}},\ }\href {https://www.gurobi.com} {\bibinfo {title} {{Gurobi Optimizer
  Reference Manual}}} (\bibinfo {year} {2025})\BibitemShut {NoStop}%
\bibitem [{\citenamefont {Schulman}\ \emph
  {et~al.}(2017{\natexlab{a}})\citenamefont {Schulman}, \citenamefont {Levine},
  \citenamefont {Moritz}, \citenamefont {Jordan},\ and\ \citenamefont
  {Abbeel}}]{schulman2017a}%
  \BibitemOpen
  \bibfield  {author} {\bibinfo {author} {\bibfnamefont {J.}~\bibnamefont
  {Schulman}}, \bibinfo {author} {\bibfnamefont {S.}~\bibnamefont {Levine}},
  \bibinfo {author} {\bibfnamefont {P.}~\bibnamefont {Moritz}}, \bibinfo
  {author} {\bibfnamefont {M.~I.}\ \bibnamefont {Jordan}},\ and\ \bibinfo
  {author} {\bibfnamefont {P.}~\bibnamefont {Abbeel}},\ }\href
  {https://arxiv.org/abs/1502.05477} {\bibinfo {title} {Trust region policy
  optimization}} (\bibinfo {year} {2017}{\natexlab{a}}),\ \Eprint
  {https://arxiv.org/abs/1502.05477} {arXiv:1502.05477 [cs.LG]} \BibitemShut
  {NoStop}%
\bibitem [{\citenamefont {Schulman}\ \emph
  {et~al.}(2017{\natexlab{b}})\citenamefont {Schulman}, \citenamefont {Wolski},
  \citenamefont {Dhariwal}, \citenamefont {Radford},\ and\ \citenamefont
  {Klimov}}]{schulman2017b}%
  \BibitemOpen
  \bibfield  {author} {\bibinfo {author} {\bibfnamefont {J.}~\bibnamefont
  {Schulman}}, \bibinfo {author} {\bibfnamefont {F.}~\bibnamefont {Wolski}},
  \bibinfo {author} {\bibfnamefont {P.}~\bibnamefont {Dhariwal}}, \bibinfo
  {author} {\bibfnamefont {A.}~\bibnamefont {Radford}},\ and\ \bibinfo {author}
  {\bibfnamefont {O.}~\bibnamefont {Klimov}},\ }\href
  {https://arxiv.org/abs/1707.06347} {\bibinfo {title} {Proximal policy
  optimization algorithms}} (\bibinfo {year} {2017}{\natexlab{b}}),\ \Eprint
  {https://arxiv.org/abs/1707.06347} {arXiv:1707.06347 [cs.LG]} \BibitemShut
  {NoStop}%
\bibitem [{\citenamefont {Lu}\ \emph {et~al.}(2022)\citenamefont {Lu},
  \citenamefont {Kuba}, \citenamefont {Letcher}, \citenamefont {Metz},
  \citenamefont {Schroeder~de Witt},\ and\ \citenamefont
  {Foerster}}]{lu2022discovered}%
  \BibitemOpen
  \bibfield  {author} {\bibinfo {author} {\bibfnamefont {C.}~\bibnamefont
  {Lu}}, \bibinfo {author} {\bibfnamefont {J.}~\bibnamefont {Kuba}}, \bibinfo
  {author} {\bibfnamefont {A.}~\bibnamefont {Letcher}}, \bibinfo {author}
  {\bibfnamefont {L.}~\bibnamefont {Metz}}, \bibinfo {author} {\bibfnamefont
  {C.}~\bibnamefont {Schroeder~de Witt}},\ and\ \bibinfo {author}
  {\bibfnamefont {J.}~\bibnamefont {Foerster}},\ }\bibfield  {title} {\bibinfo
  {title} {Discovered policy optimisation},\ }\href@noop {} {\bibfield
  {journal} {\bibinfo  {journal} {Advances in Neural Information Processing
  Systems}\ }\textbf {\bibinfo {volume} {35}},\ \bibinfo {pages} {16455}
  (\bibinfo {year} {2022})}\BibitemShut {NoStop}%
\bibitem [{\citenamefont {Heek}\ \emph {et~al.}(2024)\citenamefont {Heek},
  \citenamefont {Levskaya}, \citenamefont {Oliver}, \citenamefont {Ritter},
  \citenamefont {Rondepierre}, \citenamefont {Steiner},\ and\ \citenamefont
  {van {Z}ee}}]{flax2020github}%
  \BibitemOpen
  \bibfield  {author} {\bibinfo {author} {\bibfnamefont {J.}~\bibnamefont
  {Heek}}, \bibinfo {author} {\bibfnamefont {A.}~\bibnamefont {Levskaya}},
  \bibinfo {author} {\bibfnamefont {A.}~\bibnamefont {Oliver}}, \bibinfo
  {author} {\bibfnamefont {M.}~\bibnamefont {Ritter}}, \bibinfo {author}
  {\bibfnamefont {B.}~\bibnamefont {Rondepierre}}, \bibinfo {author}
  {\bibfnamefont {A.}~\bibnamefont {Steiner}},\ and\ \bibinfo {author}
  {\bibfnamefont {M.}~\bibnamefont {van {Z}ee}},\ }\href
  {http://github.com/google/flax} {\bibinfo {title} {{F}lax: A neural network
  library and ecosystem for {JAX}}} (\bibinfo {year} {2024})\BibitemShut
  {NoStop}%
\bibitem [{\citenamefont {Lange}(2022)}]{gymnax2022github}%
  \BibitemOpen
  \bibfield  {author} {\bibinfo {author} {\bibfnamefont {R.~T.}\ \bibnamefont
  {Lange}},\ }\href {http://github.com/RobertTLange/gymnax} {\bibinfo {title}
  {{gymnax}: A {JAX}-based reinforcement learning environment library}}
  (\bibinfo {year} {2022})\BibitemShut {NoStop}%
\bibitem [{\citenamefont {DeepMind}\ \emph {et~al.}(2020)\citenamefont
  {DeepMind}, \citenamefont {Babuschkin}, \citenamefont {Baumli}, \citenamefont
  {Bell}, \citenamefont {Bhupatiraju}, \citenamefont {Bruce}, \citenamefont
  {Buchlovsky}, \citenamefont {Budden}, \citenamefont {Cai}, \citenamefont
  {Clark}, \citenamefont {Danihelka}, \citenamefont {Dedieu}, \citenamefont
  {Fantacci}, \citenamefont {Godwin}, \citenamefont {Jones}, \citenamefont
  {Hemsley}, \citenamefont {Hennigan}, \citenamefont {Hessel}, \citenamefont
  {Hou}, \citenamefont {Kapturowski}, \citenamefont {Keck}, \citenamefont
  {Kemaev}, \citenamefont {King}, \citenamefont {Kunesch}, \citenamefont
  {Martens}, \citenamefont {Merzic}, \citenamefont {Mikulik}, \citenamefont
  {Norman}, \citenamefont {Papamakarios}, \citenamefont {Quan}, \citenamefont
  {Ring}, \citenamefont {Ruiz}, \citenamefont {Sanchez}, \citenamefont
  {Sartran}, \citenamefont {Schneider}, \citenamefont {Sezener}, \citenamefont
  {Spencer}, \citenamefont {Srinivasan}, \citenamefont {Stanojevi\'{c}},
  \citenamefont {Stokowiec}, \citenamefont {Wang}, \citenamefont {Zhou},\ and\
  \citenamefont {Viola}}]{deepmind2020jax}%
  \BibitemOpen
  \bibfield  {author} {\bibinfo {author} {\bibnamefont {DeepMind}}, \bibinfo
  {author} {\bibfnamefont {I.}~\bibnamefont {Babuschkin}}, \bibinfo {author}
  {\bibfnamefont {K.}~\bibnamefont {Baumli}}, \bibinfo {author} {\bibfnamefont
  {A.}~\bibnamefont {Bell}}, \bibinfo {author} {\bibfnamefont {S.}~\bibnamefont
  {Bhupatiraju}}, \bibinfo {author} {\bibfnamefont {J.}~\bibnamefont {Bruce}},
  \bibinfo {author} {\bibfnamefont {P.}~\bibnamefont {Buchlovsky}}, \bibinfo
  {author} {\bibfnamefont {D.}~\bibnamefont {Budden}}, \bibinfo {author}
  {\bibfnamefont {T.}~\bibnamefont {Cai}}, \bibinfo {author} {\bibfnamefont
  {A.}~\bibnamefont {Clark}}, \bibinfo {author} {\bibfnamefont
  {I.}~\bibnamefont {Danihelka}}, \bibinfo {author} {\bibfnamefont
  {A.}~\bibnamefont {Dedieu}}, \bibinfo {author} {\bibfnamefont
  {C.}~\bibnamefont {Fantacci}}, \bibinfo {author} {\bibfnamefont
  {J.}~\bibnamefont {Godwin}}, \bibinfo {author} {\bibfnamefont
  {C.}~\bibnamefont {Jones}}, \bibinfo {author} {\bibfnamefont
  {R.}~\bibnamefont {Hemsley}}, \bibinfo {author} {\bibfnamefont
  {T.}~\bibnamefont {Hennigan}}, \bibinfo {author} {\bibfnamefont
  {M.}~\bibnamefont {Hessel}}, \bibinfo {author} {\bibfnamefont
  {S.}~\bibnamefont {Hou}}, \bibinfo {author} {\bibfnamefont {S.}~\bibnamefont
  {Kapturowski}}, \bibinfo {author} {\bibfnamefont {T.}~\bibnamefont {Keck}},
  \bibinfo {author} {\bibfnamefont {I.}~\bibnamefont {Kemaev}}, \bibinfo
  {author} {\bibfnamefont {M.}~\bibnamefont {King}}, \bibinfo {author}
  {\bibfnamefont {M.}~\bibnamefont {Kunesch}}, \bibinfo {author} {\bibfnamefont
  {L.}~\bibnamefont {Martens}}, \bibinfo {author} {\bibfnamefont
  {H.}~\bibnamefont {Merzic}}, \bibinfo {author} {\bibfnamefont
  {V.}~\bibnamefont {Mikulik}}, \bibinfo {author} {\bibfnamefont
  {T.}~\bibnamefont {Norman}}, \bibinfo {author} {\bibfnamefont
  {G.}~\bibnamefont {Papamakarios}}, \bibinfo {author} {\bibfnamefont
  {J.}~\bibnamefont {Quan}}, \bibinfo {author} {\bibfnamefont {R.}~\bibnamefont
  {Ring}}, \bibinfo {author} {\bibfnamefont {F.}~\bibnamefont {Ruiz}}, \bibinfo
  {author} {\bibfnamefont {A.}~\bibnamefont {Sanchez}}, \bibinfo {author}
  {\bibfnamefont {L.}~\bibnamefont {Sartran}}, \bibinfo {author} {\bibfnamefont
  {R.}~\bibnamefont {Schneider}}, \bibinfo {author} {\bibfnamefont
  {E.}~\bibnamefont {Sezener}}, \bibinfo {author} {\bibfnamefont
  {S.}~\bibnamefont {Spencer}}, \bibinfo {author} {\bibfnamefont
  {S.}~\bibnamefont {Srinivasan}}, \bibinfo {author} {\bibfnamefont
  {M.}~\bibnamefont {Stanojevi\'{c}}}, \bibinfo {author} {\bibfnamefont
  {W.}~\bibnamefont {Stokowiec}}, \bibinfo {author} {\bibfnamefont
  {L.}~\bibnamefont {Wang}}, \bibinfo {author} {\bibfnamefont {G.}~\bibnamefont
  {Zhou}},\ and\ \bibinfo {author} {\bibfnamefont {F.}~\bibnamefont {Viola}},\
  }\href {http://github.com/google-deepmind} {\bibinfo {title} {The
  {D}eep{M}ind {JAX} {E}cosystem}} (\bibinfo {year} {2020})\BibitemShut
  {NoStop}%
\bibitem [{\citenamefont {Zhou}\ \emph {et~al.}(2023)\citenamefont {Zhou},
  \citenamefont {Dedieu}, \citenamefont {Kumar}, \citenamefont {Lehrach},
  \citenamefont {L{\'a}zaro-Gredilla}, \citenamefont {Kushagra},\ and\
  \citenamefont {George}}]{zhou2023}%
  \BibitemOpen
  \bibfield  {author} {\bibinfo {author} {\bibfnamefont {G.}~\bibnamefont
  {Zhou}}, \bibinfo {author} {\bibfnamefont {A.}~\bibnamefont {Dedieu}},
  \bibinfo {author} {\bibfnamefont {N.}~\bibnamefont {Kumar}}, \bibinfo
  {author} {\bibfnamefont {W.}~\bibnamefont {Lehrach}}, \bibinfo {author}
  {\bibfnamefont {M.}~\bibnamefont {L{\'a}zaro-Gredilla}}, \bibinfo {author}
  {\bibfnamefont {S.}~\bibnamefont {Kushagra}},\ and\ \bibinfo {author}
  {\bibfnamefont {D.}~\bibnamefont {George}},\ }\href
  {https://arxiv.org/abs/2202.04110} {\bibinfo {title} {Pgmax: Factor graphs
  for discrete probabilistic graphical models and loopy belief propagation in
  jax}} (\bibinfo {year} {2023}),\ \Eprint {https://arxiv.org/abs/2202.04110}
  {arXiv:2202.04110 [cs.LG]} \BibitemShut {NoStop}%
\end{thebibliography}%

\begin{acknowledgments}
This material is based upon work supported by the Defense Advanced Research Projects Agency (DARPA) 
under the Air Force Research Laboratory (AFRL) Agreement No.~FA8650-23-3-7313. 
The views, opinions, and/or findings expressed are those of the authors and should 
not be interpreted as representing the official views or policies of 
the Department of Defense or the U.S. Government.
We also gratefully acknowledge the generous funding of this work under NEUROTEC II 
(Verbundkoordinator / Förderkennzeichen: Forschungszentrum Jülich / 16ME0398K) 
by the Bundesministerium für Bildung und Forschung. 

D.D. would like to thank Gili Rosenberg, Martin Schuetz, Helmut Katzgraber, as well as Jan Finkbeiner and Jamie Lohoff 
for valuable comments and suggestions during the preparation of the manuscript. Authors acknowledge the open source version of 
NMC algorithm available at \cite{aadit2023}. M.M. would like to thank Aaron Lott for useful discussions.
\end{acknowledgments}

\section*{Author contributions}
D.\,D., M.\,M., and  J.\,P.\,S., conceived the idea, analyzed the data, and wrote the manuscript; D.\,D. wrote the code, designed, and 
ran the simulations.

\section*{Competing interests}
Authors declare no competing interests.

\newpage
\appendix
\section{Supplementary materials}
\subsection{\label{app:benchmarks}4-SAT benchmarks}
\subsubsection{\label{app:4sat_uf}Uniform random 4-SAT in the rigidity phase}
Uniform random k-SAT problems are a common combinatorial optimization benchmark 
exhibiting a rich variety of phase transition phenomena \cite{mezard2009}. The ``uniformity'' of this class 
(as opposed to the scale-free problems below) refers to the equal
probability of each variable $x_i,\, i \in [1, N]$ to appear in the clauses of the conjugate normal form of Eq.~\ref{eq:K-SAT}.

To benchmark our solvers, we have generated $384$ uniform random problems at sizes
$N = 500,\,1000,\,2000$ at the clause-to-variable ratio $\alpha = 9.884$ used in \cite{mohseni2021}. 
The clauses are not repeating, and the variables cannot appear in the same clause more than once.
The chosen ratio $\alpha$ corresponds to the rigidity phase that appears past the threshold $\alpha_r = 9.883$ 
for this random problem class \cite{montanari2008}. The generated instances are very hard for exact SAT solvers 
(CDCL-type \cite{biere2009} algorithms typically time-out).

\subsubsection{\label{app:4sat_sf}Industrial-inspired scale-free random 4-SAT}
\begin{figure}[ht]
    \centerline{\includegraphics[width=0.8\linewidth]{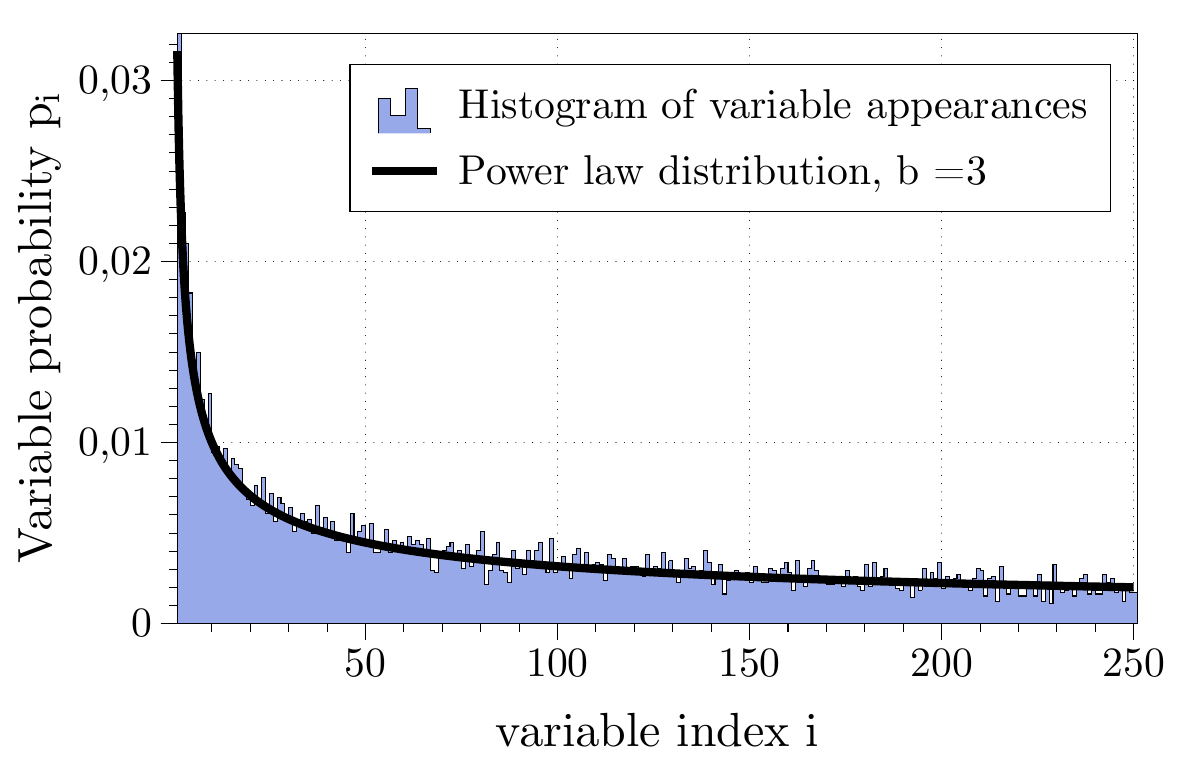}}
    \caption{
      Distribution of variables $x_i$ in a random scale-free 4-SAT instance of size $N=250$ 
      generated using the power law of Eq.~\ref{eq:pwr_law} with $\beta = 3$.
    }
    \label{fig:pwr_law_var_dist}
\end{figure}

Common k-SAT problems from industrial applications (structured instances)
were shown to have a distribution of variables close to a power-law (scale free) 
\cite{ansotegui2009a, friedrich2017}. As a result, it was suggested that \textit{random} instances 
generated with such power law, 
\begin{equation}
    p_i =  \frac{1}{N}\frac{b - 2}{b - 1} \left(\frac{N}{i}\right)^{1/(b - 1)},\; i\in \{1,\dots,N\}\,,
    \label{eq:pwr_law}
\end{equation}
would be representative of industrial applications while allowing for simpler benchmarking 
and prediction because of the ability to easily generate many problem instances. 
In addition, such problems were shown to exhibit a sat-unsat phase 
transition featuring an increasing algorithmic hardness akin to the uniform random case. 
Eq.~\ref{eq:pwr_law} corresponds to a steep distribution at smaller 
values of $b > 1$ (example in Fig.~\ref{fig:pwr_law_var_dist} for $b = 3$) 
and to the uniform distribution at $b \to \infty$.

\begin{figure}[ht]
\centerline{\includegraphics[width=0.8\linewidth]{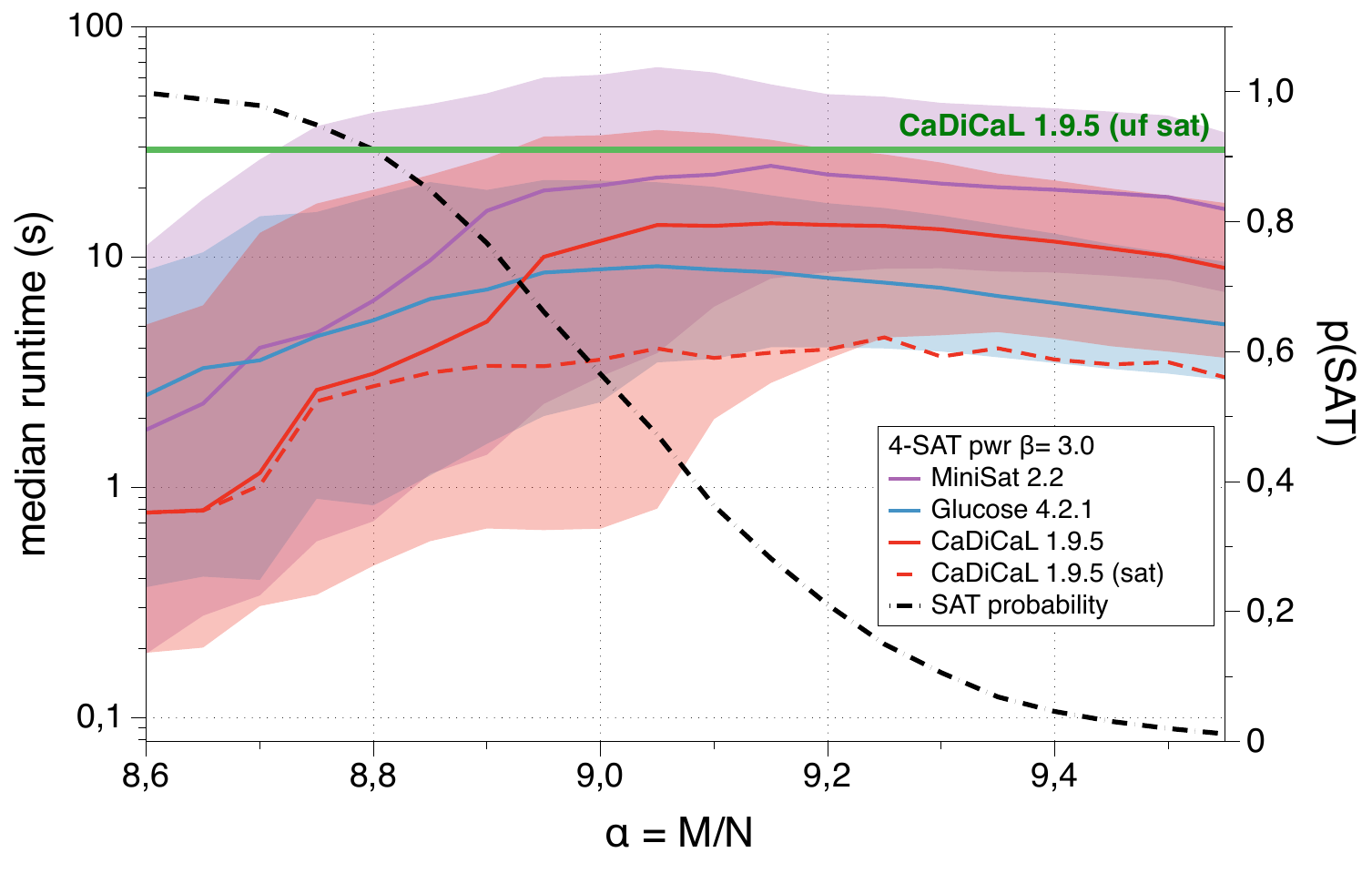}}
\caption{
    Median and 10/90 percentiles of runtimes for scale-free problems at $N = 250$, $b = 3.0$. 
    For comparison, the green line corresponds to the runtime of solving satisfiable uniform random problems at 
    $N = 250$. ``sat'' stands for statistics of only satisfiable instances. 
    The solvers are from the PySAT library \cite{ignatiev2018}.
}
\label{fig:4sat_pwr_hardness}
\end{figure}

As in the uniform random case, we have generated $384$ 4-SAT instances for training/benchmarking  
at the problem size $N = 250$ with $b = 3$ using the software provided in \cite{friedrich2017}. 
The clause-to-variable ratio $\alpha$ giving approximately highest runtime of exact solvers 
was $\alpha = 9.2$ (see Fig.~\ref{fig:4sat_pwr_hardness}). For comparison, we also show the time it takes an exact solver to 
find a solution for satisfiable uniform random $4$-SAT problems at the same problem size $N = 250$.
The uniform random problems at $N \ge 500$ used in this work are many orders of 
magnitude harder to satisfy.


\subsection{\label{app:metrics}Metrics}
\subsubsection{\label{app:time_to_solution}Time-to-solution}
One quantity of interest in this paper is time-to-solution Eq.~\ref{eq:tts}, which 
consists of a product of a monotonically \textit{increasing} term $\tau (N_{\rm{sw}})$
and a monotonically \textit{decreasing} term $\log{(1 - 0.99)}/\log{(1 - p(N_{\rm{sw}}))}$.
The resulting typically observed curves of $\mathrm{TTS}_{99}$ are given in Fig.~\ref{fig:sf250_80_tts}.
At first, $\mathrm{TTS}_{99}$ decreases, provided that some solutions are being found.
Later, $\mathrm{TTS}_{99}$ increases; i.e. new solutions are not being discovered quickly enough to justify the increased
runtime $\tau$ of an algorithm. In this case, an independent restart is preferred with a shorter $\tau$. 
If an algorithm continues to decrease $\mathrm{TTS}_{99}$ with increasing $\tau$, then it indicates effective 
exploration of the configuration space. An optimum value $\min[\mathrm{TTS}_{99}(N_{\rm{sw}})]$ characterizes 
the best observed performance of the algorithm with the other hyperparameters fixed.

Similarly to App.~\ref{app:residual_energy} we estimate the mean and standard deviation of 
$[\mathrm{TTS}_{99}]_{x}$ for typical (median $x = 0.5$) and hard ($x = 0.8$ percentile) instances with 
bootstrap resampling. Following the method in \cite{troels2014}, 
the probability of success for each instances is modeled with 
the beta distribution $\beta[N_{\rm{success}} + 0.5, N_{\rm{failure}} + 0.5]$, where 
$N_{\rm{success}} + N_{\rm{failure}} = N_{\rm{repl}}$.
The following number of independent repetitions $N_{\rm{repl}}$ (replicas) is used for benchmarking: 
$N_{\rm{repl}} = 4096$ for scale-free problems of $N = 250$, $N_{\rm{repl}} = 2048,\,1024,\,512$ for uniform random problems of 
$N = 500,\,1000,\,2000$ respectively. For hyperparameter optimization of SA/NMC the number of replicas is half of the 
aforementioned values.

We found that $\mathrm{TTS}_{99}$ of the chosen scale-free 4-SAT $N = 250$ benchmark is of the same order of magnitude (in MC sweeps)
as the time to approximation for the uniform random 4-SAT of this work at $N = 500$ with $2\times 10^{-4}$ approximation ratio
(1 unsatisfied clause). As a result, it takes approximately the same effort (in MC sweeps) to perform hyperparameter optimization, training, 
and benchmarking of algorithms for these problem classes.

\begin{figure}[ht]
    \centering
    \subfloat[\label{fig:uf1000_50_energy}]{
        \includegraphics[width=0.85\linewidth]{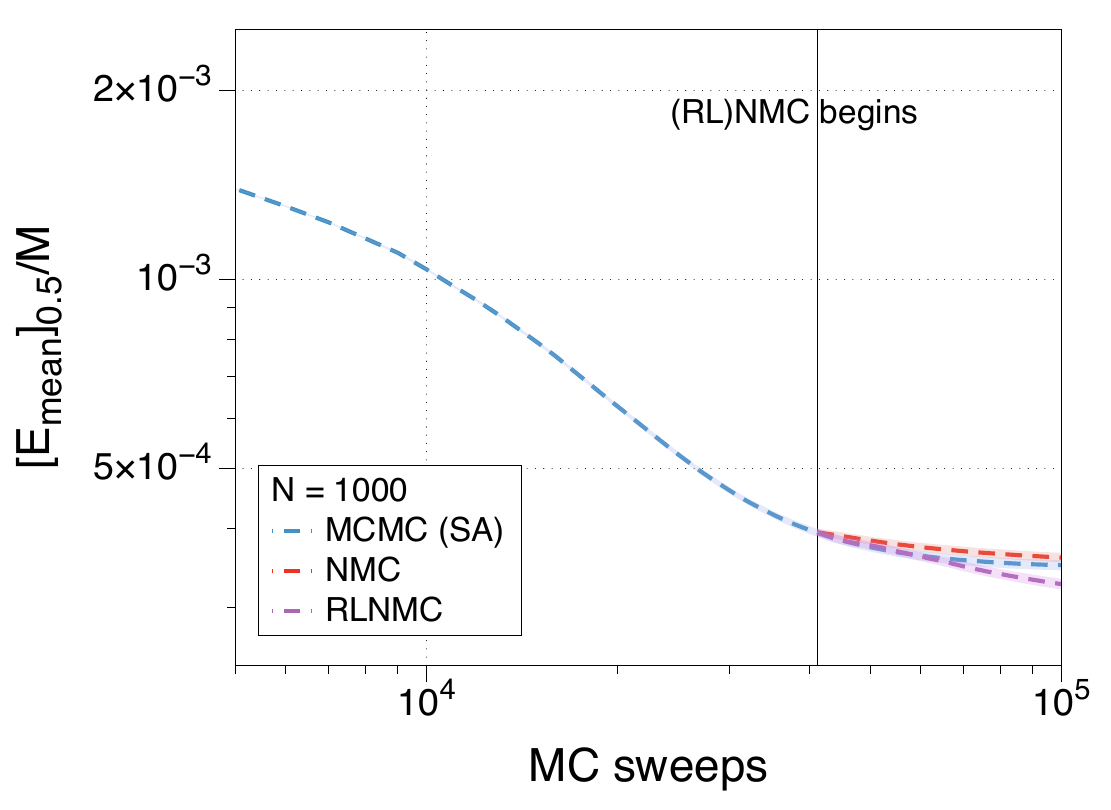}
    }
    \hfill
    \subfloat[\label{fig:uf2000_50_energy}]{
        \includegraphics[width=0.85\linewidth]{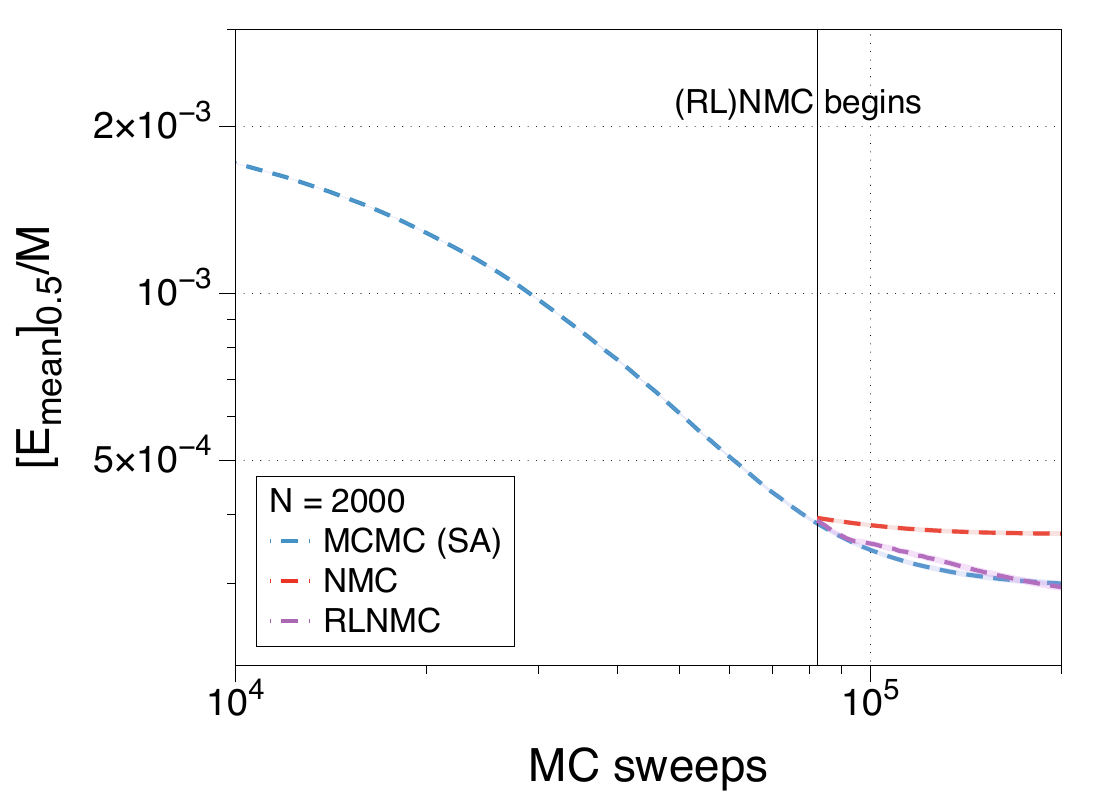}
    }
    \caption{\label{fig:uf_50_energy_extra}
        Median residual energy for uniform random 4-SAT (a) $N = 1000$, 
        (b) $N = 2000$ vs MC sweeps. NMC and RLNMC nonlocal moves begin at the indicated step.  
        For each instance the mean is over $1024$ and $512$ replicas respectively.
        The median and its standard deviation are estimated with bootstrap resampling
        of 320 used instances.
    }
\end{figure}

\subsubsection{\label{app:residual_energy}Residual energy}
In addition to TTS, we are interested in the residual energy $\langle E_{_{\rm{min}}}\rangle$, where 
$E_{_{\rm{min}}}$ is the minimum energy reached by a replica during its runtime (not necessarily the final energy), and 
the averaging is performed across all replicas for each instance of interest. To quantify the residual energy 
for typical instances in the chosen benchmark sets, we also define $[\langle E_{_{\rm{min}}}\rangle]_{0.5}$, 
i.e. the median across the set of instances.
Its expected value and standard deviation are estimated using bootstrap resampling of the instances with replacement.
Fig.~\ref{fig:uf_50_energy_extra} shows the energy vs MC sweeps at different problem sizes complementary to Fig.~\ref{fig:energy_scaling}
of the main text.

\subsubsection{\label{app:diversity_of_solutions}Diversity of solutions}
We define diversity of solutions following the prescription of \cite{mohseni2023}.
Consider $K$ independent replicas of SA/NMC/RLNMC, 
each possibly containing a solution $\bm{\sigma}_k,\, k \le K$ within a given approximation ratio.
We collect a set of solutions $\{\bm{\sigma}_k\}$ from each replica.
First, for every pair of solutions $\bm{\sigma}_{k'}, \bm{\sigma}_{k''}$ from the set 
we compute their mutual normalized Hamming distance $d_{k', k''} = d(\bm{\sigma}_{k'}, \bm{\sigma}_{k''})/N$. 
Second, for a given diversity threshold $R$ we construct an undirected graph $G(R)$, where each node 
$k$ corresponds to the solution $\bm{\sigma}_k$, and edges are present if $d_{k', k''} \le R$. 
When $R = 0$, then the resulting graph has no edges: there are no two identical solutions in the set $\{\bm{\sigma}_k\}$.
When $R = 1$, then we obtain a fully connected graph. 

\begin{figure}[ht]
    \centerline{\includegraphics[width=0.82\linewidth]{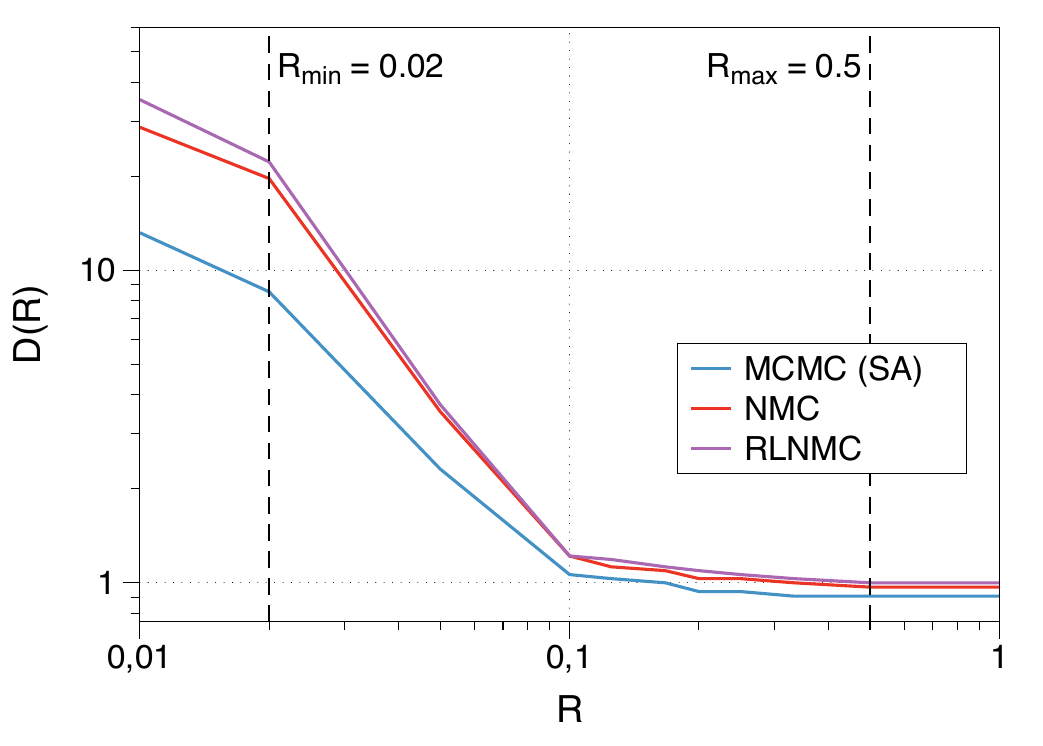}}
    \caption{
      Average $D(R)$ for instances in Fig.~\ref{fig:diversity} at differenct values of $R$.
    }
    \label{fig:sf250_diversity_r}
\end{figure}

For a chosen $R \in [0, 1]$, the diversity of solutions is 
defined as the maximum independent set (MIS) of the $G(R)$ graph. The fully connected graph $G(R = 1)$ has the 
smallest diversity $D = 1$, and the disconnected graph $G(R = 0)$ has diversity 
equal to the cardinality of the set of solutions $D = |\{\bm{\sigma}_k\}|$ (size of the graph).
Finally, we define the diversity integral of Eq.~\ref{eq:diversity_definition}
as the metric characterizing the performance of solvers in Sec.~\ref{sec:diversity_of_solutions}.

First, the set of solutions $\{\bm{\sigma}_k\}$ is accumulated from $2048$ independent replicas of SA/NMC/RLNMC.
Next, the integral is approximated by exactly solving the MIS problem
with Gurobi \cite{gurobi} for several diversity graphs $G(R_i)$ at $R_i \in [R_{\rm{min}}, R_{\rm{max}}]$ 
and using the corresponding finite sum. The lower cutoff value $R_{\rm{min}}$ is chosen when the average value of $D(R)$ over the tested instances starts to sharply decline indicating a possible typical size of basins of attraction (see Fig.~\ref{fig:sf250_diversity_r}). The higher cutoff $R_{\rm{max}}$ is chosen as the value of $R$ so that at $R > R_{\rm{max}}$ diversity $D(R)$ does not change.

\subsection{\label{app:training_details}Reinforcement learning details}
Proximal policy optimization (PPO) is a reinforcement learning algorithm within the large family of 
policy gradient methods. PPO clips an RL objective function 
so that during training updates a new policy $\pi_\theta$ is not too far from the old $\pi_{\theta_{\mathrm{old}}}$,
determined by a hyperparameter $\epsilon_{\rm{clip}}\in (0,1)$.
This results in an increased stability of RL training and higher performance 
across multiple benchmarks \cite{schulman2017a, schulman2017b}.

\begin{algorithm}[H]
    \caption{RLNMC training (simplified)}
    \label{alg:rlnmc_training}
    \begin{algorithmic}[1]
    \renewcommand{\algorithmicrequire}{\textbf{Input:}}
    \renewcommand{\algorithmicensure}{\textbf{Output:}}
    \REQUIRE Training instances; RL and NMC hyperparameters; 
        initial policy $\pi_{\theta_0}$, value function $V_{\phi_0}$.
    \FOR{\textit{repetition} $<$ $N_{\rm{train\,reps.}}$}
        \FOR{\textit{instance} in a set of instances}
            \FOR{\textit{episode} $<$ $N_{\rm{eps.}}$}
            \STATE Initialize $N_{\rm{repl}}$ replicas of \textit{instance} in a random state
                and reach a local minimum by SA running from $\beta_i$ to $\beta^0 \equiv \beta_{\rm{NMC}}$: 
                RL state $\mathbf{s}^0$.
                \FOR{NMC step $<$ $N_{\rm{NMC\,steps}}$}
                    \STATE Infer the backbone probability $\bm{p}^t$ in state $\mathbf{s}^t$ 
                        by the policy $\pi_{\theta}$ for each replica in parallel; sample action $\bm{a}^t$ from $\bm{p}^t$.
                    \STATE Perform NMC jump of Alg.~\ref{alg:nmc} at $\beta^t$ for each replica in parallel: 
                    $\mathbf{s}^t \to \mathbf{s}^{t+1}$.
                    \STATE Compare $\mathbf{s}^t$ to $\mathbf{s}^{t+1}$ and collect rewards $r^t$.
                    \STATE [each $N_{\rm{steps\,per\,upd.}}$] Do PPO training
                        with the collected trajectories $\tau = [\bm{a}^t, \bm{s}^t, r^t]$
                        updating the parameters $\theta$ of the policy $\pi_\theta$:
                        \begin{equation}
                            \theta = \mathrm{argmax}\sum_{\tau_i}\sum_{t = 0}^T L(\theta|\theta_{\mathrm{old}}, \bm{a}^t, \bm{s}^t, r^t)\,,
                            \label{eq:PPO_update}
                        \end{equation}
                    where $L$ is given by Eq.~\ref{eq:ppo_objective}.
                    Update the value function $V_{\phi}$ minimizing the least squares with rewards-to-go $R_{t}$.
                    \STATE Change temperature following the SA schedule: 
                        $\beta^{t+1} = \beta^t + \Delta \beta \in [\beta_{\rm{NMC}}, \beta_{f}]$
                \ENDFOR
            \ENDFOR
        \ENDFOR
    \ENDFOR
    \ENSURE Trained policy $\pi_\theta$.
    \end{algorithmic}
\end{algorithm}

We use the GPU accelerated JAX \cite{jax2018github} implementation of PPO given in \cite{lu2022discovered}, 
which searches with SGD for $\theta$ that maximize the objective function (see also Alg.~\ref{alg:rlnmc_training})
\begin{multline}
    L(\theta|\theta_{\mathrm{old}}, \bm{a}^t, \bm{s}^t, r) = \\
    = \min{\left(
        \frac{\pi_{\theta}(\bm{a}^t| \bm{s}^t)}{\pi_{\theta_{\mathrm{old}}}(\bm{a}^t| \bm{s}^t)}A^t_\gamma(\bm{s}^{t}, \bm{a}^{t}), 
        g[\epsilon_{\rm{clip}}, A^t_\gamma(\bm{s}^{t}, \bm{a}^{t})]
    \right)}\,,
    \label{eq:ppo_objective}
\end{multline}
where $A^t_\gamma(\bm{s}^{t}, \bm{a}^{t})$ is the advantage of taking the action $\bm{a}^{t}$ 
with the discount factor $\gamma$, policy function $\pi_{\theta}$ was specified in App.~\ref{sec:rlnmc_details}; 
the clipping function is $g(\epsilon_{\rm{clip}}, A) = (1+\epsilon_{\rm{clip}})A$, if $A > 0$, 
and $g(\epsilon_{\rm{clip}}, A) = (1-\epsilon_{\rm{clip}})A$, if $A < 0$.
Simply put, the advantage $A^t_\gamma$ compares the observed discounted rewards 
$R_{t} = \sum_{t'= t}^T\gamma^{t'-t} r^{t'}$ 
when following a policy $\pi_{\theta_{\rm{old}}}$ to the expectation estimated by the value function $V_\phi$ (which shares
many of its parameters with $\pi_{\theta}$). As a result, the updates of the policy $\pi_{\theta}$
reinforce actions having a positive advantage (better than expected) and discourage actions leading to negative advantage.
However, the difference of $\pi_{\theta}$ from $\pi_{\theta_{\rm{old}}}$ cannot exceed the bound $\epsilon_{\rm{clip}}$.

\begin{table}
    \centering
    \caption{\label{tab:ppo_hyperparameters} PPO RLNMC training hyperparameters.}
    \begin{tabular}{l|c|c}
        \toprule
        \textbf{Parameters} & \textbf{Uniform} & \textbf{Scale-free} \\
        \midrule
        Learning rate & $10^{-3}\to 10^{-4}$ & $10^{-3}\to 10^{-5}$ \\
        Epochs & 5 & 5 \\
        $N_{\rm{NMC\,steps}}$ & 51 & 54 \\
        minibatch & 64 & 64 \\
        $N_{\rm{replicas}}$ & 2048 & 2048 \\
        $N_{\rm{sw}}$ & 200 & 100 \\
        $N_{\rm{steps\,per\,upd.}}$ & 17 & 18 \\
        $N_{\rm{eps.}}$ & 2 & 5 \\
        $K$ & 64 & 64 \\
        $N_{\rm{train\,reps.}}$ & 5 & 5 \\
        $\gamma$ & 0.75 & 0.75 \\
        $\lambda_{\rm{GAE}}$ & 0.95 & 0.95 \\
        $c_{\rm{vf}}$ & 0.25 & 0.25 \\
        $c_{\rm{ent}}$ & $10^{-3}$ & $10^{-3}$ \\
        $\epsilon_{\rm{clip}}$ & 0.25 & 0.25 \\
        \bottomrule
    \end{tabular}
\end{table}

We summarize the RLNMC training in Alg.~\ref{alg:rlnmc_training}. A set of $K = 64$ instances is used for training 
of the RLNMC policies $\pi_{\theta}$ for each problem class. For $N_{\rm{train\,reps.}}$ number of repetitions we sequentially 
choose an instance that would be used for training. This instance is run for $N_{\rm{eps.}}\times N_{\rm{NMC\,steps}}$ 
number of of NMC steps in each of the $N_{\rm{repl}}$ replicas in parallel. If a replica reaches the ground state 
($2\times 10^{-4}$ approximation) in the scale-free $N=250$ (uniform random $N=500$) case, then the episode is restarted.
Every replica at the beginning of an episode is initialized 
with SA and then $N_{\rm{NMC\,steps}}$ NMC steps are performed.
Every $N_{\rm{steps\,per\,upd.}}$ steps, the accumulated trajectories are collected (rollout data) and used for 
the PPO update of $\pi_\theta$ with a random shuffling of the mini-batches for a certain number of epochs. 
As a result, the total number of NMC steps used for RLNMC training is 
$64\times N_{\rm{train\,reps.}}\times N_{\rm{eps.}}\times N_{\rm{NMC\,steps}} \times N_{\rm{repl}}$.
The learning rate is gradually decreased over the course of the training from $\rm{LR}_{\rm{init}}$ to 
$\rm{LR}_{\rm{final}}$. The training hyperparameters are explicitly given in Tab.~\ref{tab:ppo_hyperparameters}.

Random solver initializations, MCMC sampling trajectories, RL stochastic policy actions all depend on the random seed
given to RLNMC. As a result, we have performed multiple training attempts of RLNMC described above. At the end 
of the training trials, each resulting final policy is tested for its performance at solving the 4-SAT problems used for RL training,
because the discounted reward maximization of Eq.~\ref{eq:rl_reward} is not the quantity of interest in this work 
(TTS$_{99}$ and energy minimization are). The best performing policy is used for benchmarking of instances not seen during training 
and its results are reported in this paper in the scale-free problem class case, giving a clear advantage as shown in Fig.~\ref{fig:sf250_80_tts}. 
In the case of the uniform random problem class, we further fine-tuned the best model in addition to the effort reported in 
Tab.~\ref{tab:ppo_hyperparameters} with learning rate 
$10^{-4} \to 10^{-5}$, $N_{\rm{NMC\,steps}} = 50$, $N_{\rm{steps\,per\,upd.}} = 25$, $N_{\rm{train\,reps.}} = 3$, 
and the minibatch size 32 (other hyperparameters being the same). 

\subsubsection{\label{app:nmc_schedules}RLNMC schedules}

\begin{figure}[ht]
    \centering
    \subfloat[\label{fig:uf500_bb_dist}]{
        \includegraphics[width=0.485\linewidth]{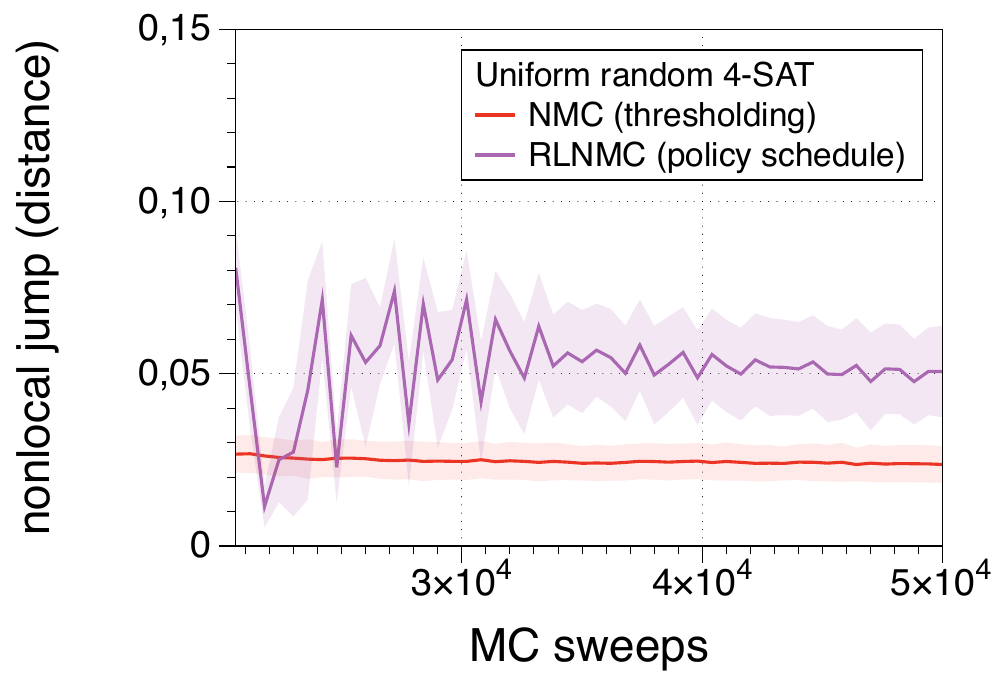}
    }
    \subfloat[\label{fig:sf250_bb_dist}]{
        \includegraphics[width=0.485\linewidth]{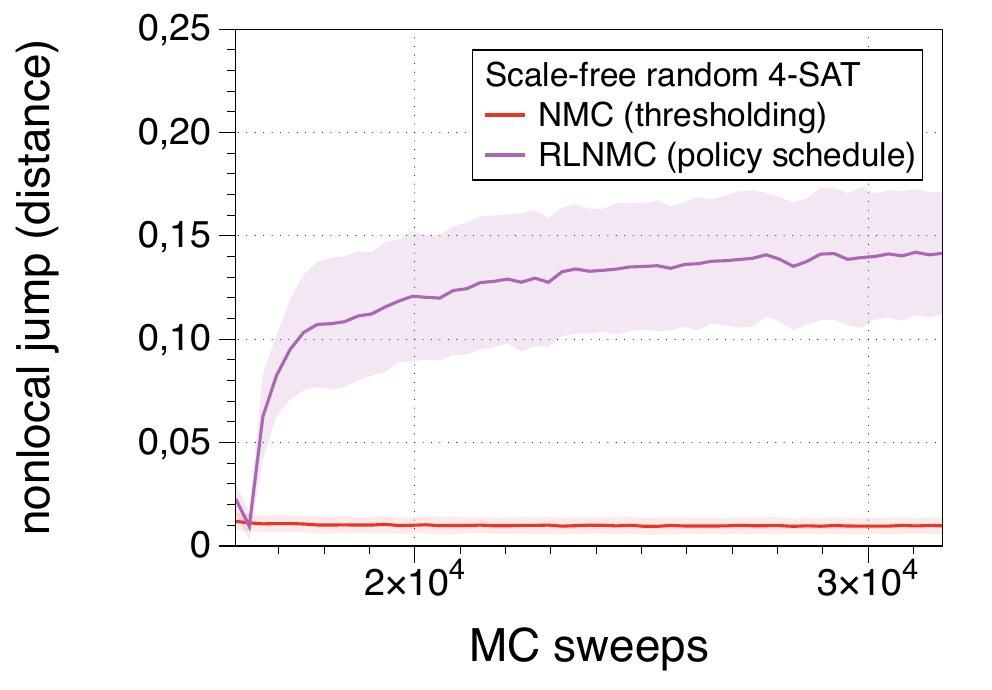}
    }
    \hfill
    \subfloat[\label{fig:uf500_bb_de}]{
        \includegraphics[width=0.485\linewidth]{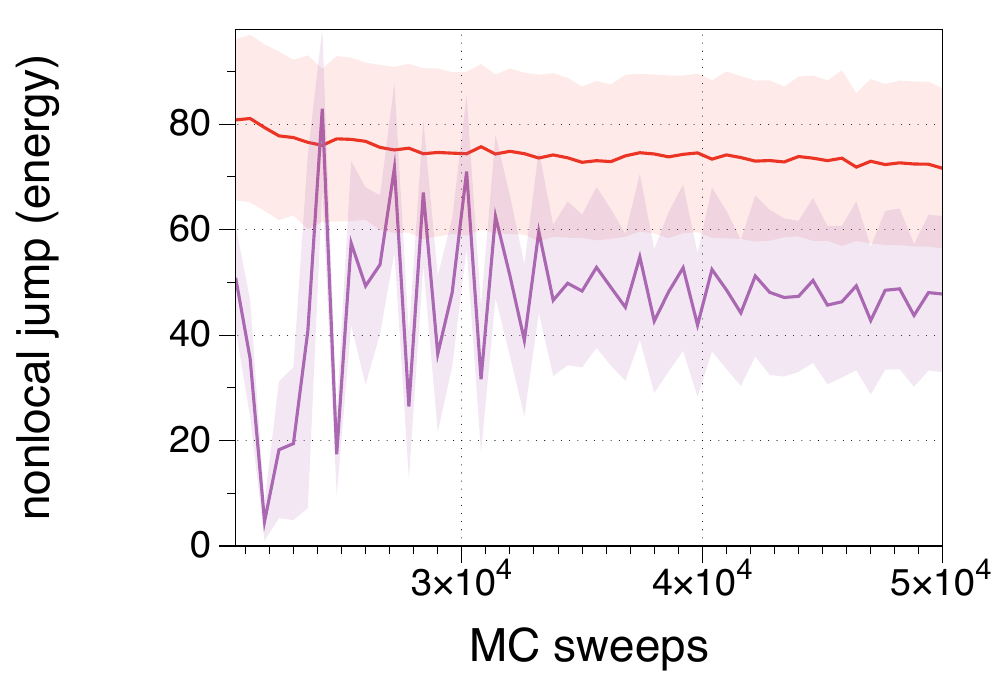}
    }
    \subfloat[\label{fig:sf250_bb_de}]{
        \includegraphics[width=0.485\linewidth]{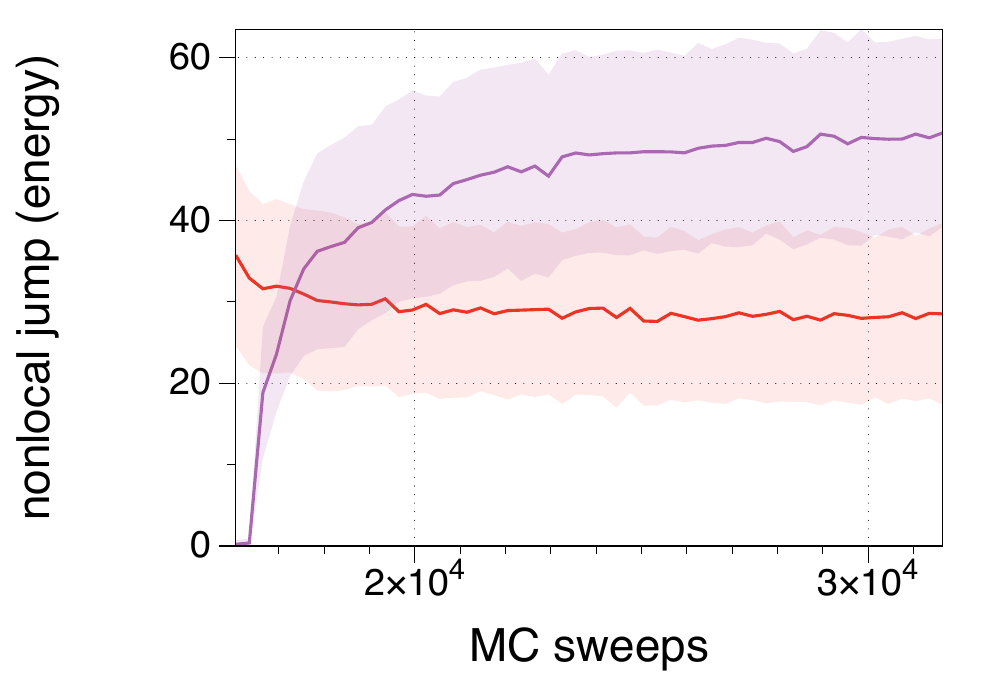}
    }
    \caption{\label{fig:bb_stats}
        Nonlocal nonequilibrium jump statistics for optimized NMC and trained RLNMC 
        algorithms for (a, c) uniform random class, (b, d) scale-free random class. 
        The trajectories are for a single instance averaged over multiple (512) replicas: 
        the mean and the standard deviation shown.
        The distance is normalized by the problem size, the energy increase of the 
        NMC excitation (see Fig.~\ref{fig:nmc_sketch}) is the number of unsatisfied clauses.
    }
\end{figure}

In Fig.~\ref{fig:bb_stats} we show the nonlocal move schedules created by the NMC and RLNMC policies for
both problem classes. The jump distance (fraction of variables flipped) in Figs.~\ref{fig:uf500_bb_dist}-\ref{fig:sf250_bb_dist}, 
as well as the excitation energy in Figs.~\ref{fig:uf500_bb_de}-\ref{fig:sf250_bb_de} are shown (step 1 in Fig.~\ref{fig:nmc_sketch}). 

Firstly, on average RLNMC follows a nonlocal move schedule initially reducing the size of the 
jump and later increasing the size of jumps with some saturation. A similar schedule achieved by RL, but for the SA temperature, 
was demonstrated in \cite{mills2020}. NMC does not follow a schedule, which would have needed to be handcrafted; however, 
there is a small reduction of the backbone size over time because the basins with the strongest magnetizations of variables are easily escaped from.
Secondly, we observe that NMC makes relatively small jumps of high energy excitation,
while RLNMC has learned to perform considerably more distant moves 
(``horizontal'' and not ``vertical'' in Fig.~\ref{fig:figure_one}). In this sense, RLNMC has learned more nonlocal moves, 
which holds promise for research aimed at addressing the overlap-gap-property's algorithmic challenges.

\subsubsection{\label{app:cost_at_test_time}Implementation and computational cost of RLNMC}

\begin{figure}[ht]
    \centering
    \includegraphics[width=0.9\linewidth]{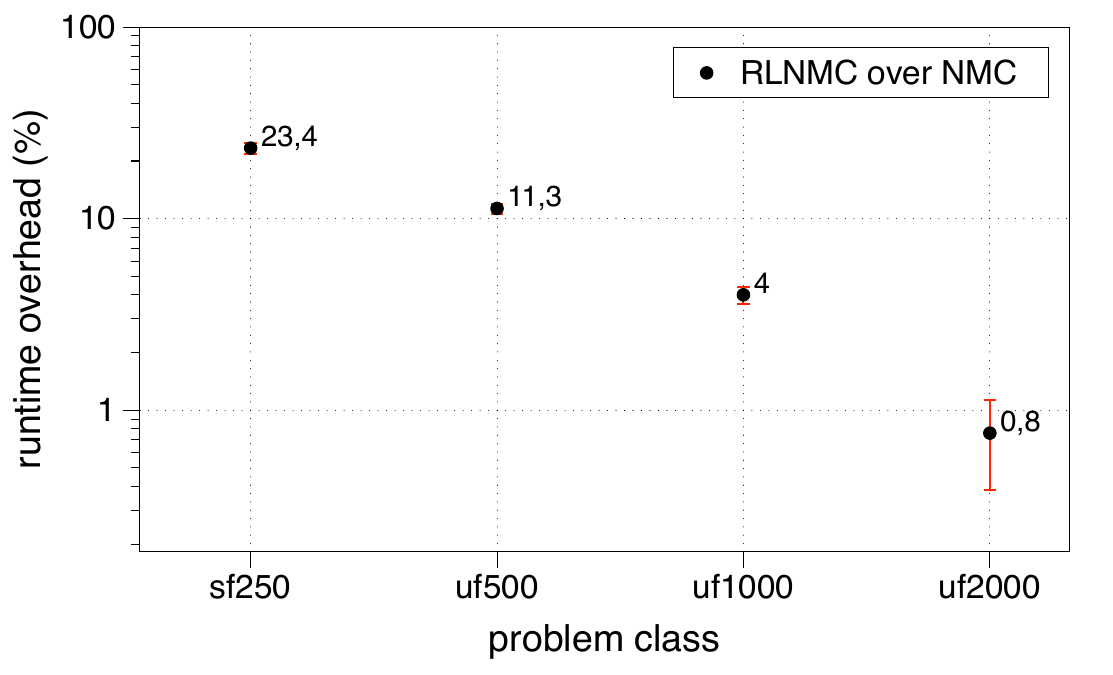}
    \caption{
        \label{fig:RLNMC_overhead} 
        RLNMC policy inference overhead compared to NMC for scale-free (``sf250'') and uniform random
        (``uf500'', ``uf1000'', ``uf2000'') problems of this work. 
        Data averaged over different instances and varying number of parallel replicas on a GPU.
        The scale-free problems overhead at $N=250$ is taken into account in Fig.~\ref{fig:sf250_80_tts}.
    }
\end{figure}

All routines of SA/NMC/RLNMC algorithms, including the MCMC sampling, policy neural network inference, RL training
in this work are implemented using high performance array computing python library JAX \cite{jax2018github}. 
The used packages include: 
Flax \cite{flax2020github}, 
gymnax \cite{gymnax2022github}, 
purejaxrl \cite{lu2022discovered}, 
distrax and optax \cite{deepmind2020jax}.
Our implementation of NMC/RLNMC also supports Loopy Belief Propagation with surrogate Hamiltonians
of \cite{mohseni2021} using the GPU accelerated PGMAX library \cite{zhou2023}; 
however, we have not used it in this paper and leave exploring this method with RL for future work.

The realised implementation supports three higher-order problem formulations: 
p-spin Ising ($p \ge 2$), PUBO, and weighted CNF, including the Belief Propagation 
estimation of correlations natively in each graph. The code is tailored for sparse problems, e.g.
the number of factors (non-zero coupling terms) scaling with the problem size as $O(N)$.

The computational cost of the policy shown in Fig.~\ref{fig:policy_arch} needs to be taken into account, when reporting 
the time-to-solution results. To get the scaling results in Sec.~\ref{sec:residual_energy} we have
chosen to increase the total number of sweeps but keep the number of NMC/RLNMC policy calls. This has proven to be successful  
for RLNMC as shown in Fig.~\ref{fig:energy_scaling}. An additional benefit is the reduced overhead shown 
in Fig.~\ref{fig:RLNMC_overhead}. At $N = 2000$, compared to the computational cost of NMC, the cost of the recurrent policy 
is less than $1\%$. The tests were performed on the NVIDIA L40S GPU for different numbers of parallel replicas 
within the allowed memory limits.

\end{document}